\documentclass[10pt,twocolumn,letterpaper]{article}

\usepackage{cvpr}              
\definecolor{cvprblue}{rgb}{0.21,0.49,0.74}
\usepackage[pagebackref,breaklinks,colorlinks,allcolors=cvprblue]{hyperref}

\usepackage{graphicx}
\usepackage{amsmath}
\usepackage{amssymb}
\usepackage[utf8]{inputenc} 
\usepackage[T1]{fontenc}    
\usepackage{url}            
\usepackage{booktabs}       
\usepackage{amsfonts}       
\usepackage{nicefrac}       
\usepackage{microtype}      
\usepackage[table,dvipsnames]{xcolor}         
\usepackage{colortbl}
\usepackage{amsmath}
\usepackage{makecell}
\usepackage{multirow}
\usepackage{subcaption}
\usepackage{float}
\usepackage{pifont}
\usepackage{arydshln} 

\usepackage{times}
\usepackage{epsfig}
\usepackage{graphicx}
\usepackage{amsmath}
\usepackage{amssymb}
\usepackage{colortbl}
\usepackage{booktabs} 
\usepackage{multirow}
\usepackage{makecell}
\usepackage{float}
\usepackage{enumitem} 
\usepackage{arydshln}
\usepackage{bbding}
\usepackage{animate}
\usepackage{grffile}
\usepackage{siunitx} %
\usepackage[table, usenames, dvipsnames]{xcolor}
\usepackage{tabularx}
\usepackage{bbm}
\usepackage{caption} 
\usepackage{subcaption}
\usepackage{pifont}     %
\usepackage{amsfonts}
\usepackage{tikz}
\usepackage{gensymb}
\usepackage{siunitx}

\usepackage[T1]{fontenc}
\usepackage[scaled=0.9]{inconsolata}

\newcommand{\project}{\textbf{DROID-W (Ours)}}

\colorlet{colorFst}{Green!25}       
\colorlet{colorSnd}{SpringGreen!45} 
\colorlet{colorTrd}{Yellow!30}      
\colorlet{colorLow}{darkgray!30}    
\newcommand{\fs}{\cellcolor{colorFst}\bf}   
\newcommand{\nd}{\cellcolor{colorSnd}}      
\newcommand{\rd}{\cellcolor{colorTrd}}      

\def\paperID{} 
\def\confName{CVPR}
\def\confYear{2026}

\title{DROID-SLAM in the Wild}

\author{Moyang Li$^{1}$\footnotemark[1]\qquad Zihan Zhu$^{1}$\footnotemark[1]\qquad Marc Pollefeys$^{1,2}$\qquad Daniel Barath$^{1}$\\
$^{1}$ETH Zurich\qquad $^{2}$Microsoft\\
}

\newcommand{\bd}{\mathbf{d}}\newcommand{\bD}{\mathbf{D}}
\newcommand{\be}{\mathbf{e}}\newcommand{\bE}{\mathbf{E}}
\newcommand{\bF}{\mathbf{F}} 
\newcommand{\bG}{\mathbf{G}}

\newcommand{\bI}{\mathbf{I}}

\newcommand{\bp}{\mathbf{p}}

\newcommand{\bu}{\mathbf{u}}

\newcommand{\bw}{\mathbf{w}}

\newcommand{\figref}[1]{Fig.~\ref{#1}}
\newcommand{\secref}[1]{Sec.~\ref{#1}}
\newcommand{\secrefn}[1]{Sec.\textcolor[rgb]{0,0.55,0.85}{~\ref*{#1}}}

\newcommand{\eqnref}[1]{Eq.~\eqref{#1}}
\newcommand{\tabref}[1]{Table~\ref{#1}}
\newcommand{\tabrefn}[1]{Table\textcolor[rgb]{0,0.55,0.85}{~\ref*{#1}}}

\makeatletter
\DeclareRobustCommand\onedot{\futurelet\@let@token\@onedot}
\def\@onedot{\ifx\@let@token.\else.\null\fi\xspace}
\def\eg{e.g\onedot} 
\def\ie{i.e\onedot}

\makeatother

\newcommand{\boldparagraph}[1]{\vspace{0.2em}\noindent{\bf #1.}}

\renewcommand{\paragraph}[1]{\boldparagraph{#1}}

\definecolor{darkgreen}{rgb}{0,0.7,0}
\definecolor{newyellow}{rgb}{1,0.8,0.05}
\definecolor{newgreen}{rgb}{0.2,0.8,0.2}

\definecolor{lightyellow}{RGB}{255,255,180} 

\newcommand{\cmark}{\color{ForestGreen}{\ding{51}}}%
\newcommand{\xmark}{\color{WildStrawberry}{\ding{55}}}%

\makeatletter
\def\adl@drawiv#1#2#3{%
        \hskip.5\tabcolsep
        \xleaders#3{#2.5\@tempdimb #1{1}#2.5\@tempdimb}%
                #2\z@ plus1fil minus1fil\relax
        \hskip.5\tabcolsep}
\newcommand{\cdashlinelr}[1]{%
  \noalign{\vskip\aboverulesep
           \global\let\@dashdrawstore\adl@draw
           \global\let\adl@draw\adl@drawiv}
  \cdashline{#1}
  \noalign{\global\let\adl@draw\@dashdrawstore
           \vskip\belowrulesep}}
\makeatother

\begin{document}
\twocolumn[{%
\renewcommand\twocolumn[1][]{#1}%
\maketitle
\vspace{-10mm}
\begin{center}
    \includegraphics[width=0.95\textwidth]{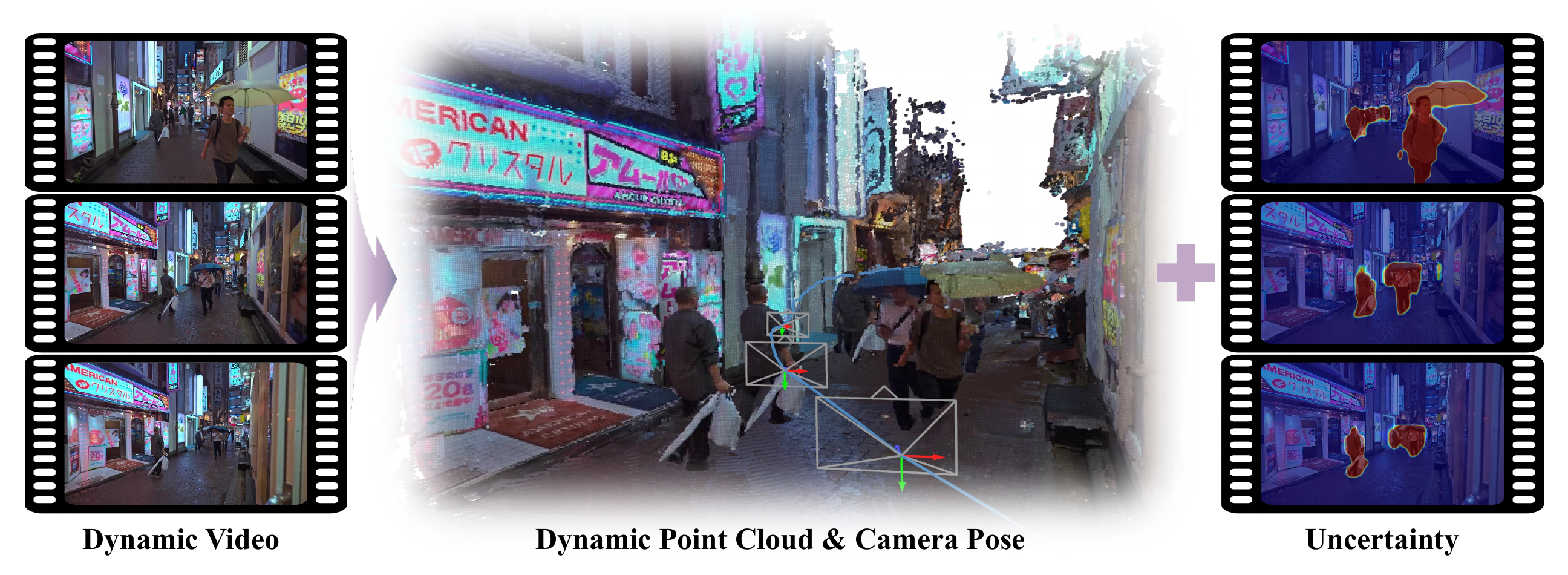}
    \vspace{-2mm}
    \captionof{figure}{\textbf{DROID-W.} Given a casually captured \textit{in-the-wild} video, our method estimates accurate dynamic uncertainty, camera trajectory, and scene structure, where existing SLAM baselines fail.
    \textit{Left}: frames of the input video. 
    \textit{Middle}: reconstructed dynamic point clouds with estimated camera poses. 
    \textit{Right}: overlay of optimized uncertainty on the corresponding input frames.}
    \label{fig:teaser}
\end{center}
}]

\renewcommand{\thefootnote}{\fnsymbol{footnote}}
\footnotetext[1]{Equal contribution.}

\begin{abstract}
We present a robust, real-time RGB SLAM system that handles dynamic environments by leveraging differentiable Uncertainty-aware Bundle Adjustment. Traditional SLAM methods typically assume static scenes, leading to tracking failures in the presence of motion. Recent dynamic SLAM approaches attempt to address this challenge using predefined dynamic priors or uncertainty-aware mapping, but they remain limited when confronted with unknown dynamic objects or highly cluttered scenes where geometric mapping becomes unreliable. In contrast, our method estimates per-pixel uncertainty by exploiting multi-view visual feature inconsistency, enabling robust tracking and reconstruction even in real-world environments. The proposed system achieves state-of-the-art camera poses and scene geometry in cluttered dynamic scenarios while running in real time at around 10 FPS. Code and datasets are available at \href{https://github.com/MoyangLi00/DROID-W.git}{https://github.com/MoyangLi00/DROID-W.git}.
\end{abstract}    
\section{Introduction}
\label{sec:intro}

Simultaneous Localization and Mapping (SLAM) is a fundamental task in computer vision, with broad applications in autonomous driving~\cite{badue2021self, geiger201kitti}, robotics~\cite{liu2024slideslam, arm2023scientific, zhou2022swarm}, and embodied intelligence~\cite{hoque2025egodex, chen2023not, krishnan2025benchmarking}. Despite remarkable progress, achieving reliable SLAM in real-world environments is challenging. Dynamic and non-rigid objects often compromise pose estimation and 3D reconstruction, limiting the robustness and applicability of SLAM systems in practice.

Although this task has been extensively studied, many existing methods~\cite{newcombe2011kinectfusion, Mur2015TRO, Engel2017DSO, Mur2017orb2, teed2021droid, teed2023dpvo} still assume a static environment and ignore non-rigid motion, which results in errors in both camera tracking and scene reconstruction. Some recent works~\cite{bescos2018dynaslam, schischka2023dynamon, xu2024dgslam, jiang2024rodyn, wu2025addslam} attempt to handle dynamic scenes by detecting or segmenting moving objects and masking out those regions. However, they rely heavily on prior knowledge of dynamic objects, which limits their robustness in complex and diverse real-world environments.

Recently, uncertainty-aware methods~\cite{ren2024nerf, kulhanek2024wildgaussians, zheng2025wildgs, zheng2025upslam} have attracted increasing attention for handling scene dynamics without relying on predefined motion priors. These approaches typically employ a shallow multi-layer perceptron (MLP) to estimate pixel-wise uncertainty from DINO~\cite{oquab2023dinov2} features and optimize the predictor through an online update. However, these approaches rely on constructing a perfectly static neural implicit~\cite{Mildenhall2020ECCV} or Gaussian Splatting~\cite{kerbl3Dgaussians} map to optimize uncertainty. Consequently, their performance remains limited in complex real-world environments, where dynamic and cluttered scenes pose significant challenges for stable scene representation.

To address these limitations, we propose \textbf{DROID-W}, a novel dynamics-aware SLAM system that adapts prior deep visual SLAM system DROID-SLAM \cite{teed2021droid} to dynamic environments. 
We incorporate uncertainty optimization into the differentiable bundle adjustment (BA) layer to iteratively update dynamic uncertainty, camera poses, and scene geometry.
The pixel-wise uncertainty of the frame is updated by leveraging multi-view visual feature similarity.
In contrast with prior approaches, our uncertainty estimation is not constrained by high-quality geometric mapping or predefined motion priors.
In addition, we introduce the DROID-W dataset, capturing diverse and unconstrained outdoor dynamic scenes, and further include \textit{YouTube} clips for truly in-the-wild evaluation. In contrast to the saturated indoor benchmarks prevalent in prior works, our sequences feature challenging real-world settings with various object dynamics.
Experimental results demonstrate that our approach achieves robust uncertainty estimation in real-world environments, leading to state-of-the-art camera tracking accuracy and scene geometry reconstruction while running in \textit{real time} at approximately 10 FPS.

\section{Related Works}
\label{sec:related_works}

\paragraph{Traditional Visual SLAM}
Many existing traditional visual SLAM methods~\cite{engel2014lsd, Mur2015TRO, Engel2017DSO, Mur2017orb2, teed2021droid, teed2023dpvo} assume a static environment, which often leads to feature mismatching and degrades both tracking accuracy and mapping quality.
To mitigate the disruption caused by object motion, some prior works~\cite{kerl2013dense, kerl2013robust} implicitly handle dynamic elements through penalizing large frame-to-frame residuals during optimization.
Other methods \cite{scona2018staticfusion, palazzolo2019iros} identify dynamic areas based on frame-to-model alignment residuals.
StaticFusion~\cite{scona2018staticfusion} employs keypoint clustering and frame-to-model alignment to detect regions with large residuals, introducing a penalization term to constrain the map to static regions.
ReFusion~\cite{palazzolo2019iros} adopts a TSDF~\cite{curless1996tsdf} representation and removes uncertain regions with large depth residuals to maintain a consistent background map.

A complementary line of approaches~\cite{runz2017cofusion, bescos2018dynaslam, yu2018ds-slam, zhong2018detect-slam, runz2018maskfusion} exploits object detection and segmentation to explicitly filter out dynamic regions.
DynaSLAM~\cite{bescos2018dynaslam} and DS-SLAM~\cite{yu2018ds-slam}, both built upon ORB-SLAM2~\cite{Mur2017orb2}, employ segmentation networks~\cite{he2017mask, badrinarayanan2017segnet} to detect moving objects and reconstruct a static background.
Detect-SLAM~\cite{zhong2018detect-slam} integrates the SSD detector~\cite{liu2016ssd} and propagates the moving probability of keypoints to reduce latency caused by object detection.
Co-Fusion~\cite{runz2017cofusion} and MaskFusion~\cite{runz2018maskfusion} extend to the object level, jointly segmenting, tracking, and reconstructing multiple independently moving objects.
FlowFusion~\cite{zhang2020flowfusion} instead leverages optical flow residuals to highlight dynamic regions.

\paragraph{NeRF- and GS-based SLAM}
Recent advances in Neural Radiance Fields (NeRF)~\cite{Mildenhall2020ECCV} have garnered substantial attention for their integration into SLAM systems, owing to their dense representation and photorealistic rendering capabilities. The pioneering work iMAP~\cite{Sucar2021ICCV} introduces the first neural implicit SLAM framework, achieving high-quality dense mapping. However, iMAP \cite{Sucar2021ICCV} suffers from the loss of fine details and catastrophic forgetting, as it represents the entire scene in a single MLP. To overcome these limitations, NICE-SLAM~\cite{Zhu2022CVPR} incorporates hierarchical feature grids to enhance scalability and reconstruction fidelity. Subsequent methods~\cite{yang2022voxfusion, Johari2022ESLAM, wang2023coslam, sandstrom2023point, zhang2023go, zhu2024nicer} further improve the efficiency and robustness of such SLAM systems.
More recently, the emergence of 3D Gaussian Splatting (3DGS)~\cite{kerbl3Dgaussians} inspired numerous SLAM approaches~\cite{keetha2024splatam, sandstrom2024splat, matsuki2024monogs, yan2024gs-slam, hu2024cg, ha2024gs-icp} that adopt Gaussian primitives. 
However, these methods typically assume predominantly static environments, which limits their applicability in real-world scenarios with dynamic objects.

\begin{figure*}
    \centering
    \vspace{-4mm}
    \includegraphics[width=1.02\linewidth]{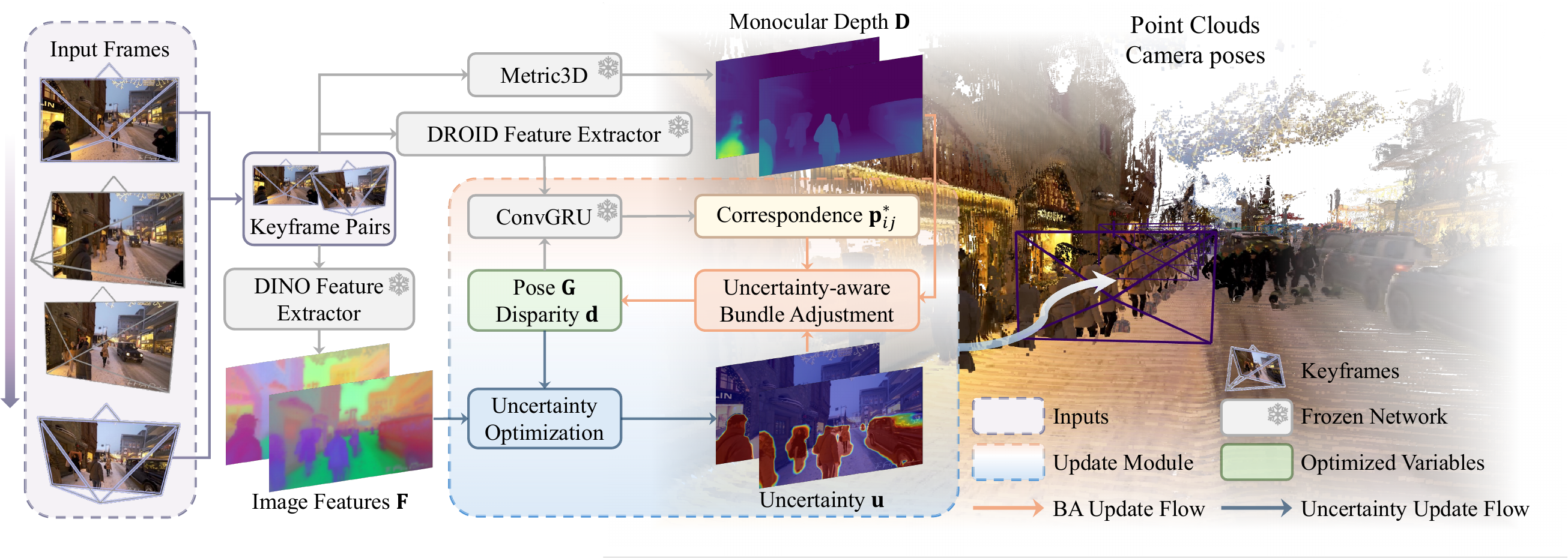}
    \vspace{-8mm}
    \caption{\textbf{System Overview.} The proposed DROID-W takes a sequence of RGB images as inputs and simultaneously estimates camera poses while recovering scene geometry. It alternatingly performs pose-depth refinement and uncertainty optimization in an iterative manner. The proposed uncertainty-aware dense bundle adjustment weights reprojection residuals with per-pixel uncertainty $\bu$ to mitigate the influence of dynamic distractors. In addition, we use predicted monocular depth ${\mathbf{D}}$ as regularization of bundle adjustment, to improve its robustness under highly dynamic environments. For the uncertainty optimization module, we first extract DINOv2~\cite{oquab2023dinov2} features from the input images and then iteratively update the dynamic uncertainty map by leveraging multi-view feature consistency. Specifically, feature consistency is measured by the cosine similarity between features of image $\bI_i$ and its corresponding features in image $\bI_j$, where the rigid-motion correspondences $\bp_{ij}$ are derived using the current pose and depth estimates.}
    \label{fig:pipeline}
    \vspace{-4mm}
\end{figure*}

To overcome this limitation, several dynamic NeRF-based~\cite{schischka2023dynamon, li2024ddn, jiang2024rodyn, wu2025dvn-slam} and GS-based SLAM systems~\cite{xu2024dgslam, zheng2025upslam, wu2025addslam, zheng2025wildgs, liu2025sdd-slam, li20254dgs-slam} have been proposed. Most of them~\cite{li2024ddn, liu2025sdd-slam, li20254dgs-slam, wu2025addslam} rely on object detection or semantic segmentation to mask out dynamic regions, but struggle to handle undefined or unseen object classes. To address this, DynaMoN~\cite{schischka2023dynamon} introduces an additional CNN to predict motion masks from forward optical flow, while RoDyn-SLAM~\cite{jiang2024rodyn} and DG-SLAM~\cite{xu2024dgslam} combine semantic segmentation with warping masks to improve motion mask estimation.
WildGS-SLAM~\cite{zheng2025wildgs} and UP-SLAM~\cite{zheng2025upslam} employ uncertainty modeling to handle scene dynamics. They utilize a shallow MLP to estimate per-pixel motion uncertainty from DINOv2~\cite{oquab2023dinov2} features, as these features are robust to appearance variations and can represent abundant semantic information. The uncertainty MLP is optimized under the supervision of photometric and depth losses between  input and rendered images. Furthermore, UP-SLAM~\cite{zheng2025upslam} extends high-dimensional visual features into the 3DGS feature space and introduces a similarity loss as additional uncertainty constraints. 

However, the optimization of uncertainty in these methods remains tightly coupled with scene representation, leading to performance degradation in complex environments where mapping struggles. 
In contrast, our approach adopts visual feature similarity between frames to estimate dynamic uncertainty, demonstrating robustness and effectiveness in challenging real-world environments.

\paragraph{Feed-forward Approaches}
Recent feed-forward reconstruction and pose estimation methods have achieved remarkable progress. DUSt3R~\cite{wang2024dust3r} and VGGT~\cite{wang2025vggt} demonstrate strong performance in scene geometry estimation. MonST3R~\cite{zhang2024monst3r} extends DUSt3R~\cite{wang2024dust3r} to dynamic environments by estimating the dynamic mask from optical flow and pointmaps. Easi3R~\cite{chen2025easi3r} introduces a training-free 4D reconstruction framework that isolates motion information from the attention maps of DUSt3R~\cite{wang2024dust3r}. However, these methods are restricted to short sequences. CUT3R~\cite{wang2025cut3r} and TTT3R~\cite{chen2025ttt3r} further advance feed-forward reconstruction by handling long sequences in an online continuous manner. 
Despite these approaches achieving visually convincing geometry estimation, purely feed-forward pipelines often struggle to recover accurate camera trajectories and metrically consistent structure compared to SLAM-style systems. In contrast, our method, grounded in a visual SLAM framework, yields more accurate camera trajectories and reconstructions.
\section{Proposed Method}
\label{sec:method}

Our approach adapts prior deep visual SLAM DROID-SLAM \cite{teed2021droid} by introducing a differentiable Uncertainty-aware Bundle Adjustment (UBA) that explicitly models per-pixel uncertainty to handle dynamic objects.
Given RGB sequences from cluttered real-world scenes, our system optimizes camera poses, depth, and uncertainty to achieve robust tracking and accurate geometry estimation.

Next, we will first summarize the key components of DROID-SLAM  designed for static environments (\secref{sec:preliminary}). We then present our proposed differentiable Uncertainty-aware Bundle Adjustment (\secref{sec:udba}) and dynamic uncertainty update (\secref{sec:uncer_opt}) modules. Finally, we introduce the proposed overall dynamic SLAM system (\secref{sec:slam_system}). The overview of \textbf{DROID-W} is shown in \figref{fig:pipeline}.

\subsection{Preliminaries}
\label{sec:preliminary}
DROID-SLAM  leverages a differentiable bundle adjustment (BA) layer to update camera poses and depths in an iterative manner. For each RGB image in the input sequence $\{\bI_t\}_{t=0}^N$, it maintains two state variables: camera pose $\mathbf{G}_t \in SE(3)$, inverse depth $\bd_t \in \mathcal{R}^{\frac{H}{8} \times \frac{W}{8}}$.
In addition, it constructs the frame-graph $(\mathcal{V}, \mathcal{E})$ to represent co-visibility across frames, where an edge $(i,j) \in \mathcal{E}$ means that the images $\mathbf{I}_i$ and $\mathbf{I}_j$ overlap. The set of camera poses $\{\mathbf{G}_t\}_{t=0}^N$ and inverse depths $\{\mathbf{d}_t\}_{t=0}^N$ are iteratively updated through the differentiable BA layer, operating on a set of image pairs $(\mathbf{I}_i, \mathbf{I}_j)$.

\paragraph{Differential Bundle Adjustment} For each pair of images $(\mathbf{I}_i, \mathbf{I}_j)$,  we can derive the rigid-motion correspondence as:
\begin{equation}
    \footnotesize
    \bp_{ij} = \Pi_c \Big(\bG_{ij}' \circ \Pi_c^{-1}(\bp_i, \bd_i') \Big),
    \label{eq:rigid_corres}
\end{equation}
where $\Pi_c$ denotes the camera projection function, and $\bG'_{ij}$ is the relative pose between frames $i$ and $j$. 
Variable 
$\bp_i \in \mathcal{R}^{\frac{H}{8} \times \frac{W}{8} \times 2}$ represents a grid of pixel coordinates in frame $i$.
DROID-SLAM  predicts the 2D dense correspondence $\mathbf{p}_{ij}^* \in \mathcal{R}^{\frac{H}{8} \times \frac{W}{8} \times 2}$ and confidence map $\bw_{ij} \in \mathcal{R}^{\frac{H}{8} \times \frac{W}{8} \times 2}$ in an iterative manner. The differentiable BA jointly refines camera poses and inverse depths by minimizing dense correspondence residuals as follows:
\begin{equation}
\footnotesize
\begin{aligned}
    \bE(\bG', \bd') &= 
    \sum_{(i,j) \in \mathcal{E}} 
    \left\| 
        \bp_{ij}^* - \bp_{ij}
    \right\|_{\boldsymbol{\Sigma}_{ij}}^2, \nonumber\\
    \boldsymbol{\Sigma}_{ij} &= 
    \mathrm{diag}\,(\bw_{ij}).
\label{eq:energy_droid}
\end{aligned}
\end{equation}
where $\|\cdot\|_{\Sigma}$ denotes Mahalanobis distance that weights the residuals according to the confidence map predicted by DROID-SLAM. The pose and disparity are optimized using the Gauss-Newton algorithm as follows:
\begin{equation}
\footnotesize
\label{eq:Schur}
    \begin{bmatrix}
        \mathbf{B} & \mathbf{E} \\
        \mathbf{E}^{\mathsf{T}} & \mathbf{C}
    \end{bmatrix}
    \begin{bmatrix}
        \Delta \boldsymbol{\xi} \\
        \Delta \mathbf{d}
    \end{bmatrix}
    =
    \begin{bmatrix}
        \mathbf{v} \\
        \mathbf{w}
    \end{bmatrix},
\end{equation}

\begin{equation}
    \footnotesize
    \begin{aligned}
        \Delta \boldsymbol{\xi} &=
        [\mathbf{B} - \mathbf{E}\mathbf{C}^{-1}\mathbf{E}^{\mathsf{T}}]^{-1}
        (\mathbf{v} - \mathbf{E}\mathbf{C}^{-1}\mathbf{w}),\\
        \Delta \mathbf{d} &=
        \mathbf{C}^{-1}(\mathbf{w} - \mathbf{E}^{\mathsf{T}}\Delta \boldsymbol{\xi}).
    \end{aligned}
    \label{eq:update}
\end{equation}
where $(\boldsymbol{\Delta \xi}, \mathbf{\Delta d})$ represents pose and disparity update. Matrix $\mathbf{C}$ is diagonal as each term in \eqnref{eq:energy_droid} depends only on a single depth value, thus it can be inverted by $\mathbf{C}^{-1} = 1 / \mathbf{C}$.

\subsection{Uncertainty-aware Bundle Adjustment}
\label{sec:udba}
Dynamic objects violate the rigid-motion assumption, yielding unreliable residuals that destabilize the BA layer of DROID-SLAM. To address this, we introduce a per-pixel dynamic uncertainty $\mathbf{u}_t \in \mathcal{R}^{\frac{H}{8} \times \frac{W}{8}}$ that downweights inconsistent correspondences during optimization. Intuitively, $\mathbf{u}_t$ acts as a confidence term penalizing high residuals caused by dynamic objects. Thus, we define uncertainty-aware Mahalanobis distance term $\| \cdot \|_{\Sigma_{ij}^{\text{uncer}}}$ as follows:
\begin{equation}
    \footnotesize
    \boldsymbol{\Sigma}_{ij}^{\text{uncer}} = 
    \mathrm{diag}\,(\bw_{ij} \cdot \frac{1}{\mathbf{{u}}'_i}).
\label{eq:mahalanobis_droid_w}
\end{equation}
However, jointly optimizing pose, depth, and uncertainty via Gauss-Newton algorithms is computationally prohibitive. We thus adopt an interleaved optimization strategy that alternates between pose-depth refinement and uncertainty optimization. The pose-depth refinement is performed by minimizing the following uncertainty-aware energy function:
\begin{equation}
    \footnotesize
    \hat{\bE}(\bG', \bd') = 
    \sum_{(i,j) \in \mathcal{E}} 
    \left\| 
        \bp_{ij}^* - \bp_{ij}
    \right\|_{\boldsymbol{\Sigma}^{\text{uncer}}_{ij}}^2.
\label{eq:energy_droid_w}
\end{equation}
%

\subsection{Uncertainty Optimization}
\label{sec:uncer_opt}
For the optimization of dynamic uncertainty, we measure multi-view inconsistency via the similarity of DINOv2~\cite{oquab2023dinov2} features across image pairs rather than the reprojection residuals in \eqnref{eq:energy_droid_w}. Reprojection error can become unreliable under large dynamic motion, while 2D visual feature similarity yields a more stable and semantically meaningful measure for multi-view inconsistency.

\paragraph{Uncertainty Cost Function} For each pair of images $(\bI_i, \bI_j)$, 2D visual features $(\bF_i, \bF_j)$ are first extracted using FiT3D~\cite{yue2025FiT3D}, a refined DINOv2 model.
For each pixel \(\bp_i\) in frame $i$, we compute its rigid-motion correspondence \(\bp_{ij}\) in frame $j$ via \eqnref{eq:rigid_corres}. We then obtain corresponding feature \(\bF_{ij}\) and uncertainty \(\bu_{ij}\) through bilinear interpolation. 
Multi-view consistency of the image pair is measured by cosine similarity between the DINOv2 features $(\mathbf{F}_i, \mathbf{F}_{ij})$. The dynamic objects in the environment with multi-view inconsistency are expected to have high uncertainty. Thus, we formulate the following similarity loss:
\begin{equation}
    \footnotesize
    \mathbf{E}_{\text{sim}}(\mathbf{u}{'}) = 
    \sum_{(i,j) \in \mathcal{E}}
        \frac{1 - \frac{\bF_i \cdot {\bF}_{ij}}{ {\|\bF_i\|}_2 {\|{\bF}_{ij}\|}_2 }}{\bu'_i \cdot {\bu}'_{ij}}.
    \label{eq:sim_loss}
\end{equation}
Here, we optimize bidirectional uncertainties for each image pair to decouple inter-frame dynamics. 

To avoid the trivial solution of $\mathbf{u'} \rightarrow + \infty$, we regularize the uncertainty with a logarithmic prior:
\begin{equation}
    \footnotesize
    \mathbf{E}_{\text{prior}}(\mathbf{u'}) = \sum_{i} \mathrm{log}(\bu'_i+1.0). 
    \label{eq:prior_loss}
\end{equation}
Here, we add a bias term 1.0 to the uncertainty to prevent the prior loss from being negative.


Thus, the total uncertainty cost function is defined as:
\begin{equation}
    \footnotesize
    \mathbf{E}_{\text{uncer}}(\bu') = \mathbf{E}_{\text{sim}}(\bu') + \gamma_{\text{prior}} \mathbf{E}_{\text{prior}}(\bu').
    \label{eq:uncer_loss}
\end{equation}
\paragraph{Uncertainty Regularization} 
Direct optimization of pixel-wise uncertainty may suffer from spatial inconsistency and overfitting to noise due to various dynamic motion.
To address this, we learn a local affine mapping followed by the Softplus activation function from DINOv2 features to uncertainties.
Thus, the uncertainty is obtained via $\mathbf{u} = \text{Softplus}(\boldsymbol{\theta} \cdot \mathbf{F})$.
This affine mapping plays the role of a regularization term within the small local window, which is different from the decoder in prior works \cite{zheng2025wildgs, ren2024nerf}.

\paragraph{Optimization} To avoid the inverse computation of the large Hessian matrix, we optimize uncertainty using Gradient Descent with weight decay instead of the Newton algorithm.
All backpropagation operations are implemented in CUDA to ensure efficiency.
The learnable parameters $\boldsymbol{\theta}$ of the affine mapping layer are updated as the following Jacobians:
\begin{equation}
\footnotesize
\begin{aligned}
    \boldsymbol{g}_t
    & = \sum_{i=0}^{N}
    \frac{\partial \mathbf{E}_{\text{uncer}}}{\partial \mathbf{u}'_i}
    \cdot \frac{\partial \mathbf{u}'_i}{\partial \boldsymbol{\theta}_{t-1}} \\
    & = \sum_{i=0}^{N}
    \frac{\partial \mathbf{E}_{\text{uncer}}}{\partial \mathbf{u}'_i}
    \cdot \frac{1}{1+\text{exp}(-{\boldsymbol{\theta}_{t-1} \cdot \mathbf{F}_i})}
    \cdot \mathbf{F}_i, \\
    \boldsymbol\theta_t & = \boldsymbol{\theta}_{t-1} - \lambda \cdot \boldsymbol{g}_{t} - \eta \cdot \boldsymbol{\theta}_{t-1}.
\label{eq:jacob_mlp}
\end{aligned}
\end{equation}
For more details about the gradient derivations, please refer to the supplementary material.

\subsection{SLAM System}
\label{sec:slam_system}

Following DROID-SLAM, we accumulate 12 keyframes with sufficient motion to initialize the SLAM system. 
DROID-SLAM initializes the disparities as the constant value of 1, which can cause inaccurate tracking in high-dynamic scenes. 
Thus, we adopt the metric monodepth $\bD_t \in \mathcal{R}^{\frac{H}{8} \times \frac{W}{8}}$ predicted by Metric3D \cite{hu2024metric3d} to penalize the disparity and improve accuracy. Thus, the cost function of BA with depth regularization is defined as follows:
\begin{equation*}
    \footnotesize
    \bE^{+}(\bG', \bd') = 
    \sum_{(i,j) \in \mathcal{E}} 
    \left\| 
        \bp_{ij}^* - \bp_{ij}
        \right\|_{\boldsymbol{\Sigma}^{\text{uncer}}_{ij}}^2 + \gamma_d
        \sum_i
    \left\| \bd_i' - \bD_i \right\|^2.
\label{eq:energy_w_d_reg}
\end{equation*}
After the initialization, we process incoming keyframes in an incremental manner. For newly added keyframes, we follow DROID-SLAM to perform local bundle adjustment in a sliding window and adopt depth regularization.
For both initialization and frontend tracking stages, we optimize poses, disparities, and uncertainties. 
After frontend tracking, we perform global BA over all keyframes to refine camera poses and disparities. We freeze the dynamic-uncertainty parameters during global BA, since the affine transformation is intended to regularize uncertainty locally within the sliding window rather than at global scale.

\section{Experiments}
\label{sec:exp}

\begin{figure*}[ht]
  \vspace{-1mm}
  \centering
  \scriptsize
  \setlength{\tabcolsep}{1.1pt}
  \newcommand{\sz}{0.132}
  \newcommand{\sza}{0.115} 
  \begin{tabular}{ccccccc}
  \raisebox{3.5\normalbaselineskip}[0pt][0pt]{\rotatebox[origin=c]{90}{Input}} &
  \includegraphics[height=\sza\linewidth]{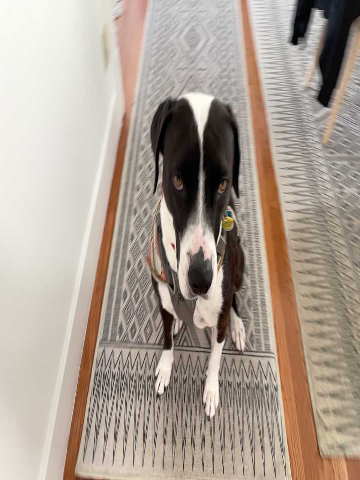}  &
  \includegraphics[height=\sza\linewidth]{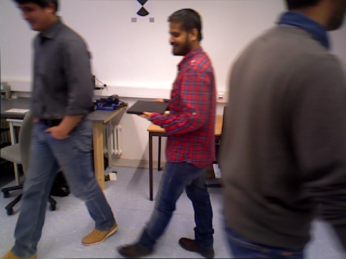} &
  \includegraphics[height=\sza\linewidth]{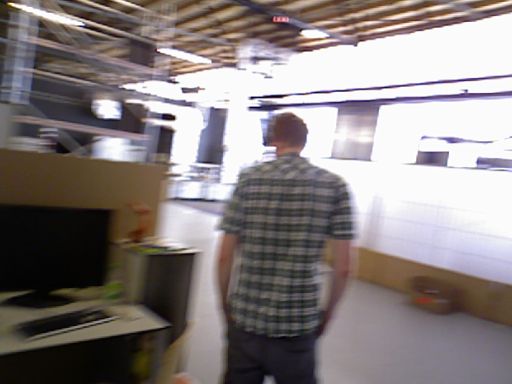} &
  \includegraphics[height=\sza\linewidth]{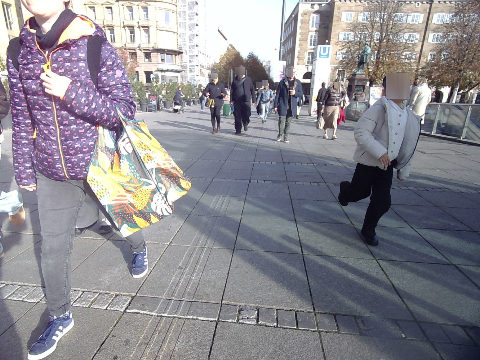}  &
  \includegraphics[height=\sza\linewidth]{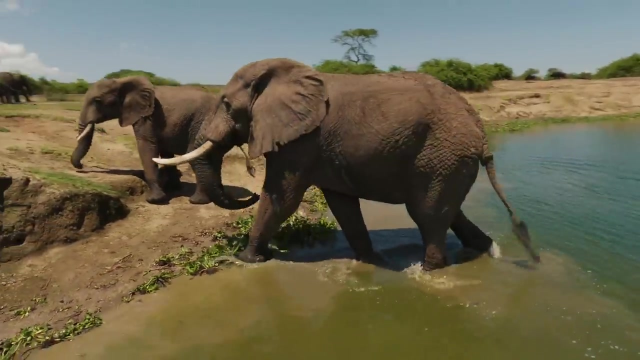}  &
  \includegraphics[height=\sza\linewidth]{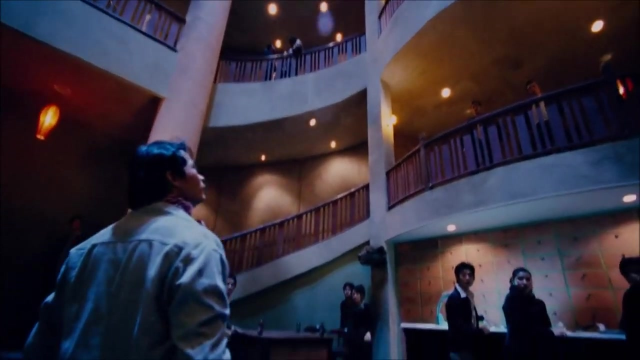}  
  \\[-0.1mm]

  \raisebox{3.5\normalbaselineskip}[0pt][0pt]{\rotatebox[origin=c]{90}{MonST3R \cite{zhang2024monst3r}}}   & \includegraphics[height=\sza\linewidth]{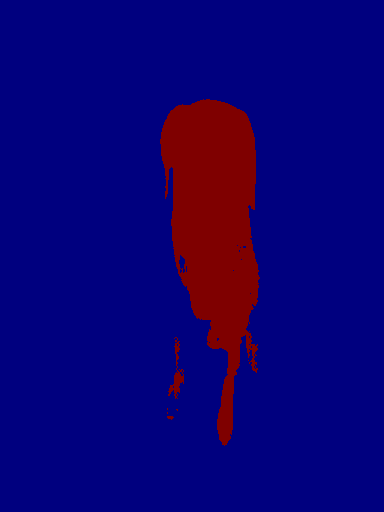} &
  \includegraphics[height=\sza\linewidth]{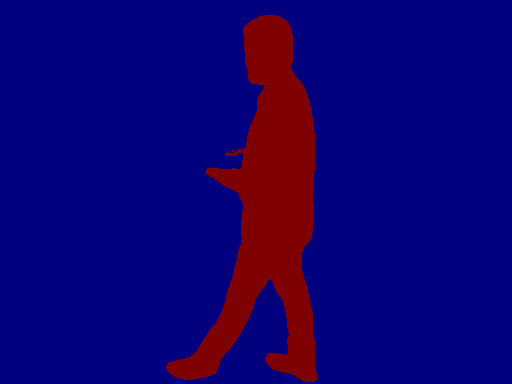} &
  \includegraphics[height=\sza\linewidth]{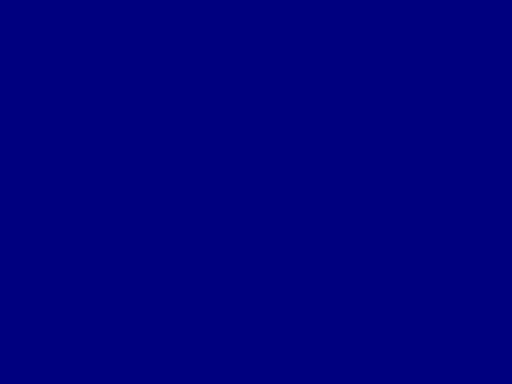} &
  \includegraphics[height=\sza\linewidth]{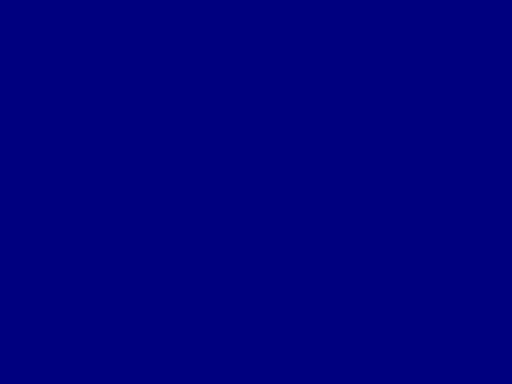} &
  \includegraphics[height=\sza\linewidth]{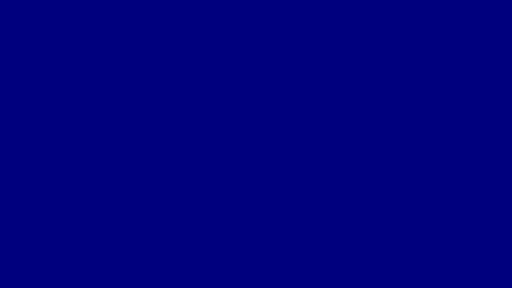} &
  \includegraphics[height=\sza\linewidth]{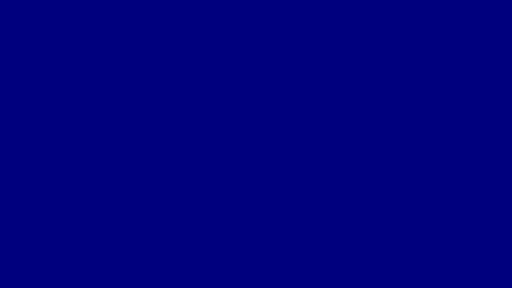}
  \\[-0.1mm]
  
  \raisebox{3.5\normalbaselineskip}[0pt][0pt]{\rotatebox[origin=c]{90}{WildGS-SLAM \cite{zheng2025wildgs}}}  &
  \includegraphics[height=\sza\linewidth]{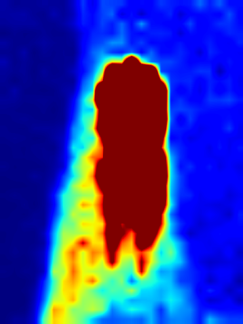} &
  \includegraphics[height=\sza\linewidth]{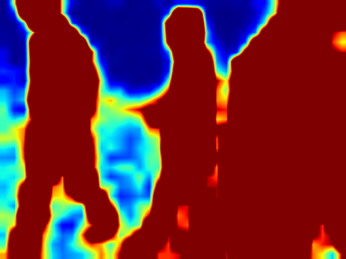} &
  \includegraphics[height=\sza\linewidth]{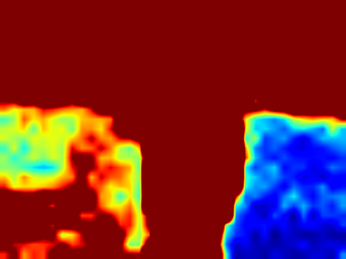} &
  \includegraphics[height=\sza\linewidth]{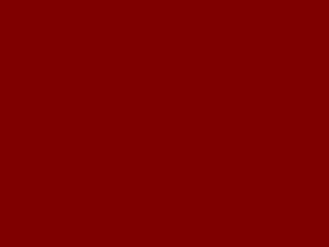}  &
  \includegraphics[height=\sza\linewidth]{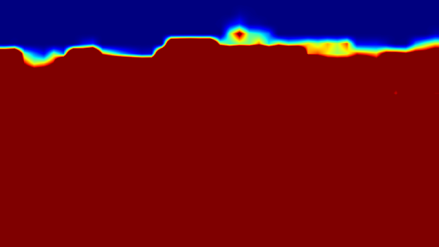}  &
  \includegraphics[height=\sza\linewidth]{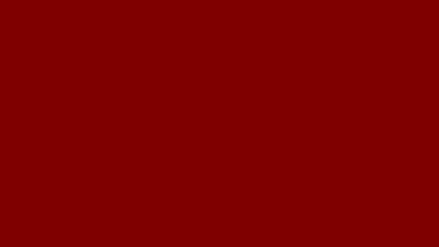}
  \\[-0.1mm]

  \raisebox{3.5\normalbaselineskip}[0pt][0pt]{\rotatebox[origin=c]{90}{\textbf{Ours}}}  &
  \includegraphics[height=\sza\linewidth]{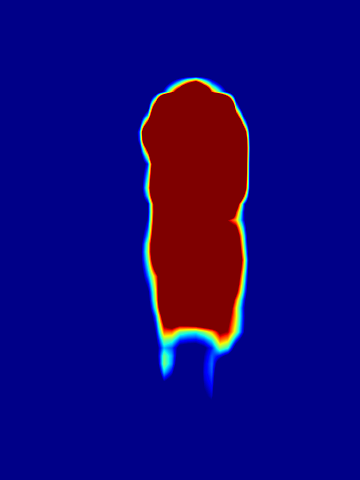} &
  \includegraphics[height=\sza\linewidth]{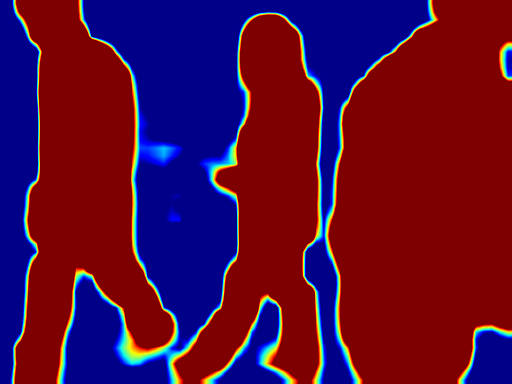} &
  \includegraphics[height=\sza\linewidth]{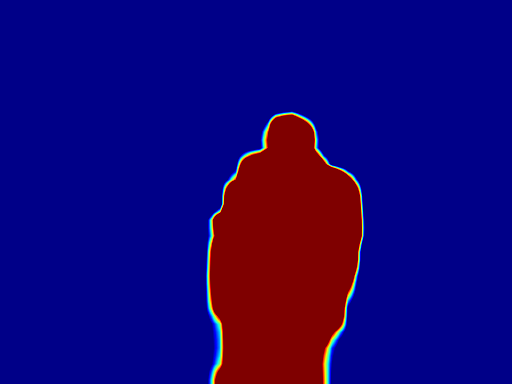} &
  \includegraphics[height=\sza\linewidth]{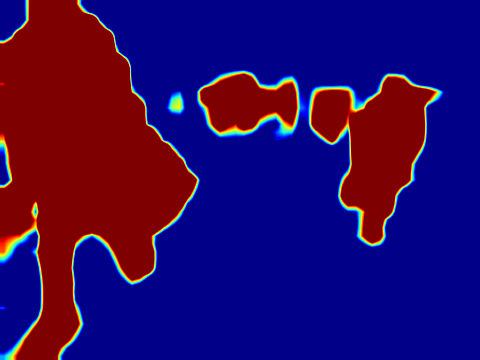} &
  \includegraphics[height=\sza\linewidth]{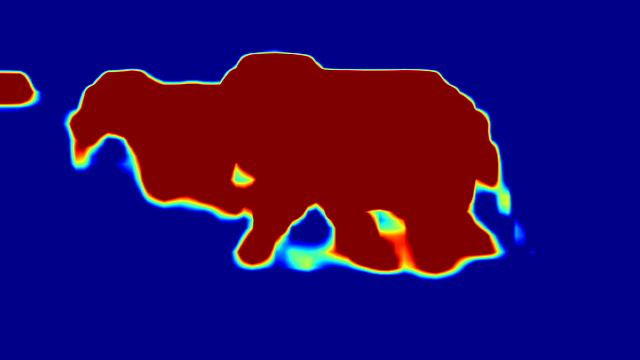}  &
  \includegraphics[height=\sza\linewidth]{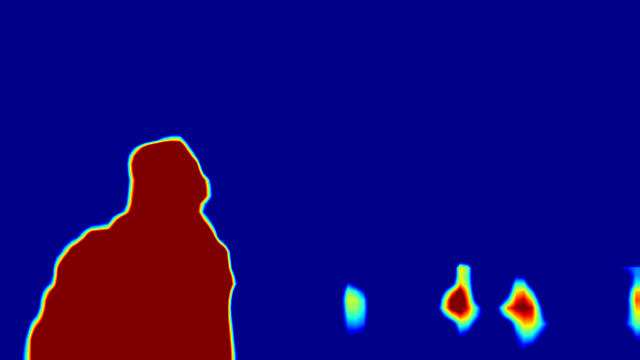} 
  \\
  
  & {\fontsize{7}{8} \selectfont \texttt{DyCheck Haru}}
  & {\fontsize{7}{8} \selectfont \texttt{Bonn Crowd}} 
  & {\fontsize{7}{8} \selectfont \texttt{TUM f3/wr}} 
  & {\fontsize{7}{8} \selectfont \texttt{DROID-W Downtown 5}} 
  & {\fontsize{7}{8} \selectfont \texttt{YouTube Elephant Herd}}  
  & {\fontsize{7}{8} \selectfont \texttt{YouTube Tomyum 1}}
  \\
     
  \end{tabular}
  \vspace{-3mm}
    \caption{\textbf{Uncertainty Estimation.} WildGS-SLAM~\cite{zheng2025wildgs} and our approach estimate dynamic uncertainty, whereas MonST3R~\cite{zhang2024monst3r} predicts a binary motion mask. Our approach produces more accurate and spatially consistent uncertainty estimations across all challenging sequences.}
    \vspace{-1mm}
  \label{fig:uncer_comparisons}
\end{figure*} 


\paragraph{Datasets} 
We evaluate our approach on the Bonn RGB-D Dynamic dataset~\cite{palazzolo2019iros}, TUM RGB-D dataset \cite{sturm2012benchmark}, and DyCheck \cite{gao2022dycheck} dataset.
To further assess performance in \textit{\textbf{unconstrained, outdoor}} settings, we introduce the {DROID-W} dataset, captured using a Livox Mid-360 LiDAR rigidly mounted with an RGB camera. The dataset comprises 7 sequences (\texttt{Downtown 1–7}) with RGB frames at a resolution of $1200{\times}1600$, ground-truth camera poses, and synchronized IMU and LiDAR measurements.
Since satellite-based localization is unavailable for \texttt{Downtown 1–2}, we use FAST-LIVO2~\cite{zheng2024fast_livo2} trajectories as ground truth, whereas the remaining sequences rely on RTK ground truth.

Additionally, we test on 6 dynamic videos downloaded from \textit{{YouTube}}. The sequences span 8 seconds to 30 minutes, featuring diverse object motion and cluttered scenes. 
Sequences exceeding 5 minutes are partitioned into non-overlapping 5-minute segments due to resource bottlenecks of SLAM on a single GPU. For each video, the camera intrinsics are estimated with MonST3R \cite{zhang2024monst3r} using 20 frames.

\begin{table*}[t]
\vspace{-1mm}
\centering
\footnotesize
\setlength{\tabcolsep}{5.0pt}
{
\begin{tabular}{lrrrrrrrr
!{\smash{\tikz[baseline]{\draw[densely dashed, gray!80, line width=0.8pt] (0pt,-2pt)--(0pt,8pt);}}}
r}
\toprule
Method & \texttt{Balloon} & \texttt{Balloon2} & \texttt{Crowd} & \texttt{Crowd2} & \texttt{Person} & \texttt{Person2} & \texttt{Moving} & \texttt{Moving2} & \textbf{Avg.}\\
\midrule
\multicolumn{10}{l}{\cellcolor[HTML]{EEEEEE}{\textit{RGB-D}}} \\ 
NICE-SLAM~\cite{Zhu2022CVPR} & 24.4 & 20.2 & 19.3 & 35.8 & 24.5 & 53.6 & 17.7 & 8.3 & 22.74\\

ReFusion~\cite{palazzolo2019iros} & 17.5 & 25.4 & 20.4 & 15.5 & 28.9 & 46.3 & 7.1 & 17.9 & 22.38\\

RoDyn-SLAM~\cite{jiang2024rodyn} & 7.9 & 11.5 & - & - & 14.5 & 13.8 & - & 12.3 & N/A \\

DynaSLAM (N+G)~\cite{bescos2018dynaslam}& 3.0 & 2.9 & \nd 1.6 & 3.1 & 6.1 & 7.8 & 23.2 & 3.9 & 6.45\\

ORB-SLAM2~\cite{Mur2017orb2} & 6.5 & 23.0 & 4.9 & 9.8 & 6.9 & 7.9 & 3.2 & 3.9 & 6.36\\

DG-SLAM~\cite{xu2024dgslam} & 3.7 & 4.1 & - & - & 4.5 & 6.9 & - & 3.5 & N/A  \\

DDN-SLAM (RGB-D)~\cite{li2024ddn} & \fs 1.8 & 4.1 & \rd 1.8 & \rd 2.3 & 4.3 & 3.8 & 2.0 & 3.2 & 2.91\\
UP-SLAM~\cite{zheng2025upslam} & 2.8 & 2.7 & - & - & 4.0 & 3.6 & - & 3.2 & N/A  \\
ADD-SLAM~\cite{wu2025addslam} & \rd 2.7 & \fs 2.3 & - & - & \fs 2.4 & 3.7 & - & \fs 2.1 & N/A  \\

\hdashline
\noalign{\vskip 1pt}
\multicolumn{10}{l}{\cellcolor[HTML]{EEEEEE}{\textit{RGB-only}}} \\ 
Splat-SLAM~\cite{sandstrom2024splat}& 8.8 & 3.0 & 6.8 & F & 4.9 & 25.8 & 1.7 & 3.0 & N/A\\


TTT3R~\cite{chen2025ttt3r} & 21.5 & 15.4 & 9.8 & 7.7 & 30.0 & 21.4 & 33.4 & 41.2 & 22.55 \\

DSO~\cite{Engel2017DSO}& 7.3 & 21.8 & 10.1 & 7.6 & 30.6 & 26.5 & 4.7 & 11.2 & 14.98\\

MonST3R~\cite{zhang2024monst3r} & 7.2 & 6.0 & 6.6 & 6.9 & 9.8 & 16.1 & 3.5 & 6.7 & 7.85 \\

DROID-SLAM~\cite{teed2021droid} & 7.5 & 4.1 & 5.2 & 6.5 & 4.3 & 5.4 & 2.3 & 4.0 & 4.91\\

DynaMoN (MS\&SS)~\cite{schischka2023dynamon} & 
2.8 & 2.7 & 3.5 & 2.8 & 14.8 & \fs 2.2 & \fs 1.3 & 2.7 & 4.10\\

DynaMoN (MS)~\cite{schischka2023dynamon} & 6.8 & 3.8 & 6.1&5.6& \fs 2.4 & 3.5 & \nd 1.4 & 2.6 & \rd 4.02\\

WildGS-SLAM~\cite{zheng2025wildgs} & 2.8 & \nd 2.4 & \nd 1.6 & \nd 2.2 & \rd 3.9 & \rd 3.1 & 1.7 & \rd 2.5 & \nd 2.52 \\

{\project{}} & \nd 2.6 & \rd 2.5 & \fs 1.3 & \fs 1.8 & \nd 3.3 & \nd 2.9 & \rd 1.6 & \nd 2.3 & \fs 2.30 \\
    
\bottomrule
\end{tabular}
}
\vspace{-2mm}
\caption{\textbf{Tracking Performance on the Bonn RGB-D Dynamic Dataset~\cite{palazzolo2019iros}} (ATE RMSE $\downarrow$ [cm]). 
Best results are highlighted as \colorbox{colorFst}{\textbf{first}}, \colorbox{colorSnd}{second}, and \colorbox{colorTrd}{third}.  
“-” indicates sequences without reported results in the original papers or unavailable code. “F” denotes tracking failure. 
For MonST3R~\cite{zhang2024monst3r}, we use the same keyframes as our method and perform evaluation in a window-wise manner with a window size of 20 and an overlap ratio of 0.5 to reduce memory consumption.}
\label{tab:bonn_tracking}
\vspace{-0mm}
\end{table*}

\begin{table*}[t]
\vspace{-2mm}
\centering
\footnotesize
{
\setlength{\tabcolsep}{5.5pt}
\begin{tabular}{lrrrrrrrrr
!{\smash{\tikz[baseline]{\draw[densely dashed, gray!80, line width=0.8pt] (0pt,-2pt)--(0pt,8pt);}}}
r}
\toprule

Method & \texttt{f2/dp} & \texttt{f3/ss} & \texttt{f3/sx} & \texttt{f3/sr} & \texttt{f3/shs} & \texttt{f3/ws} & \texttt{f3/wx} & \texttt{f3/wr} & \texttt{f3/whs} & \textbf{Avg.}\\
\midrule
\multicolumn{11}{l}{\cellcolor[HTML]{EEEEEE}{\textit{RGB-D}}} \\ 


NICE-SLAM~\cite{Zhu2022CVPR} & 88.8 & 1.6 & 32.0 & 59.1 & 8.6 & 79.8 & 86.5 & 244.0 & 152.0 & 83.60 \\
ORB-SLAM2~\cite{Mur2017orb2}& \fs 0.6 &  \nd 0.8 &  \rd 1.0 & 2.5 & 2.5 & 40.8 & 72.2 & 80.5 & 72.3 & 30.36 \\
ReFusion~\cite{palazzolo2019iros}& 4.9 &  \rd 0.9 &4.0 & 13.2 & 11.0 & 1.7 & 9.9 & 40.6 & 10.4 & 10.73 \\
DynaSLAM (N+G)~\cite{bescos2018dynaslam}&  \nd 0.7 & \fs 0.5 &1.5 & 2.7 & 1.7 & \rd 0.6 & 1.5 & \rd 3.5 & 2.5& 1.69 \\
DG-SLAM~\cite{xu2024dgslam} & 3.2 & -& \rd 1.0 & - & - & \rd 0.6 & 1.6 & 4.3 & - & N/A \\
RoDyn-SLAM~\cite{jiang2024rodyn}& -&-&-&-& 4.4 & 1.7&8.3&-&5.6&N/A\\
DDN-SLAM (RGB-D)~\cite{li2024ddn}& - & -& \rd 1.0 &- & 1.7 & 1.0 & \rd 1.4 & 3.9 & 2.3 &N/A\\
UP-SLAM \cite{zheng2025upslam} & 1.3 & - & \nd 0.9 & - & - & 0.7 & 1.6 & - & 2.6 & N/A \\
ADD-SLAM \cite{wu2025addslam} & - & - & - & - & \fs 1.3 & \nd 0.5 & \nd 1.4 & - & \fs 1.6 & N/A \\

\hdashline
\noalign{\vskip 1pt}
\multicolumn{11}{l}{\cellcolor[HTML]{EEEEEE}{\textit{RGB-only}}} \\


TTT3R~\cite{chen2025ttt3r} & 113.1 & 3.1 & 5.8 & 6.4 & 24.9 & 2.0 & 24.7 & 15.9 & 23.1 & 24.33 \\

MonST3R~\cite{zhang2024monst3r} & 33.9 & \nd 0.8 & 28.3 & 5.1 & 36.8 & 1.6 & 19.1 & 16.6 & 32.8 & 19.45 \\

DSO~\cite{Engel2017DSO}& 2.2 & 1.7 & 11.5 & 3.7 & 12.4 & 1.5 & 12.9 & 13.8 & 40.7 & 11.15\\

DDN-SLAM (RGB)~\cite{li2024ddn}& - & -&1.3 &-&3.1&2.5&2.8&8.9&4.1&N/A\\

Splat-SLAM~\cite{sandstrom2024splat}& \nd  0.7 & \fs 0.5 & \nd 0.9 & \rd 2.3 & \rd 1.5 &  2.3 & \nd 1.3 & 3.9 & 2.2 & 1.71\\

DynaMoN (MS)~\cite{schischka2023dynamon}   & \fs 0.6 & \fs 0.5 &  \nd  0.9 & \fs 2.1 & 1.9 & 1.4 &  \rd 1.4 & 3.9 & \nd 2.0 & 1.63 \\
DynaMoN (MS\&SS) ~\cite{schischka2023dynamon} & \nd 0.7 & \fs 0.5 &  \nd  0.9 & 2.4 & 2.3 & 0.7 &  \rd 1.4 & 3.9 & \rd 1.9 & 1.63 \\

DROID-SLAM~\cite{teed2021droid} & \fs 0.6 & \fs 0.5 &  \nd  0.9 & \nd 2.2 & \nd 1.4 & 1.2 & 1.6 & 4.0 & 2.2 & \nd 1.62\\

WildGS-SLAM~\cite{zheng2025wildgs} & 1.4 & \fs 0.5 & \fs 0.8 & 2.4 & 2.0 & \fs 0.4 & \nd 1.3 & \nd 3.3  & \fs  1.6 & \nd 1.51 \\ 

{\project{}} & \rd 1.1 & \fs 0.5 & \fs 0.8 & \fs 2.1 & \nd 1.4 & \nd 0.5 & \fs 1.2 & \fs 3.1 & \fs 1.6 & \fs 1.36 \\

\bottomrule
\end{tabular}
}
\vspace{-2mm}
\caption{\textbf{Tracking Performance on TUM RGB-D Dataset~\cite{sturm2012benchmark}} (ATE RMSE $\downarrow$ [cm]). 
Best results are highlighted as \colorbox{colorFst}{\bf first},\colorbox{colorSnd}{second}, and\colorbox{colorTrd}{third}. “-” indicates sequences without reported results in the original papers or unavailable code. 
Our approach consistently leads to the best or second-best results on the sequences, on average, outperforming \textit{all} baselines.
} 
\vspace{-2mm}
\label{tab:tum_tracking}
\end{table*}

\begin{table*}[t]
\vspace{-1mm}
\centering
\footnotesize
{
\setlength{\tabcolsep}{5.0pt}
\begin{tabular}{lrrrrrrrrrrrr
!{\smash{\tikz[baseline]{\draw[densely dashed, gray!80, line width=0.8pt] (0pt,-2pt)--(0pt,8pt);}}}
r}
\toprule

Method & \texttt{apple} & \texttt{backpack} & \texttt{block} & \texttt{creeper} & \texttt{handwavy} & \texttt{haru} & \texttt{mochi} & \texttt{paper} & \texttt{pillow} & \texttt{spin} & \texttt{sriracha} & \texttt{teddy} & \textbf{Avg.}\\
\midrule
\multicolumn{14}{l}{\cellcolor[HTML]{EEEEEE}{\textit{RGB-D}}} \\ 

NICE-SLAM~\cite{Zhu2022CVPR} & 0.186 & 0.149 & 0.099 & 0.166 & 0.059 & F & 0.042 & 0.062 & \nd0.171 & 0.211 & 0.073 & 0.060 & N/A \\

DynaSLAM (N+G)~\cite{bescos2018dynaslam} & 0.981 & 0.045 & 0.731 & 1.709 & 0.796 & 0.322 & 1.263 & F & 0.713 & 0.322 & 1.098 & 0.296 & N/A \\


\hdashline
\noalign{\vskip 1pt}
\multicolumn{14}{l}{\cellcolor[HTML]{EEEEEE}{\textit{RGB-only}}} \\
MonST3R~\cite{zhang2024monst3r} & 1.236 & 0.013 & 1.141 & 0.324 & 0.125 & 0.263 & 0.089 & 0.037 & 1.118 & 0.118 & 0.060 & 0.279 & 0.400 \\

TTT3R~\cite{zheng2025wildgs} & 0.915 & 0.026 & 0.507 & 0.699 & 0.298 & \nd 0.042 & 0.233 & 0.070 & 0.712 & 0.353 & 0.215 & 0.125 & 0.350 \\

Splat-SLAM~\cite{sandstrom2024splat} & \nd 0.038 & \fs 0.005 & \rd 0.078 & 0.052 & \rd 0.024 & \rd 0.078 & 0.211 & \fs 0.011 &  0.262 & \nd 0.007 & \nd 0.005 & \rd 0.048 & 0.068 \\

WildGS-SLAM~\cite{chen2025ttt3r} & 0.043 & \nd 0.006 & \nd 0.047 & \rd 0.029 & \nd 0.016 & 0.085 & \nd 0.017 & \rd 0.013 & 0.378 & \fs 0.005 & \nd 0.005 & \nd 0.027 & \rd 0.056 \\

DROID-SLAM~\cite{teed2021droid} & \fs 0.036 & \nd 0.008 & 0.156 & \fs 0.015 & \nd 0.016 & \fs 0.005 & \rd 0.018 & \fs 0.011 & \rd 0.179 & \rd 0.009 & \rd 0.010 & 0.061 & \nd 0.044 \\

{\project{}} & \rd 0.043 & \fs 0.005 & \fs 0.037 & \rd 0.017 & \fs 0.015 & 0.093 & \fs 0.010 & \nd 0.012 & \fs 0.145 & \fs 0.005 & \fs 0.004 & \fs 0.019 & \fs 0.034 \\

\bottomrule
\end{tabular}
}
\vspace{-2mm}
\caption{\textbf{Tracking Performance on DyCheck Dataset \cite{gao2022dycheck}} (ATE RMSE $\downarrow$). Best results are highlighted as \colorbox{colorFst}{\bf first},\colorbox{colorSnd}{second}, and\colorbox{colorTrd}{third}. “-” indicates sequences without reported results in the original papers or unavailable code. “F” means tracking failure. Our approach demonstrates the effectiveness and robustness in highly-textured, diverse environments, where prior methods relying on object segmentation or Gaussian mapping for uncertainty optimization often fail.
} 
\label{tab:dycheck_tracking}
\end{table*}

\begin{table*}[!ht]
\vspace{-2mm}
\centering
\footnotesize
{
\setlength{\tabcolsep}{8.7pt}
\begin{tabular}{lrrrrrrr
!{\smash{\tikz[baseline]{\draw[densely dashed, gray!80, line width=0.8pt] (0pt,-2pt)--(0pt,8pt);}}}
r}
\toprule

Method & \texttt{Downtown 1} & \texttt{Downtown 2} & \texttt{Downtown 3} & \texttt{Downtown 4} & \texttt{Downtown 5} & \texttt{Downtown 6} & \texttt{Downtown 7} & \textbf{Avg.}\\
\midrule
TTT3R~\cite{chen2025ttt3r} & 4.64 & 11.25 & 4.30 & 7.35 & 11.28 & 5.09 & 7.26 & 7.309 \\
Splat-SLAM~\cite{sandstrom2024splat} & \fs 0.10 & 6.44 & 0.89 & 0.66 & 0.91 & 2.11 & \nd 0.07 & 1.597 \\
DROID-SLAM~\cite{teed2021droid} & 0.26 & 7.84 & 1.05 & \nd 0.33 & 0.64 & \fs 0.06 & \fs 0.05 & 1.460 \\


WildGS-SLAM~\cite{chen2025ttt3r} & \fs 0.10 & \nd 0.95 & \nd 0.43 & 0.36 & \nd 0.87 & 1.22 & 0.53 & \nd 0.637 \\

{\project{}} & \nd 0.15 & \fs 0.25 & \fs 0.15 & \fs 0.32 & \fs 0.24 & \nd 0.43 & \nd 0.07 & \fs 0.230 \\

\bottomrule
\end{tabular}
}
\vspace{-2mm}
\caption{\textbf{Tracking Performance on DROID-W Dataset} (ATE RMSE $\downarrow$ [m]). Best results are highlighted as \colorbox{colorFst}{\bf first}, \colorbox{colorSnd}{second}.
} 
\vspace{-2mm}
\label{tab:stuttgart_tracking}
\end{table*}

\begin{table}[t]
\vspace{-2mm}
\centering
\footnotesize
\setlength{\tabcolsep}{3pt}
{
    \begin{tabular}{l
    !{\smash{\tikz[baseline]{\draw[densely dashed, gray!80, line width=0.8pt] (0pt,-2pt)--(0pt,8pt);}}}
    c
    !{\smash{\tikz[baseline]{\draw[densely dashed, gray!80, line width=0.8pt] (0pt,-2pt)--(0pt,8pt);}}}
    cccc}
        \toprule
        Method & Dynamic & Bonn \cite{palazzolo2019iros} & TUM \cite{sturm2012benchmark} & DyCheck \cite{gao2022dycheck}\\
        \midrule
        DROID-SLAM~\cite{teed2021droid} & \xmark & \textbf{19.89} & \textbf{26.97} & \textbf{17.50} \\
        WildGS-SLAM~\cite{zheng2025wildgs} & \cmark & \phantom{0}0.22 & \phantom{0}0.32 & \phantom{0}0.18 \\
        \project{} & \cmark & {\underline{10.57}} & \underline{14.92} & {\underline{11.06}} \\
        \bottomrule
    \end{tabular}
}
\vspace{-2mm}
\caption{\textbf{Runtime Comparisons} (average FPS $\uparrow$). 
All evaluations are conducted on an RTX~3090 GPU with a 16-core CPU.}
\label{tab:runtime}
\end{table}


\newcolumntype{d}{!{\smash{\tikz[baseline]{\draw[densely dashed, gray!70, line width=1.0pt] (0pt,-3pt)--(0pt,80pt);}}}}

\begin{figure*}[!ht]
  \vspace{-3mm}
  \centering
  \scriptsize
  \setlength{\tabcolsep}{2.0pt}
  \newcommand{\sz}{0.23} 

  \begin{tabular}{c c : c : c : c}

  \raisebox{4.0\normalbaselineskip}[0pt][0pt]{\rotatebox[origin=c]{90}{Inputs}} &
  \includegraphics[width=\sz\linewidth]{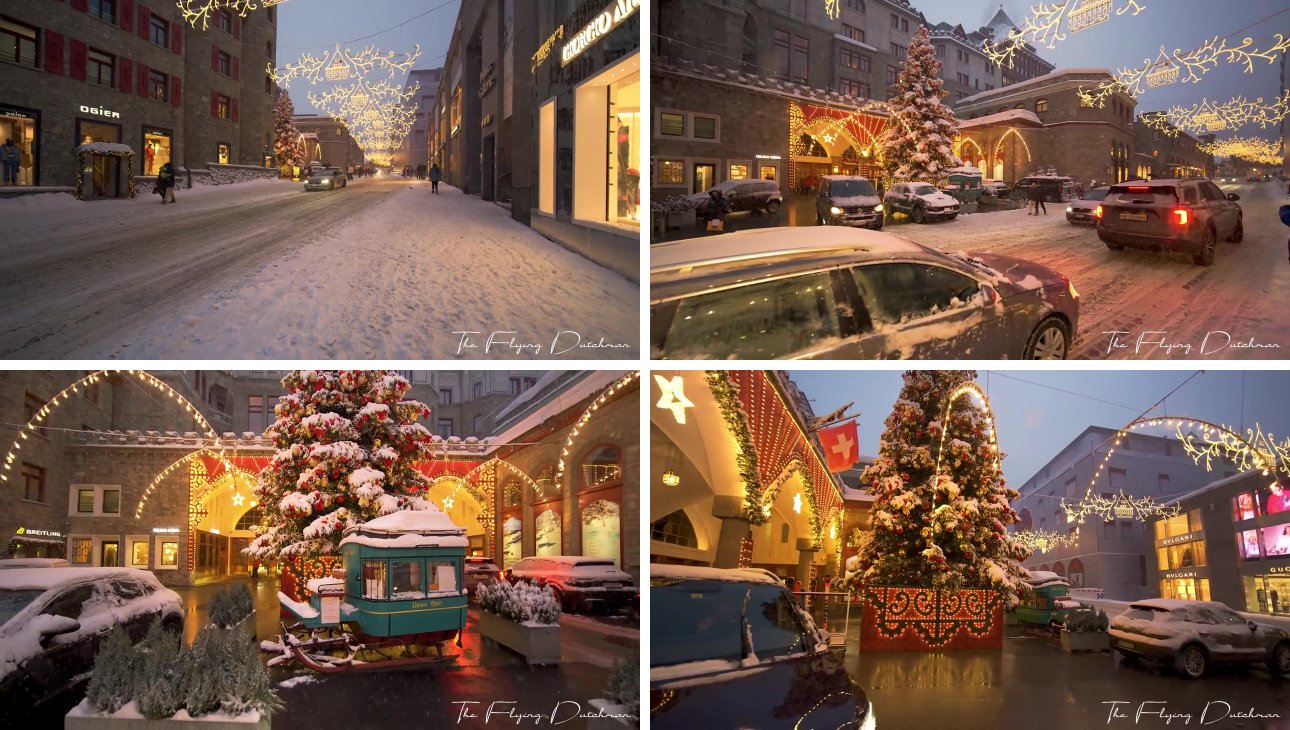} &
  \includegraphics[width=\sz\linewidth]{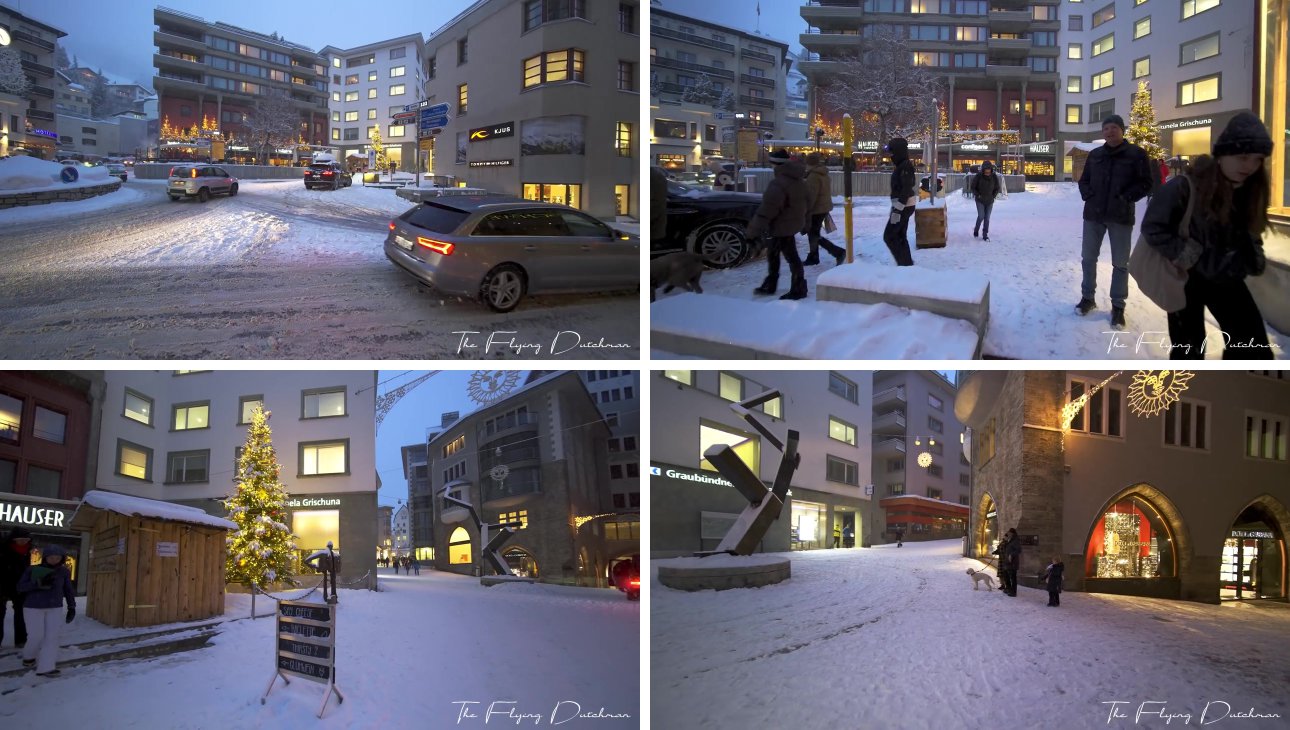} &
  \includegraphics[width=\sz\linewidth]{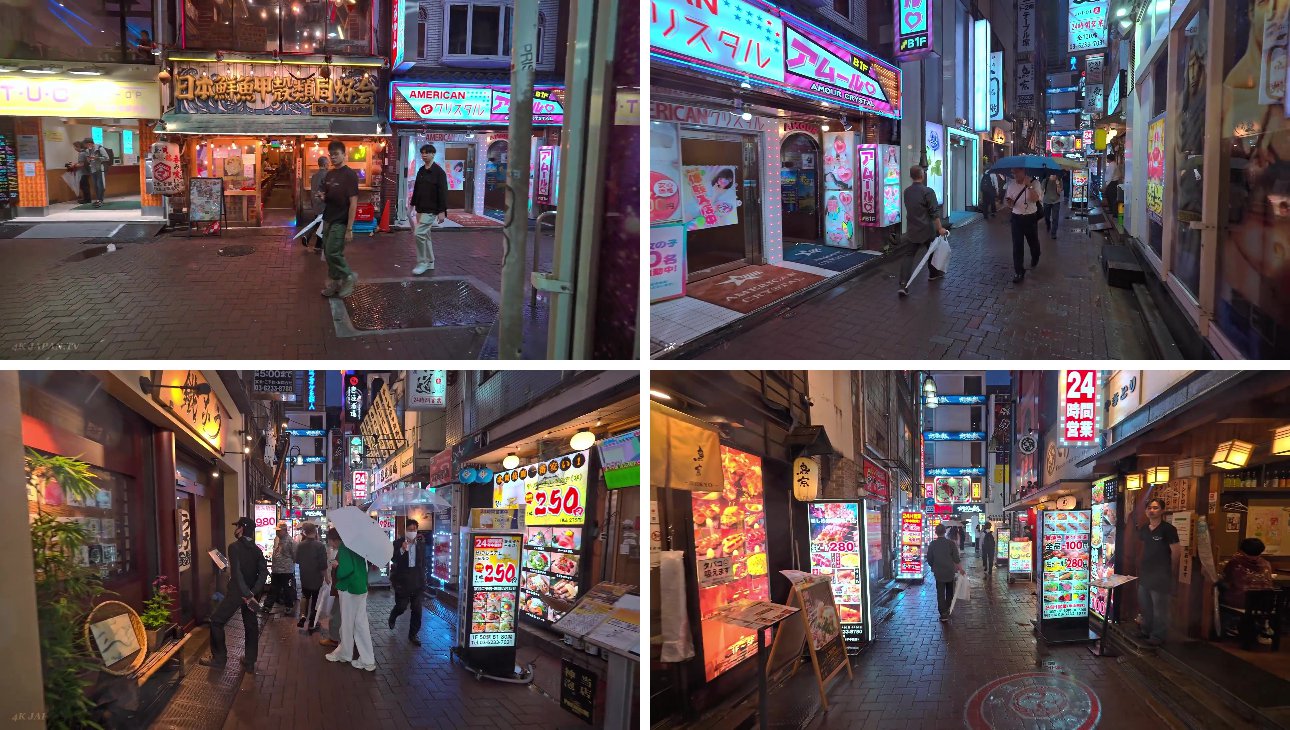} &
  \includegraphics[width=\sz\linewidth]{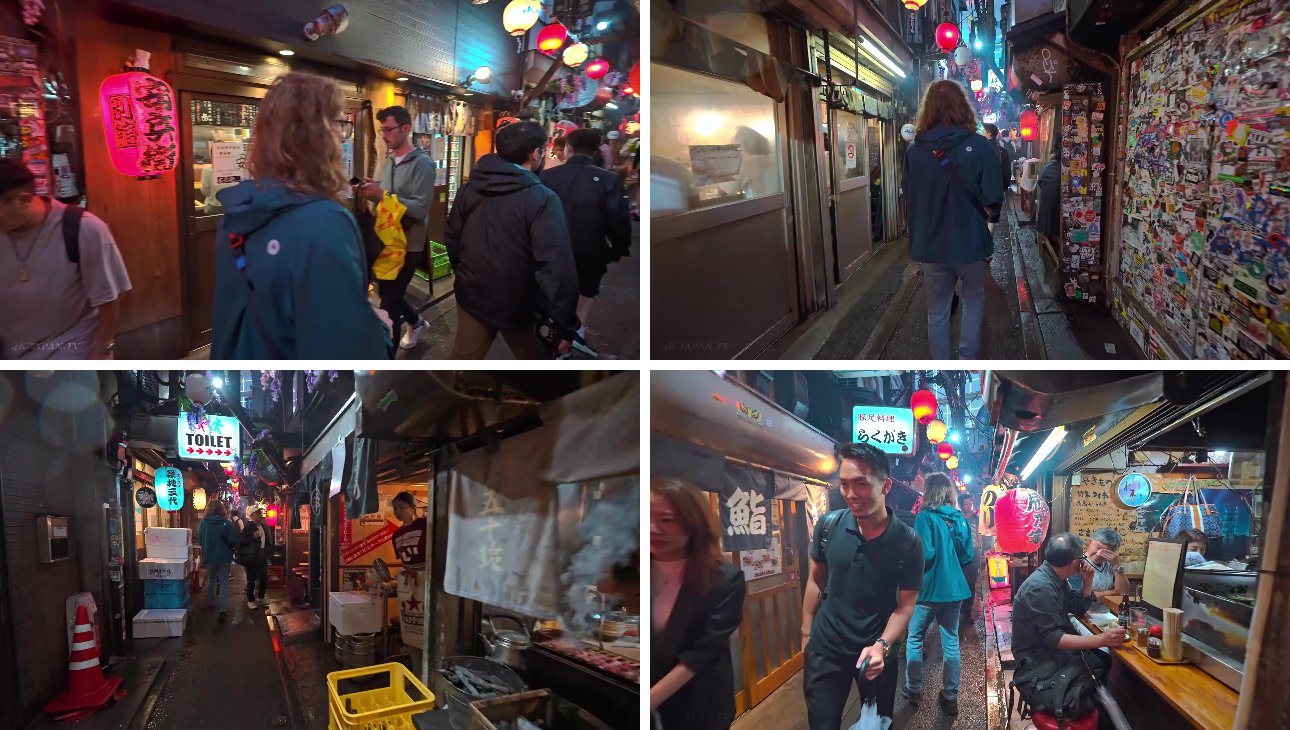}
  \\[0.5mm]

  \raisebox{5.0\normalbaselineskip}[0pt][0pt]{\rotatebox[origin=c]{90}{WildGS-SLAM \cite{zheng2025wildgs}}} &
  \includegraphics[width=\sz\linewidth]{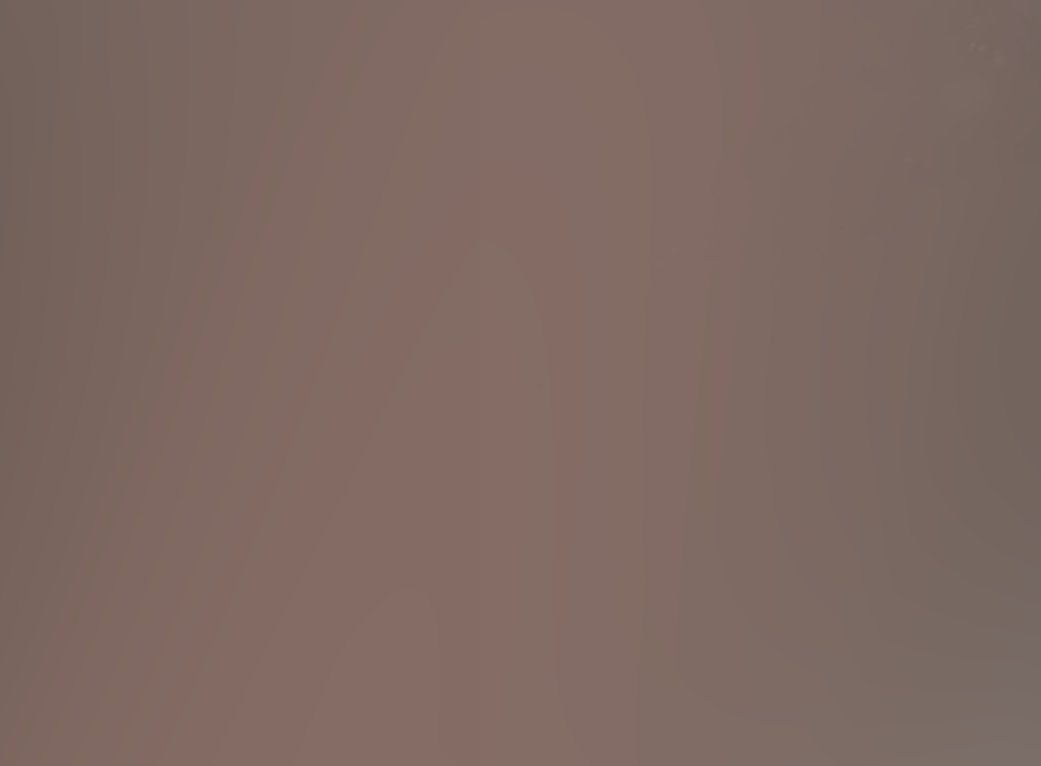} &
  \includegraphics[width=\sz\linewidth]{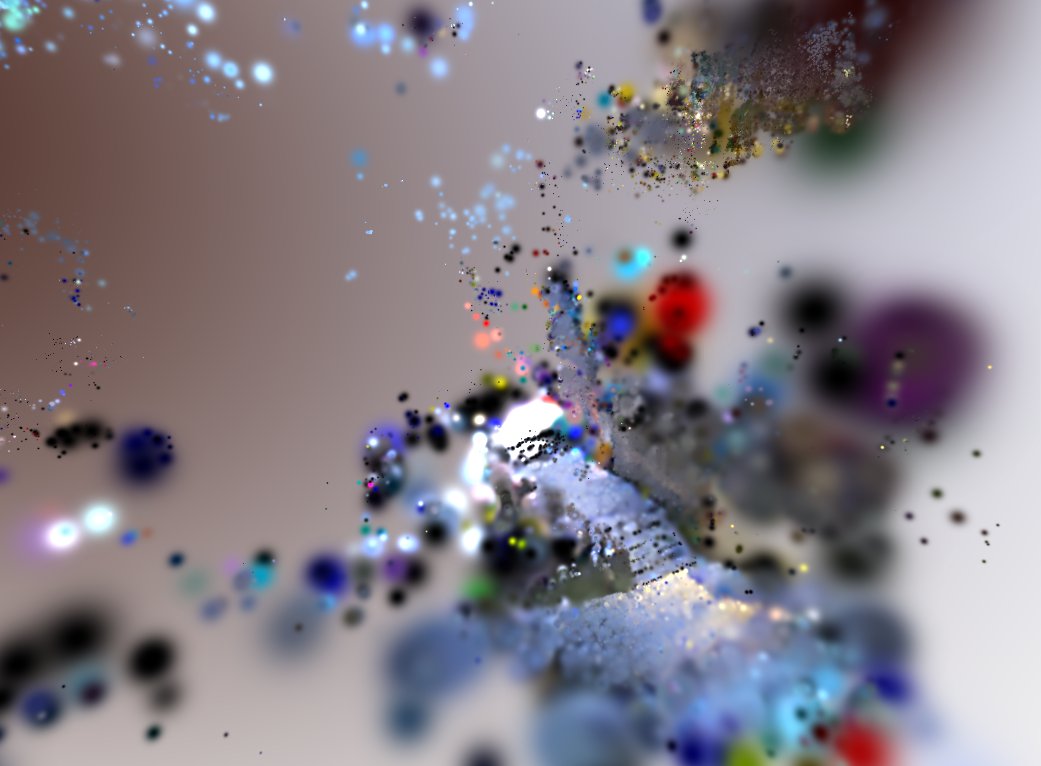} &
  \includegraphics[width=\sz\linewidth]{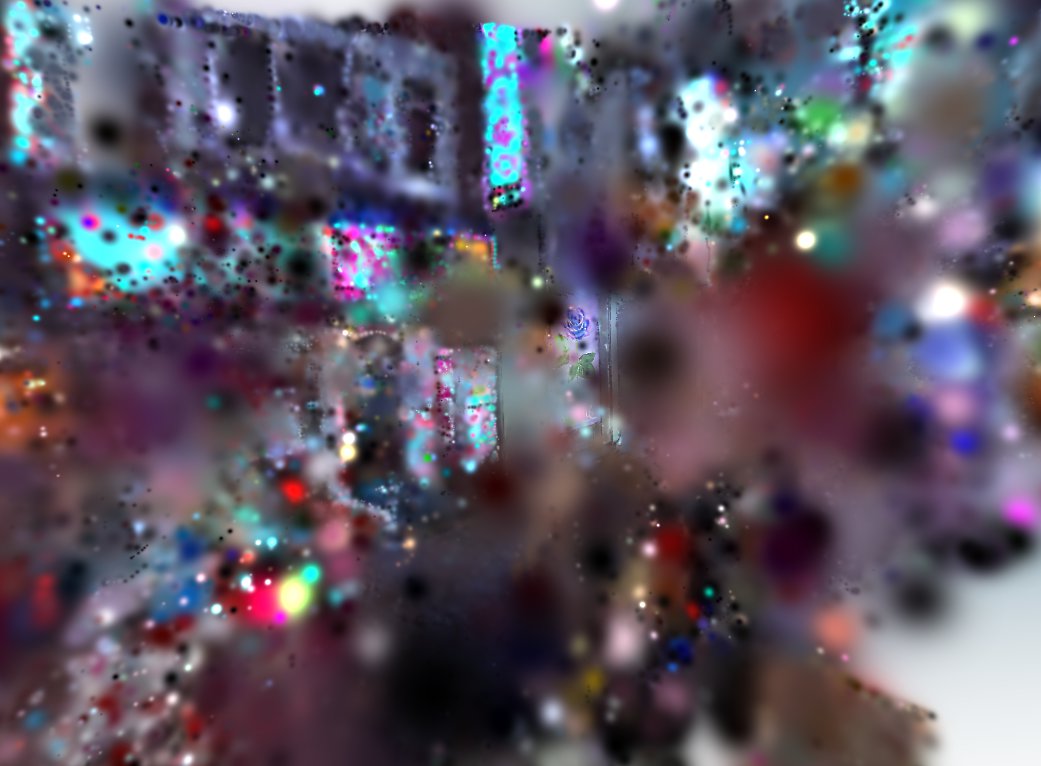} &
  \includegraphics[width=\sz\linewidth]{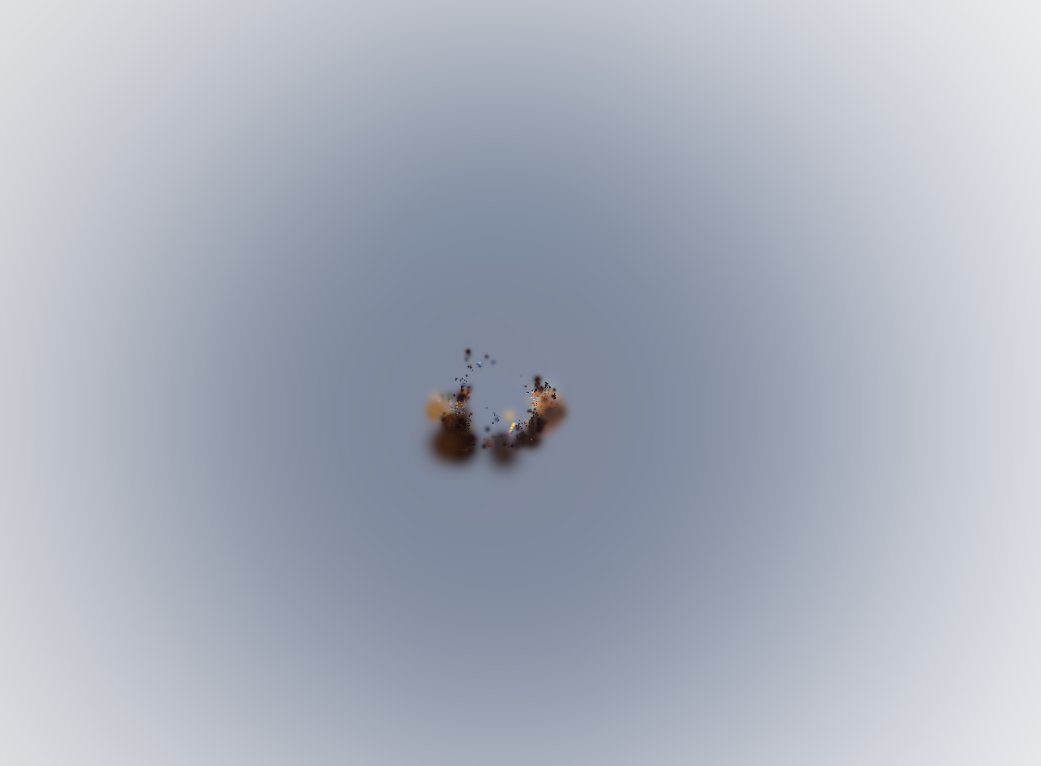}
  \\[0.5mm]

  \raisebox{5.0\normalbaselineskip}[0pt][0pt]{\rotatebox[origin=c]{90}{DROID-SLAM \cite{teed2021droid}}} &
  \includegraphics[width=\sz\linewidth]{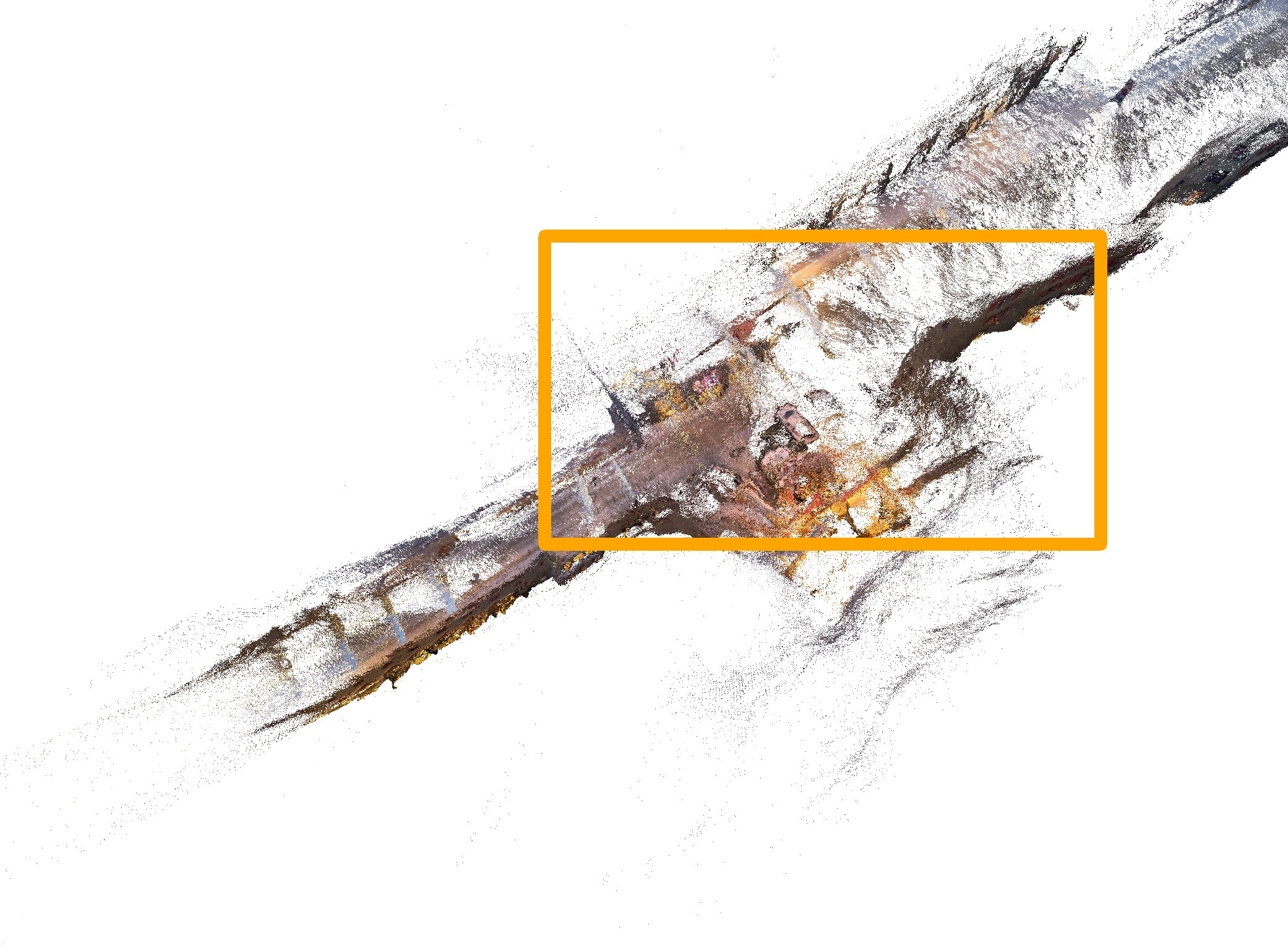} &
  \includegraphics[width=\sz\linewidth]{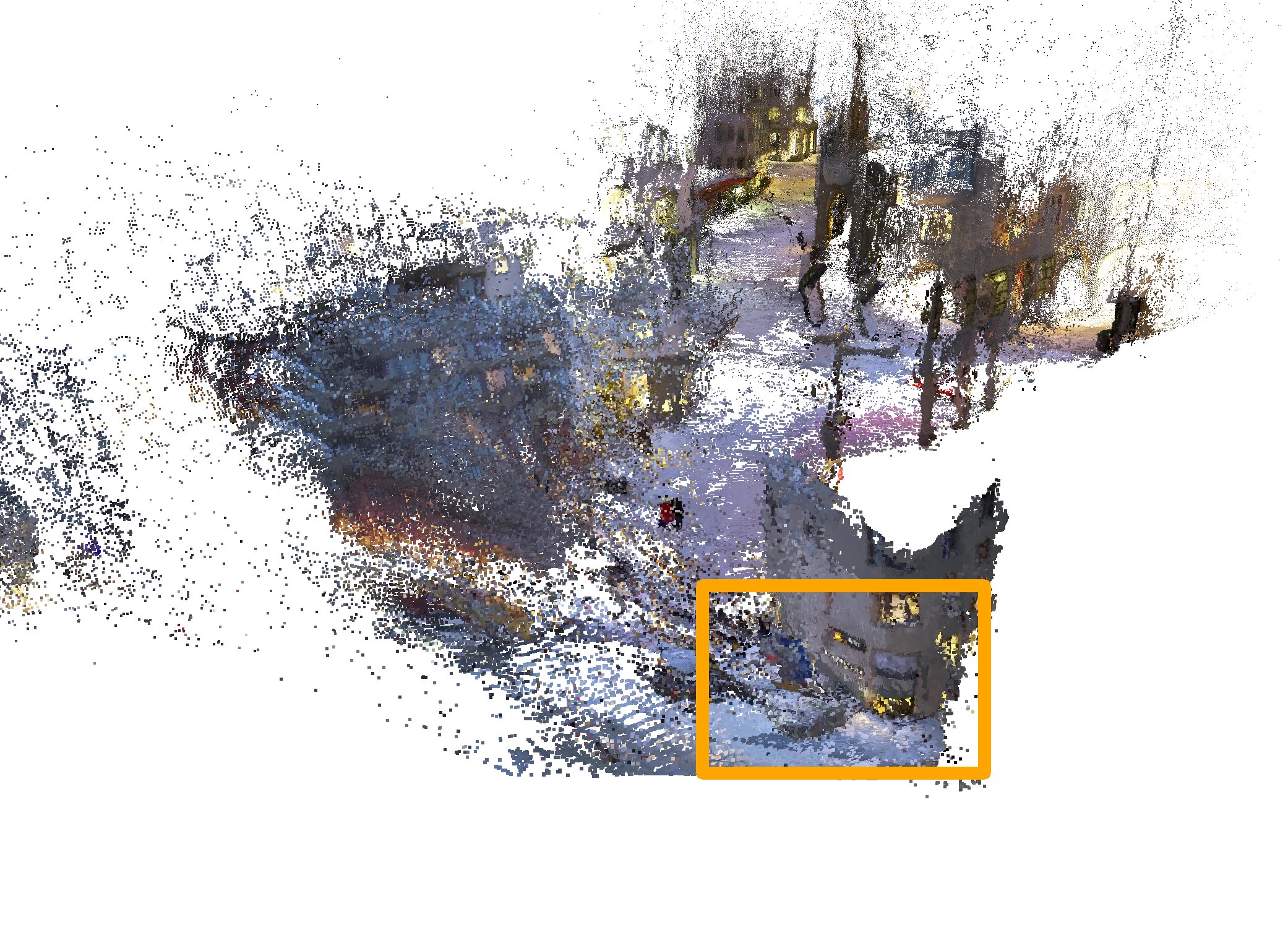} &
  \includegraphics[width=\sz\linewidth]{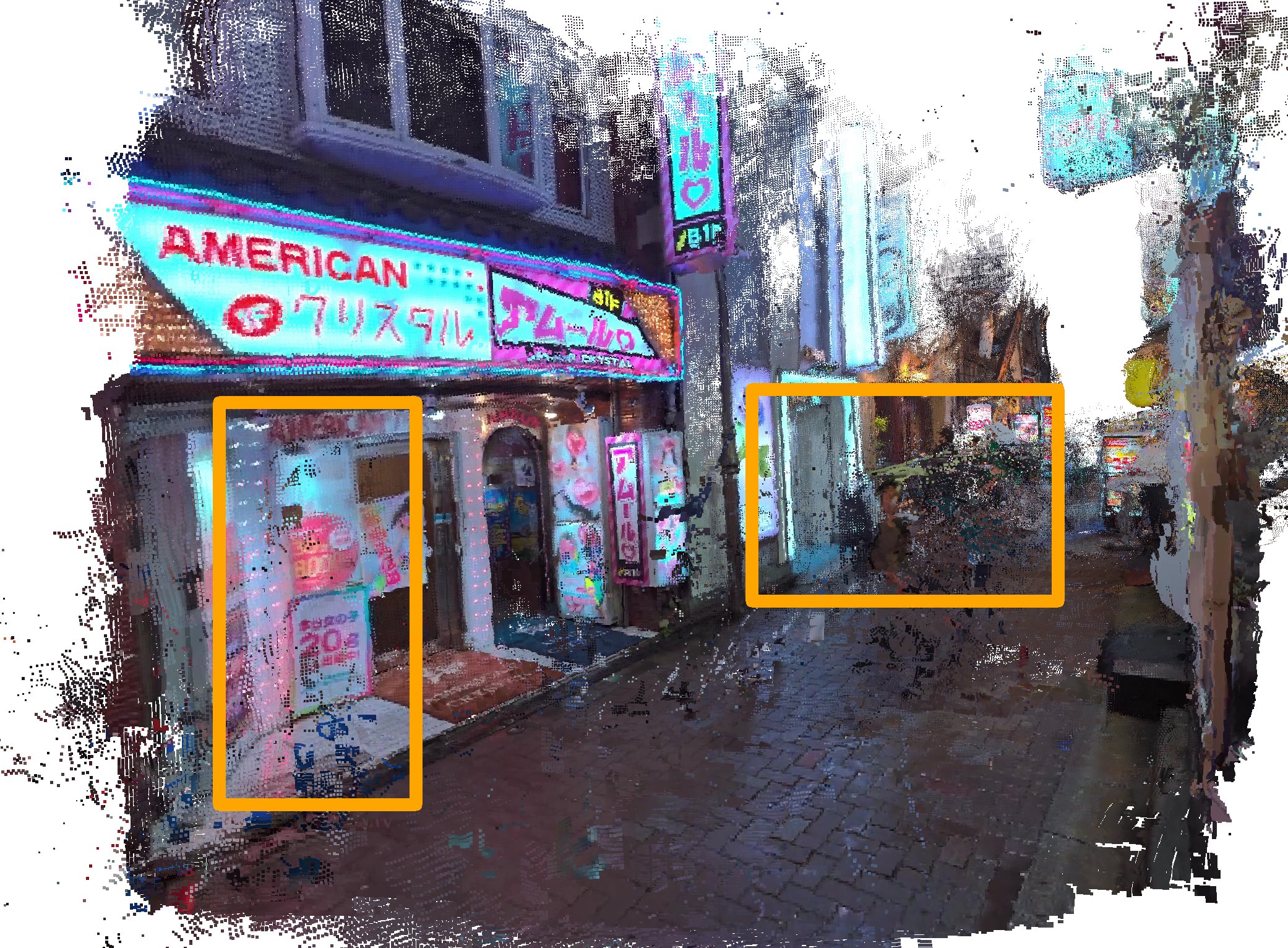} &
  \includegraphics[width=\sz\linewidth]{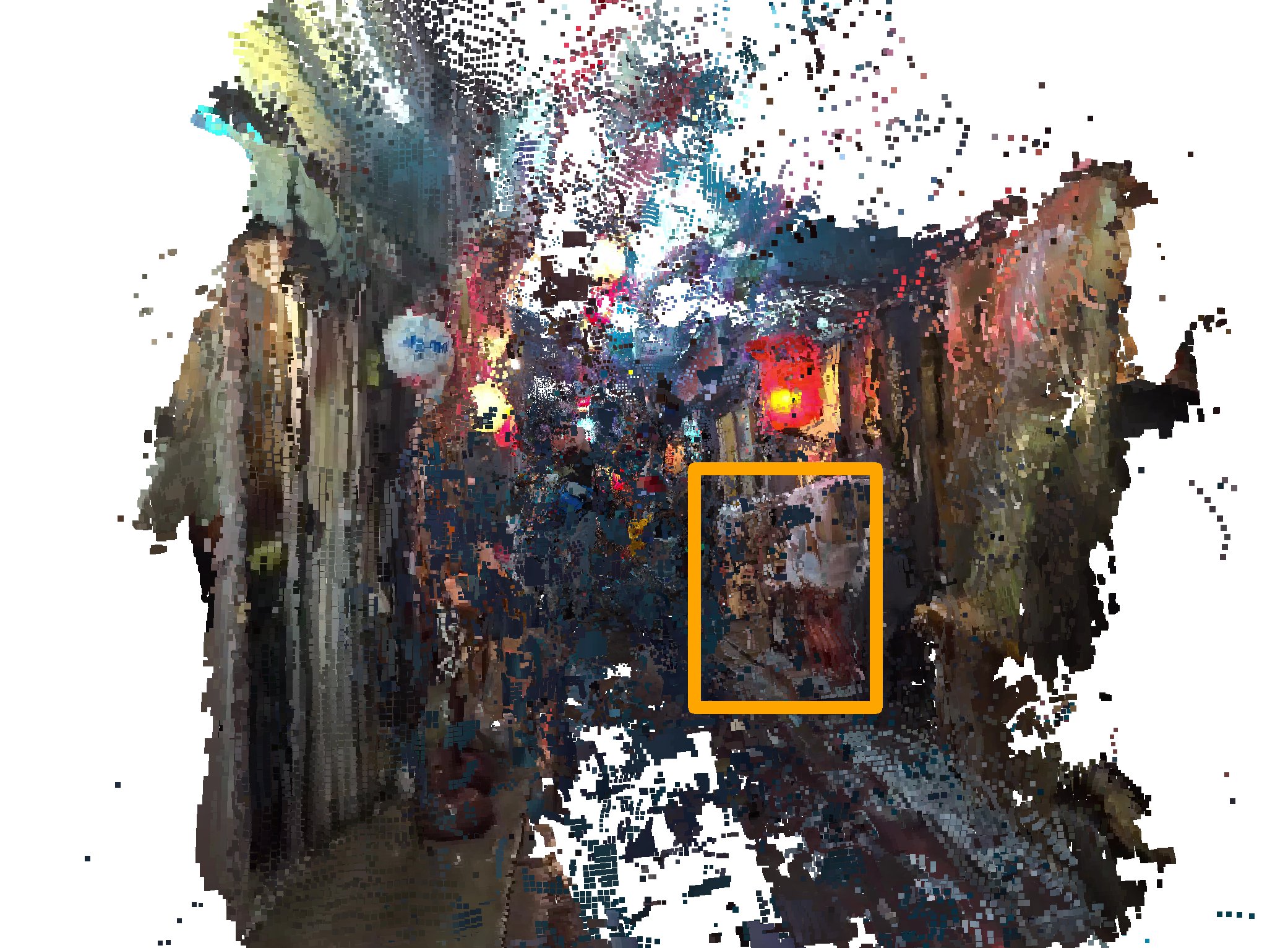}
  \\[0.5mm]

  \raisebox{5.0\normalbaselineskip}[0pt][0pt]{\rotatebox[origin=c]{90}{\textbf{Ours}}} &
  \includegraphics[width=\sz\linewidth]{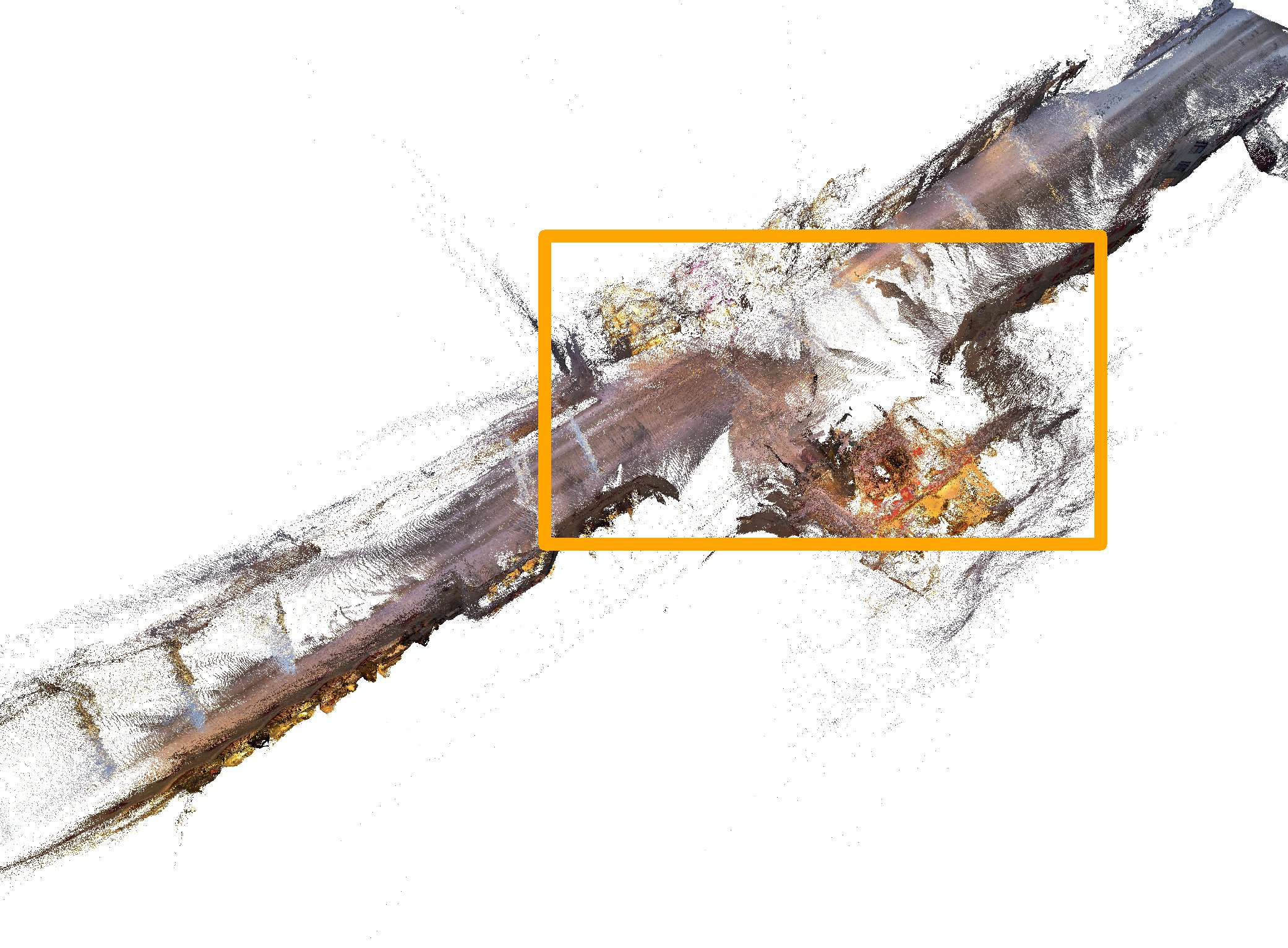} &
  \includegraphics[width=\sz\linewidth]{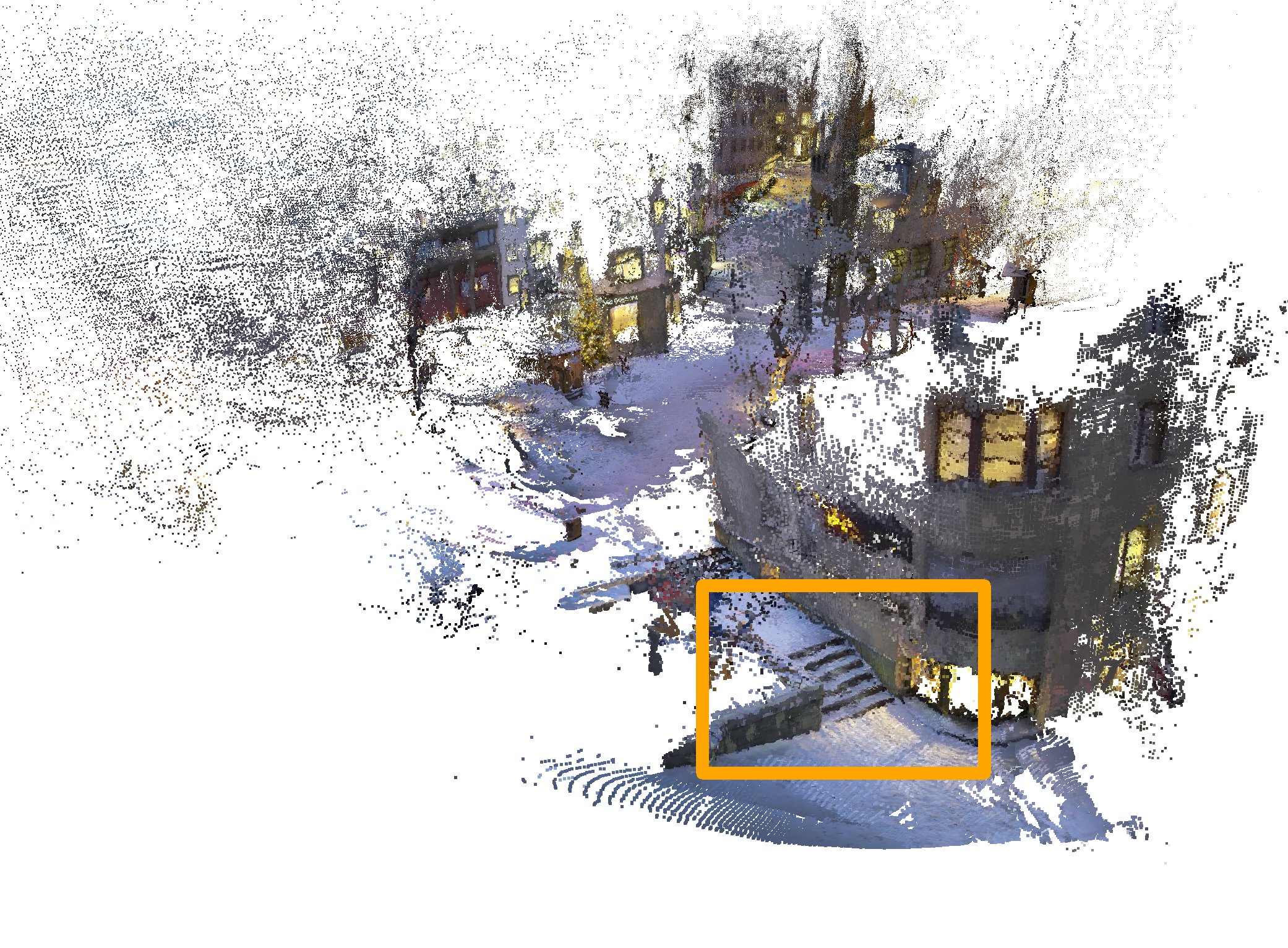} &
  \includegraphics[width=\sz\linewidth]{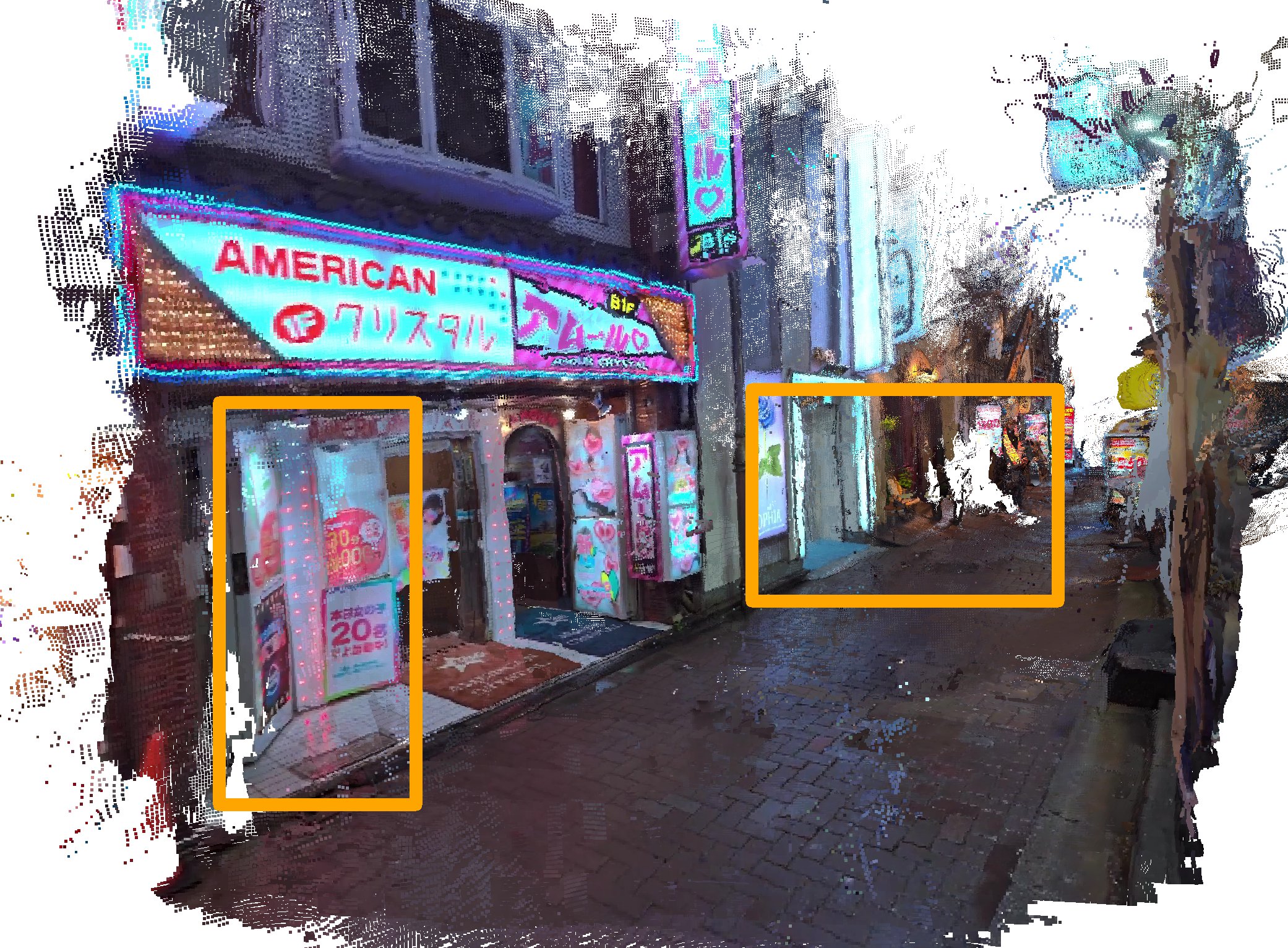} &
  \includegraphics[width=\sz\linewidth]{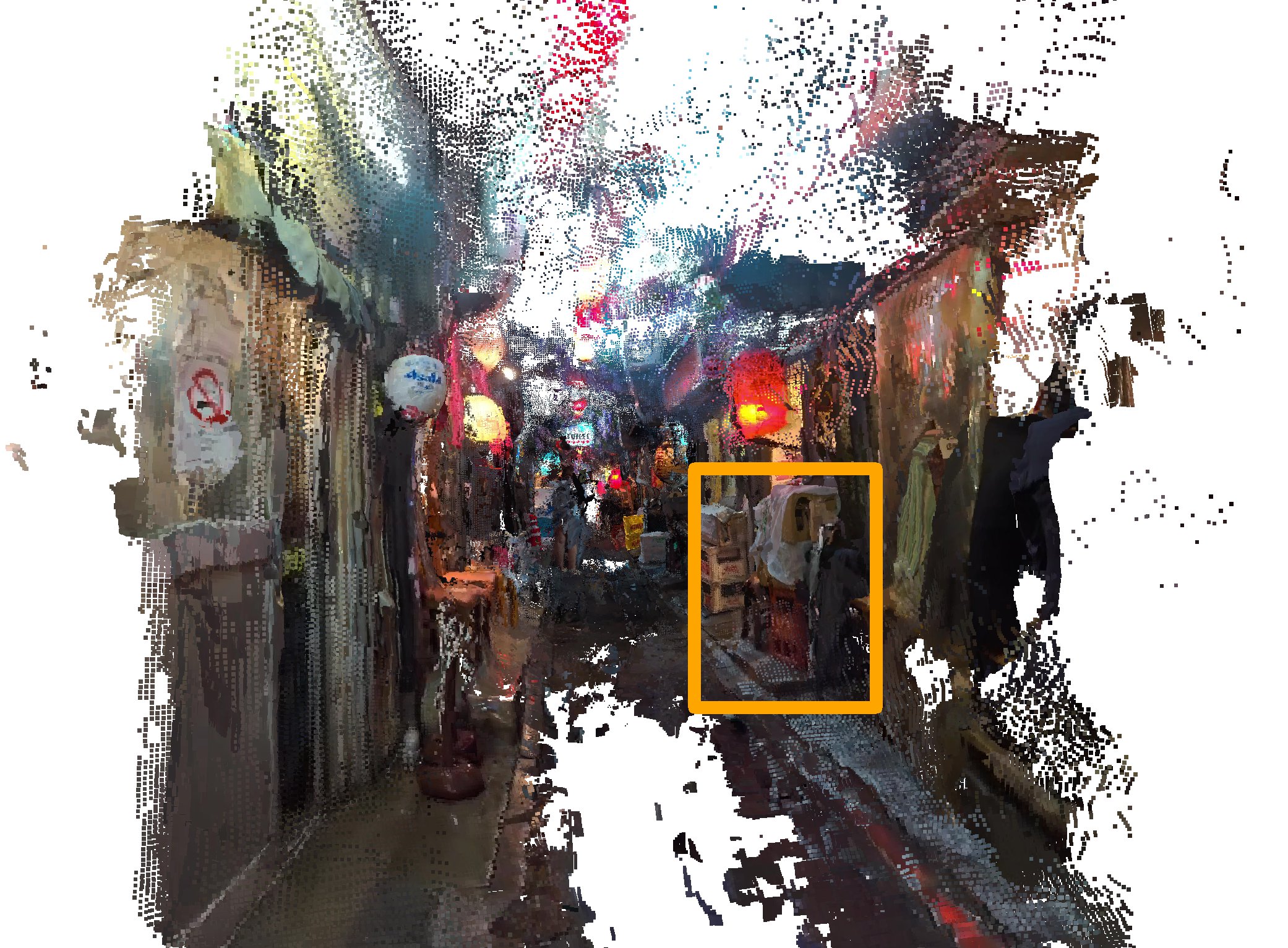}
  \\[0.5mm]

  & {\fontsize{9}{9} \selectfont {\texttt{St. Moritz 1}}} 
  & {\fontsize{9}{9} \selectfont {\texttt{St. Moritz 3}}} 
  & {\fontsize{9}{9} \selectfont {\texttt{Tokyo Walking 2}}} 
  & {\fontsize{9}{9} \selectfont {\texttt{Tokyo Walking 3}}} \\

  \end{tabular}
  \vspace{-2mm}
  \caption{\textbf{3D Reconstruction Comparisons on YouTube Sequences.}
  We compare 3D reconstruction quality of DROID-SLAM~\cite{teed2021droid}, WildGS-SLAM~\cite{zheng2025wildgs}, and our method.
  Point clouds from DROID-SLAM and ours are visualized directly, while Gaussian renderings from WildGS-SLAM are displayed using the 3DGS viewer. 
  WildGS-SLAM fails on most sequences. 
  DROID-SLAM shows obvious scale drift (\texttt{St. Moritz 1}), inaccurate geometry (\texttt{St. Moritz 3}), and noisy distractors (\texttt{Tokyo Walking 2 \& 3}) under challenging dynamic environments. Our approach produces accurate and consistent reconstructions across highly dynamic and visually challenging real-world sequences.}
  \label{fig:recon_comparisons}
  \vspace{-3mm}
\end{figure*}

\begin{table}[t]
\vspace{-2mm}
\centering
\footnotesize
\setlength{\tabcolsep}{8pt}
{
    \begin{tabular}{lc}
        \toprule
        {Method} & ATE RMSE [cm] \\
        \midrule
        a. w/o Uncertainty-aware BA & 5.13 \\
        b. w/o monocular depth & 3.30 \\
        c. w/o uncertainty decouple & 2.57 \\
        d. w/o affine mapping & 2.47 \\
        e. w/o weight decay & 2.34 \\
        \midrule
        \textbf{Full} & \textbf{2.30} \\
        \bottomrule
    \end{tabular}
}
\vspace{-2mm}
\caption{\textbf{Ablation Studies on Bonn RGB-D Dataset~\cite{palazzolo2019iros}.}
Details about each configuration are described in Sec. \ref{sec:ablation}.}
\label{tab:ablation_study}
\vspace{-6mm}
\end{table}






\paragraph{Baselines}
We conduct comparisons with both \textbf{\textit{SLAM-style}} and recent \textbf{\textit{feed-forward}} methods.
For SLAM-style methods, existing methods can be categorized into four groups:
(a) \textit{Classic SLAM}: DSO~\cite{Engel2017DSO}, ORB-SLAM2~\cite{Mur2017orb2}, and DROID-SLAM~\cite{teed2021droid};  
(b) \textit{Classic dynamic SLAM}: ReFusion~\cite{palazzolo2019iros} and DynaSLAM~\cite{bescos2018dynaslam};  
(c) \textit{NeRF-/GS-based SLAM in static environments}: NICE-SLAM~\cite{Zhu2022CVPR}, and Splat-SLAM~\cite{sandstrom2024splat};  
(d) \textit{NeRF-/GS-based SLAM in dynamic environments}: DG-SLAM~\cite{xu2024dgslam}, RoDyn-SLAM~\cite{jiang2024rodyn}, DDN-SLAM~\cite{li2024ddn}, DynaMoN~\cite{schischka2023dynamon}, UP-SLAM~\cite{zheng2025upslam}, and ADD-SLAM~\cite{wu2025addslam}.  
For feed-forward approaches, we compare with MonST3R~\cite{zhang2024monst3r} and the very recent TTT3R~\cite{chen2025ttt3r}.

\paragraph{Metrics}
We use the Absolute Trajectory Error (ATE) to evaluate camera tracking accuracy. For the DyCheck dataset~\cite{gao2022dycheck}, we follow MegaSaM~\cite{li2024megasam} and normalize the ground-truth camera trajectories to unit length, as the sequence lengths in this dataset vary significantly. Following DROID-SLAM, our approach performs optimization only for keyframes. To evaluate full trajectories, we recover non-keyframe poses through SE(3) interpolation followed by a pose graph update. 
For all methods, we align the estimated camera trajectory with the ground-truth camera trajectory through Sim(3) Umeyama alignment \cite{umeyama1991least}.
In addition to tracking accuracy, for each method, we report the average run-time by dividing the number of input frames by the total time.

\subsection{Experimental Results}

\noindent
\textbf{Quantitative Results.}
Camera tracking results on four benchmarks are reported in Tables~\ref{tab:bonn_tracking}, \ref{tab:tum_tracking}, \ref{tab:dycheck_tracking}, and~\ref{tab:stuttgart_tracking}.
\tabref{tab:bonn_tracking} indicates that our approach achieves the best camera tracking accuracy across all baselines on the Bonn RGB-D Dynamic dataset~\cite{palazzolo2019iros} due to effective uncertainty optimization.
As shown in \tabref{tab:tum_tracking}, WildGS-SLAM \cite{zheng2025wildgs} exhibits a noticeable performance drop compared to DROID-SLAM \cite{teed2021droid} on low-dynamic sequences (\texttt{f3/sr}, \texttt{f3/shs}). This gap mainly stems from the unreliable uncertainty estimation, caused by challenging mapping in visually complex environments. In contrast, our method achieves comparable tracking accuracy to DROID-SLAM on low-dynamic scenes and significantly outperforms it on high-dynamic sequences by effectively handling motion-induced inconsistencies.

The DyCheck dataset is characterized by motion and scene diversity across indoor and outdoor scenarios. \tabref{tab:dycheck_tracking} demonstrates that WildGS-SLAM often fails to achieve accurate camera tracking due to the difficulty of scene reconstruction in these complex settings and erroneous uncertainty estimation, whereas our method remains stable and accurate.
On scene \texttt{haru}, where a moving dog dominates the view, our accurate uncertainty estimation suppresses dynamic regions. Consequently, fewer reliable background features remain to support tracking, which degrades our performance.
On average, our proposed method outperforms all baselines.
\tabref{tab:stuttgart_tracking} presents the experimental results on the proposed large-scale outdoor dataset DROID-W. Our method shows superior performance over prior works under this extremely challenging condition.
Feed-forward approaches such as MonST3R~\cite{zhang2024monst3r} and TTT3R~\cite{chen2025ttt3r} suffer from substantially higher tracking errors across all benchmarks compared to optimization-based SLAM systems.

Runtime analysis is in Table~\ref{tab:runtime}. We compare with DROID-SLAM  and WildGS-SLAM, the most recent state-of-the-art baseline for monocular dynamic SLAM. Our system achieves a $\mathbf{40\times}$ speedup over WildGS-SLAM and maintains real-time performance at approximately 10~FPS. Our approach is slightly slower than DROID-SLAM due to monocular depth estimation and DINOv2 \cite{oquab2023dinov2} feature extraction.
Overall, these results highlight the effectiveness, robustness, and efficiency of our uncertainty-aware formulation compared with existing SLAM-style and feed-forward baselines.

\noindent
\textbf{Qualitative Comparisons.}
\figref{fig:uncer_comparisons} presents comparisons of the estimated uncertainty maps.
We observe that our approach delivers the most accurate dynamic uncertainty estimates, whereas WildGS-SLAM produces erroneous results near moving objects and severely incorrect predictions on challenging sequences.
As shown in \figref{fig:uncer_comparisons}, TUM RGB-D dataset features motion blur, partial overexposure, and cluttered indoor scenes that easily degrade mapping quality.
Our introduced sequences offer diverse object motion and scene configuration, which poses challenges to high-quality geometric reconstruction.
WildGS-SLAM exhibits degraded performance in these challenging sequences with low-quality imagery and highly textured backgrounds, where erroneous Gaussian reconstruction leads to unstable uncertainty estimates.
MonST3R~\cite{zhang2024monst3r} heavily depends on the alignment of dynamic point clouds predicted by the pretrained model, which often results in incomplete or missed detections of moving objects due to limited generalizability.

In contrast, our method yields spatially coherent, semantically consistent uncertainty maps. It sharply delineates dynamic regions and maintains stable confidence in static areas across challenging scenarios, demonstrating the robustness of our uncertainty optimization.

Finally, we compare the reconstruction quality in challenging YouTube sequences.
\figref{fig:recon_comparisons} illustrates that DROID-SLAM~ produces inaccurate point clouds in dynamic scenes, as moving distractors lead to unreliable reprojection residuals and disrupt pose estimation. The reconstructions of DROID-SLAM exhibit scale drift (\texttt{St. Moritz 1}), erroneous geometry (\texttt{St. Moritz 3}), and noisy distractors (\texttt{Tokyo Walking 1 \& 2}).
WildGS-SLAM struggles to reconstruct Gaussian maps under these conditions, resulting in near-complete failures on all sequences.
In contrast, our method yields geometrically accurate and temporally consistent point clouds, maintaining stable reconstruction quality even in challenging outdoor scenarios.

\subsection{Ablation Study}
\label{sec:ablation}

We conduct ablations of the main modules in \tabref{tab:ablation_study}. In the \textit{a. w/o Uncertainty-aware BA} setting, we disable the uncertainty update and weight the reprojection term solely by the confidence map. For the experiments \textit{c. w/o uncertainty decouple}, we modify the similarity loss in \eqnref{eq:sim_loss} as follows:
\begin{equation}
    \footnotesize
    \mathbf{E}_{\text{sim}}^{*}(\bu') = 
    \sum_{(i,j) \in \mathcal{E}} \frac{1 - \frac{\bF_i \cdot {\bF}_{ij}}{ {\|\bF_i\|}_2 {\|{\bF}_{ij}\|}_2 }}{{{\bu}'_i}^2}.
    \label{eq:sim_loss_no_decouple}
\end{equation}
In experiments \textit{d. w/o affine mapping}, uncertainties are updated directly rather than by optimizing parameters of the affine mapping. Removing the affine mapping introduces temporal and spatial inconsistencies in uncertainty estimation, leading to degraded performance. In case of \textit{e. w/o weight decay}, the lack of regularization term for affine mapping will cause instability, thereby leading to performance drop on some scenes. As shown in \tabref{tab:ablation_study}, the full system consistently outperforms all variants, validating the effectiveness of each component.

\vspace{-2mm}
\section{Conclusion}
\label{sec:conclusion}
\vspace{-1mm}

In this paper, we presented a novel monocular dynamic SLAM system. Our system optimizes dynamic uncertainty within a differentiable bundle adjustment framework by leveraging multi-view feature similarity. Extensive experiments demonstrate that our effective uncertainty optimization enables robust camera tracking and accurate geometric reconstruction across challenging real-world scenarios, where prior methods often struggle.
Code will be public.

\paragraph{Limitations} Our uncertainty optimization relies on frame-to-frame alignment, which can lead to inaccurate uncertainty estimation during SLAM initialization when pose estimates are still unreliable. Incorporating reconstruction priors could improve the robustness of the initialization stage. 

\section{Acknowledgements}
We thank Wei Zhang for valuable help with collecting the DROID-W dataset.

{
    \small
    \bibliographystyle{ieeenat_fullname}
    \bibliography{main}
}

\clearpage
\setcounter{page}{1}
\maketitlesupplementary

\noindent
In the supplementary material, we provide additional details about the following:
\begin{itemize}
    \item More information about the DROID-W dataset and downloaded YouTube videos (\secref{sec:dataset}).
    \item Details about the Jacobian derivation of our uncertainty optimization (\secref{sec:jacobians}).
    \item Additional qualitative comparisons for uncertainty, point clouds, and ablation study (\secref{sec:additional_exp}).
\end{itemize}

\section{Dataset}
\label{sec:dataset}
\begin{table}[t]
\centering
\footnotesize
\setlength{\tabcolsep}{7pt}
{
    \begin{tabular}{lcc}
        \toprule
        Sequence & Number of Frames & Length of Trajectory [m] \\
        \midrule
        \texttt{Downtown 1} & 1427 & \phantom{0}90.83 \\
        \texttt{Downtown 2} & 2200 & 122.25 \\
        \texttt{Downtown 3} & 1438 & \phantom{0}62.33 \\
        \texttt{Downtown 4} & 1794 & \phantom{0}85.19 \\
        \texttt{Downtown 5} & 2157 & 129.93 \\
        \texttt{Downtown 6} & 1900 & 104.99 \\
        \texttt{Downtown 7} & 1900 & 109.35 \\
        \bottomrule
    \end{tabular}
}
\caption{\textbf{Overview of Our DROID-W Dataset.}}
\label{tab:droid_w_dataset}
\end{table}

\begin{table*}[t]
\centering
\footnotesize
{
\setlength{\tabcolsep}{5pt}
\begin{tabular}{l
    >{\centering\arraybackslash}p{3.0cm}
    >{\centering\arraybackslash}p{1.4cm}
    >{\centering\arraybackslash}p{1.4cm}
    >{\centering\arraybackslash}p{1.4cm}
    >{\centering\arraybackslash}p{1.4cm}
    >{\centering\arraybackslash}p{1.4cm}
    !{\smash{\tikz[baseline]{\draw[densely dashed, gray!80, line width=0.8pt] (0pt,-2pt)--(0pt,8pt);}}}
    >{\centering\arraybackslash}p{1.4cm}
}
\toprule
Method & Inputs & \texttt{Downtown 3} & \texttt{Downtown 4} & \texttt{Downtown 5} & \texttt{Downtown 6} & \texttt{Downtown 7} & \textbf{Avg.}\\
\midrule
FAST-LIVO2 \cite{zheng2024fast_livo2} & RGB + IMU + LiDAR & 0.06 & 0.06 & 0.09 & 0.09 & 0.06 & 0.071 \\

{\project{}} & RGB
& 0.15 & 0.32 & 0.24 & 0.43 & 0.07 & 0.242 \\
\bottomrule
\end{tabular}
}
\caption{\textbf{Tracking performance of FAST-LIVO2~\cite{zheng2024fast_livo2} on the DROID-W dataset} (ATE RMSE~$\downarrow$~[m]).
FAST-LIVO2 is an efficient and accurate LiDAR-inertial-visual fusion system capable of delivering centimeter-level localization accuracy. Its performance on \texttt{Downtown 3-7} demonstrates sufficient accuracy to serve as ground truth for \texttt{Downtown 1-2}.}
\label{tab:fast_livo2}
\end{table*}

\begin{table}[t]
\centering
\footnotesize
\setlength{\tabcolsep}{11pt}
{
    \begin{tabular}{lccc}
        \toprule
        Sequence & Time & FPS & Resolution \\
        \midrule
        \texttt{Elephant Herd} & 00:08 & 24 & 1280 \(\times\) 720 \\
        \texttt{Giraffe} & 00:09 & 24 & 1280 \(\times\) 720 \\
        \texttt{Taylor} & 01:13 & 30 & 1280 \(\times\) 720 \\
        \texttt{Tomyum 1} & 01:40 & 30 & 1280 \(\times\) 720 \\
        \texttt{Tomyum 2} & 01:40 & 30 & 1280 \(\times\) 720 \\
        \texttt{St. Moritz} & 30:00 & 50 & 1280 \(\times\) 720 \\
        \texttt{Tokyo Walking 1} & 00:50 & 60 & 1920 \(\times\) 1080 \\
        \texttt{Tokyo Walking 2} & 00:22 & 60 & 1920 \(\times\) 1080 \\
        \texttt{Tokyo Walking 3} & 00:40 & 60 & 1920 \(\times\) 1080 \\
        \bottomrule
    \end{tabular}
}
\caption{\textbf{Overview of Downloaded YouTube Videos.}}
\label{tab:youtube_videos}
\end{table}

Prior benchmarks on dynamic SLAM are limited to indoor environments, exhibiting simple object motions and lacking truly challenging real-world conditions. To enable evaluation in more complex and unconstrained settings, we introduce an outdoor dataset, DROID-W, and additionally download 6 challenging videos from YouTube.

\paragraph{DROID-W Dataset}
The DROID-W dataset is captured using a Livox Mid-360 LiDAR rigidly mounted with an RGB camera. It comprises 7 outdoor sequences (\texttt{Downtown 1-7}) with RGB frames at a resolution of 1200${\times}$1600, ground-truth camera poses, and synchronized IMU and LiDAR measurements. The RGB stream is recorded at 20 FPS, while RTK provides ground-truth poses at 10 Hz. 
We use the estimated trajectories from FAST-LIVO2~\cite{zheng2024fast_livo2} as ground truth for \texttt{Downtown 1-2} due to the absence of RTK measurements. As shown in \tabref{tab:fast_livo2}, FAST-LIVO2 provides sufficiently accurate estimates to serve as reliable ground truth for these sequences.
Detailed information, including trajectory lengths and frame numbers, is reported in \tabref{tab:droid_w_dataset}. The DROID-W dataset features long camera trajectories, high scene dynamics, and partial over-exposure -- characteristics commonly encountered in real-world scenarios. We believe this dataset will provide significant value to the community and support future research on robust in-the-wild SLAM.

\paragraph{YouTube Videos}
The framerates of the YouTube videos vary substantially. Therefore, we report the FPS and sequence duration for each sequence in \tabref{tab:youtube_videos}, which provides a more meaningful characterization of camera motion and scene dynamics. Camera intrinsics are estimated using MonST3R~\cite{zhang2024monst3r} from the first 20 frames of each video. The sequences contain a large number of dynamic objects of diverse categories, with many of them moving simultaneously, leading to highly dynamic scenes. They also exhibit challenging and cluttered conditions, including motion blur, strong view-dependent effects, and low dynamic range.
\section{Uncertainty Optimization and Jacobians}
\label{sec:jacobians}
Given the definition in 
\secrefn{sec:uncer_opt}
of the main paper, we obtain the following uncertainty energy function:
\begin{equation}
    \begin{aligned}
        \bu'_i & = \mathrm{log}(\mathrm{exp}(\boldsymbol{\theta} \cdot \bF_i)+1), \\
        \mathbf{E}_{\text{uncer}}(\bu') & 
        = \sum_{(i,j) \in \mathcal{E}} \be_{ij} + \gamma_{\text{prior}} \sum_{i} \mathrm{log}(\bu'_i+1.0), \\
        & = \sum_{(i,j) \in \mathcal{E}}
        \frac{1 - \frac{\bF_i \cdot {\bF}_{ij}}{ {\|\bF_i\|}_2 {\|{\bF}_{ij}\|}_2 }}{\bu'_i \cdot {\bu}'_{ij}} + \gamma_{\text{prior}} \sum_{i} \mathrm{log}(\bu'_i+1.0). \\
    \end{aligned}
    \label{eq:energy_definition}
\end{equation}
Thus, we can derive the following Jacobians:
\begin{equation}
    \begin{aligned}
        \frac{\partial \mathbf{e}_{ij}}{\partial \mathbf{u}'_i} 
        & = - \frac{1 - \frac{\bF_i \cdot {\bF}_{ij}}{ {\|\bF_i\|}_2 {\|{\bF}_{ij}\|}_2 }}{{(\bu'_i)}^2 \cdot \bu'_{ij}} 
        = - \frac{\be_{ij}}{\bu'_i}, \\
        \frac{\partial \mathbf{e}_{ij}}{\partial \mathbf{u}'_{j}} 
        & = - \frac{1 - \frac{\bF_i \cdot {\bF}_{ij}}{ {\|\bF_i\|}_2 {\|{\bF}_{ij}\|}_2 }}{\bu'_i \cdot {(\bu'_{ij})}^2} \cdot \frac{\partial \bu'_{ij}}{\bu'_j} 
        = - \frac{\be_{ij}}{\bu'_{j}} \cdot \boldsymbol{\alpha}_{ij}. \\
    \end{aligned}
    \label{eq:de2du_jocab}
\end{equation}
where $\boldsymbol{\alpha}_{ij} \in \mathcal{R}^{(\frac{H}{8} \times \frac{W}{8}) \times (\frac{H}{8} \times \frac{W}{8})}$ is the bilinear interpolation weight matrix whose non-zero elements are of size $\frac{H}{8} \times \frac{W}{8} \times 4$.
The final Jacobians are defined as follows:
\begin{equation}
    \begin{aligned}
        \frac{\partial{\bE_{uncer}}}{\bu'_l} & = - \sum_{(l,m) \in \mathcal{E}} \frac{\be_{lm}}{\bu'_l} - \sum_{(k,l) \in \mathcal{E}} \frac{\be_{kl}}{\bu'_k} \cdot \boldsymbol{\alpha}_{kl} + \gamma_{\text{prior}} \cdot \frac{1}{\bu'_l + 1.0}, \\
        \frac{\partial{\bu'_l}}{\partial{\boldsymbol{\theta}}} & = \frac{1}{1 + \mathrm{exp}(-\boldsymbol{\theta \cdot \bF_l})} \cdot \bF_l, \\
        \frac{\partial{\bE_\text{uncer}}}{\partial\boldsymbol{\theta}} & = \sum_{l=0}^{N}\frac{\partial{\bE_\text{uncer}}}{\partial\bu'_l} \cdot \frac{\partial{\bu'_l}}{\partial{\boldsymbol{\theta}}}.
    \end{aligned}
    \label{eq:dE_dtheta_jocab}
\end{equation}
Here, frame $l$ serves as the reference frame in all edges $(l,m)$ and as the target frame in all edges $(k,l)$.
\section{Additional Experiments}
\label{sec:additional_exp}
\subsection{Uncertainty Estimation}
\label{sec:additional_uncer}
\begin{figure*}[ht]
  \vspace{-1mm}
  \centering
  \scriptsize
  \setlength{\tabcolsep}{1.1pt}
  \newcommand{\sz}{0.22}
  \newcommand{\sza}{0.115} 
  \begin{tabular}{ccccc}
  \raisebox{3.5\normalbaselineskip}[0pt][0pt]{\rotatebox[origin=c]{90}{\texttt{St. Moritz 1}}} &
  \includegraphics[width=\sz\linewidth]{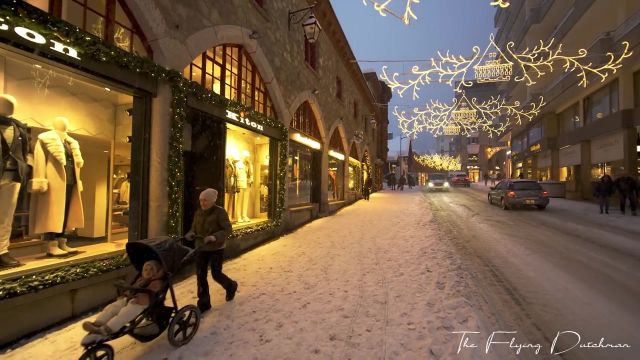}  &
  \includegraphics[width=\sz\linewidth]{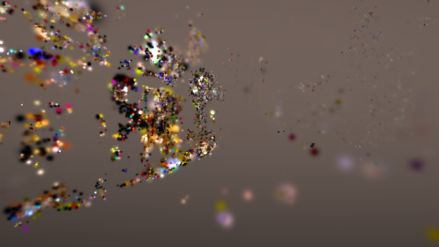} &
  \includegraphics[width=\sz\linewidth]{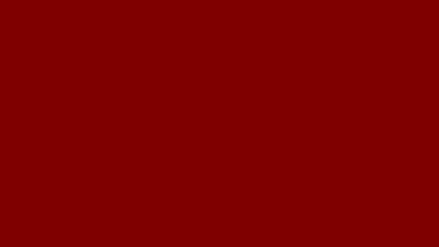} &
  \includegraphics[width=\sz\linewidth]{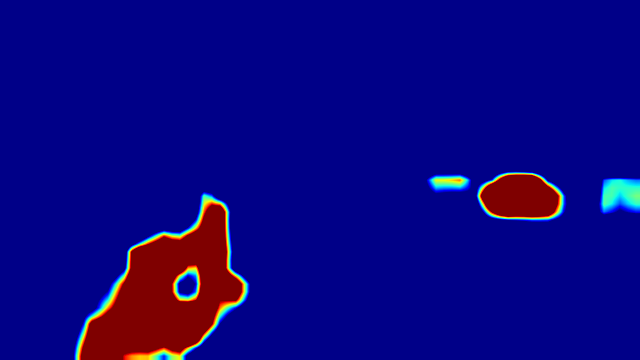}
  \\[-0.1mm]
  
  \raisebox{3.5\normalbaselineskip}[0pt][0pt]{\rotatebox[origin=c]{90}{\texttt{St. Moritz 2}}} &
  \includegraphics[width=\sz\linewidth]{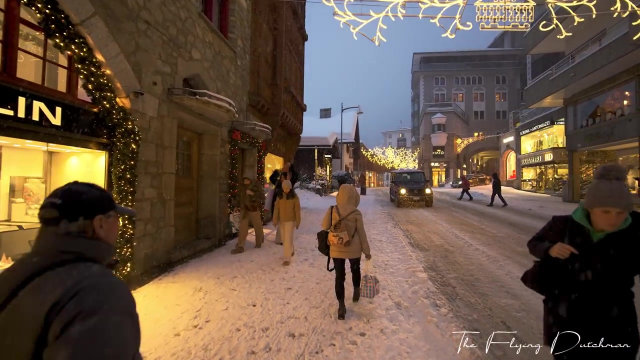}  &
  \includegraphics[width=\sz\linewidth]{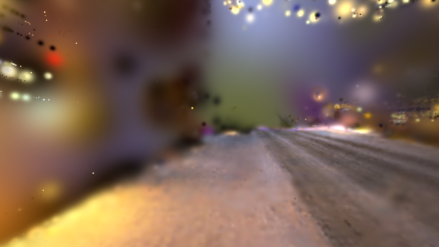} &
  \includegraphics[width=\sz\linewidth]{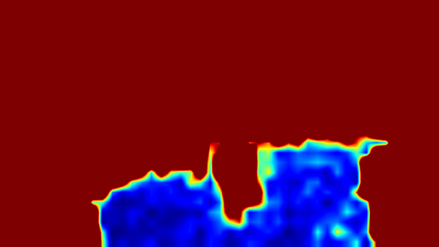} &
  \includegraphics[width=\sz\linewidth]{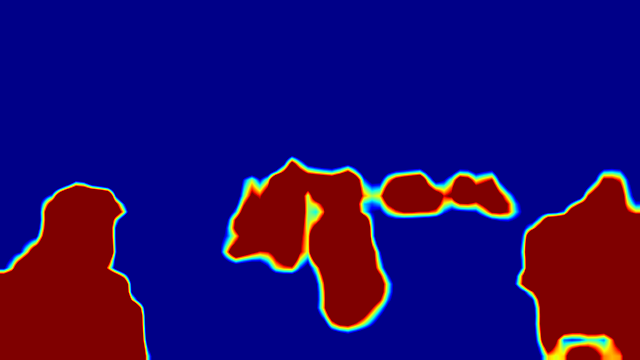}
  \\[-0.1mm]
  
  \raisebox{3.5\normalbaselineskip}[0pt][0pt]{\rotatebox[origin=c]{90}{\texttt{St. Moritz 3}}} &
  \includegraphics[width=\sz\linewidth]{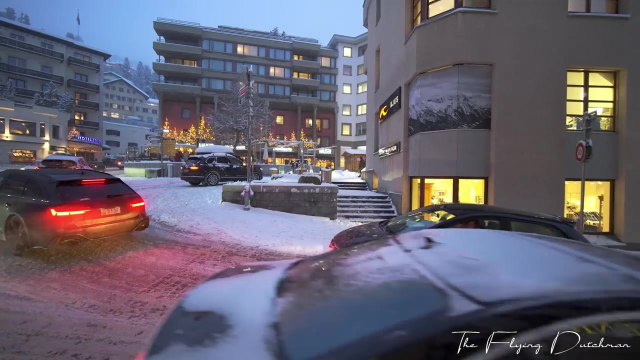}  &
  \includegraphics[width=\sz\linewidth]{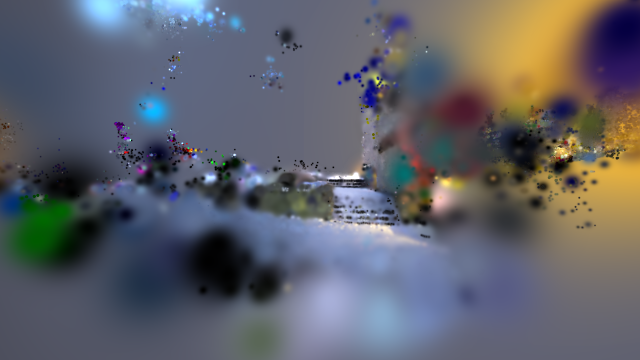} &
  \includegraphics[width=\sz\linewidth]{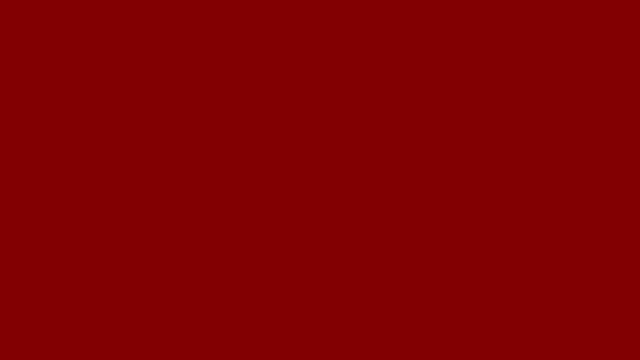} &
  \includegraphics[width=\sz\linewidth]{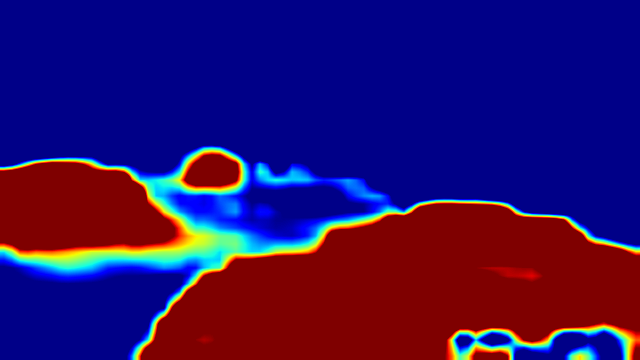}
  \\[-0.1mm]

  \raisebox{3.5\normalbaselineskip}[0pt][0pt]{\rotatebox[origin=c]{90}{\texttt{St. Moritz 4}}} &
  \includegraphics[width=\sz\linewidth]{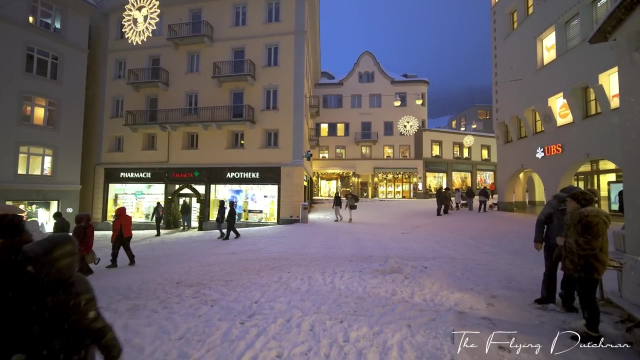}  &
  \includegraphics[width=\sz\linewidth]{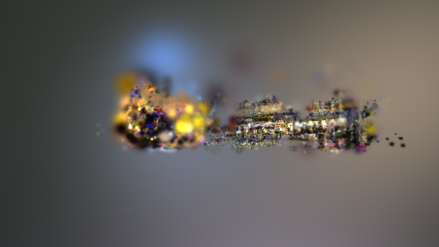} &
  \includegraphics[width=\sz\linewidth]{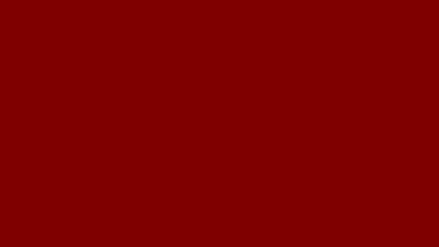} &
  \includegraphics[width=\sz\linewidth]{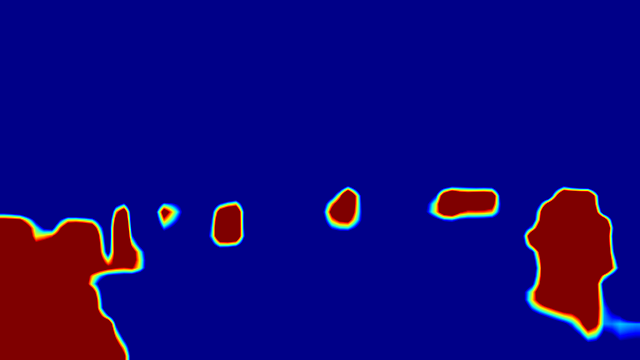}
  \\[-0.1mm]
  
  \raisebox{3.5\normalbaselineskip}[0pt][0pt]{\rotatebox[origin=c]{90}{\texttt{Giraffe}}} &
  \includegraphics[width=\sz\linewidth]{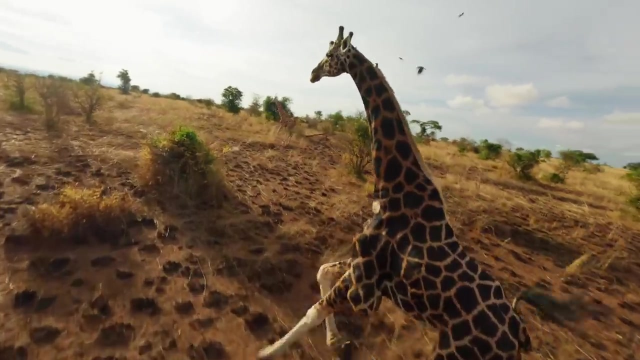}  &
  \includegraphics[width=\sz\linewidth]{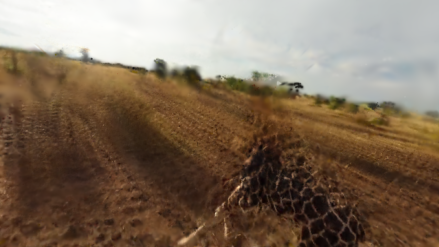} &
  \includegraphics[width=\sz\linewidth]{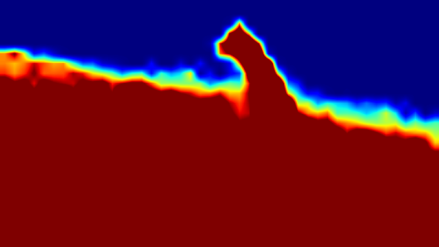} &
  \includegraphics[width=\sz\linewidth]{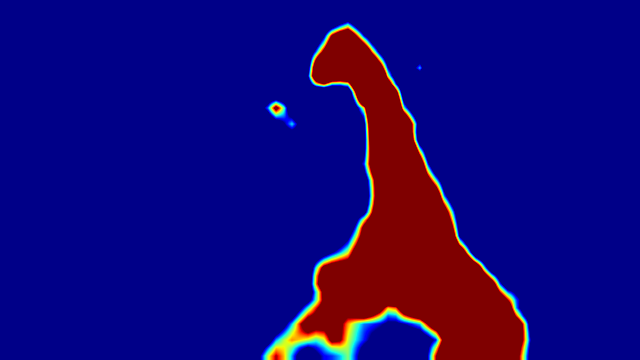}
  \\[-0.1mm]

  \raisebox{3.5\normalbaselineskip}[0pt][0pt]{\rotatebox[origin=c]{90}{\texttt{Taylor 22}}} &
  \includegraphics[width=\sz\linewidth]{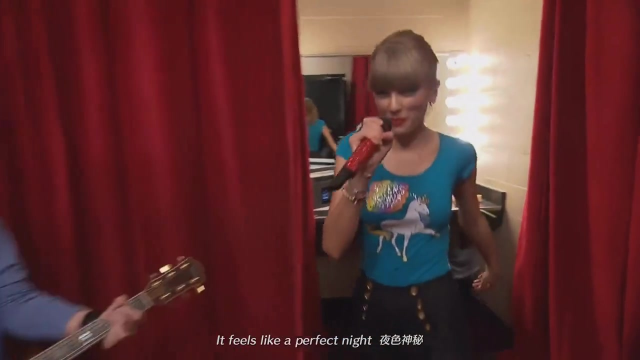}  &
  \includegraphics[width=\sz\linewidth]{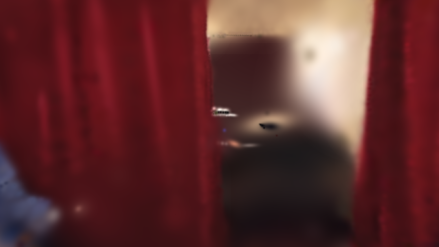} &
  \includegraphics[width=\sz\linewidth]{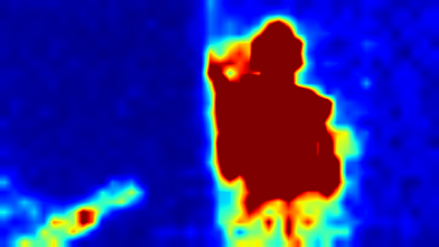} &
  \includegraphics[width=\sz\linewidth]{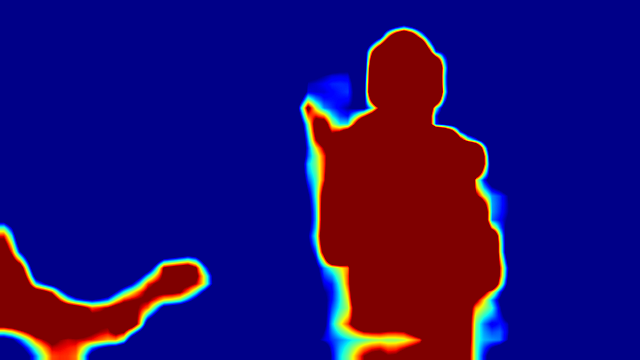}
  \\[-0.1mm]

  \raisebox{3.5\normalbaselineskip}[0pt][0pt]{\rotatebox[origin=c]{90}{\texttt{Tokyo Walking 1}}} &
  \includegraphics[width=\sz\linewidth]{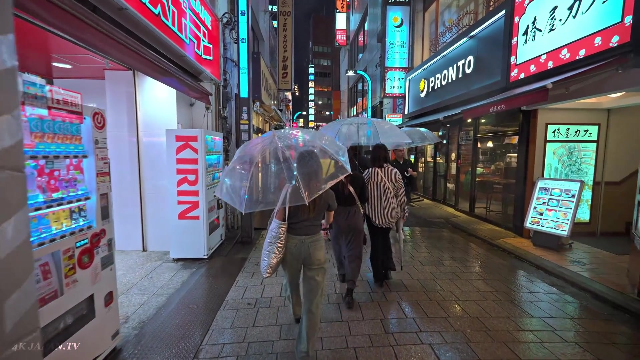}  &
  \includegraphics[width=\sz\linewidth]{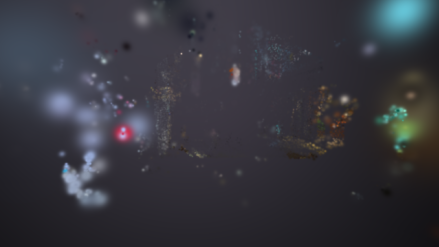} &
  \includegraphics[width=\sz\linewidth]{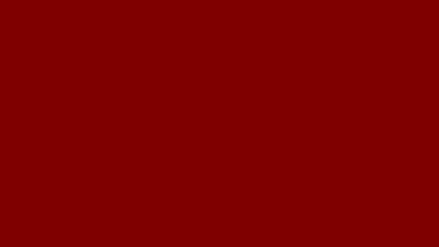} &
  \includegraphics[width=\sz\linewidth]{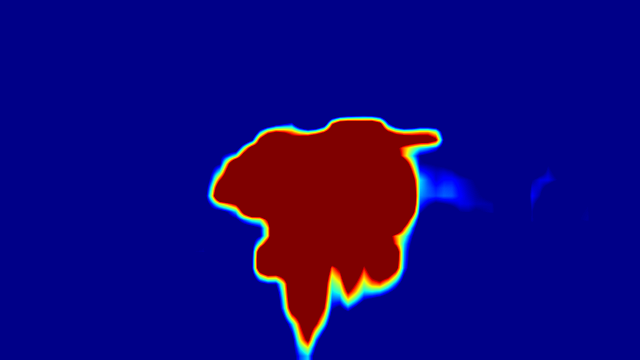}
  \\[-0.1mm]

  \raisebox{3.5\normalbaselineskip}[0pt][0pt]{\rotatebox[origin=c]{90}{\texttt{Tokyo Walking 2}}} &
  \includegraphics[width=\sz\linewidth]{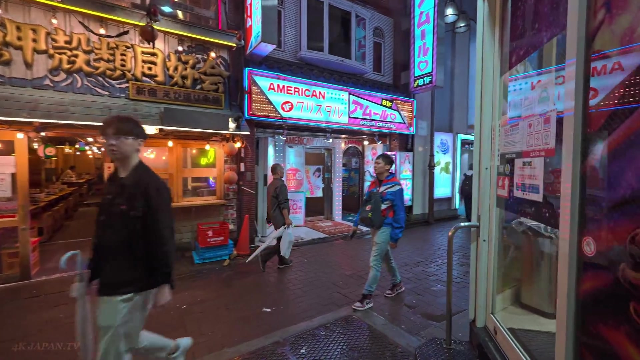}  &
  \includegraphics[width=\sz\linewidth]{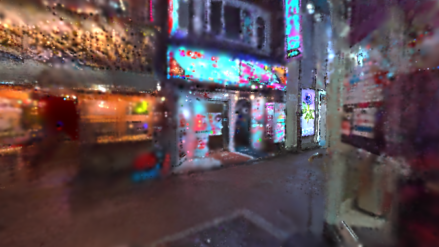} &
  \includegraphics[width=\sz\linewidth]{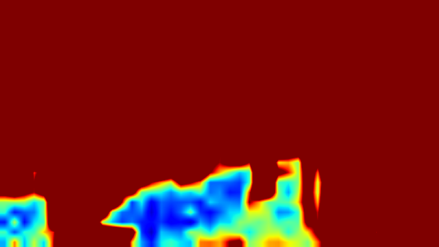} &
  \includegraphics[width=\sz\linewidth]{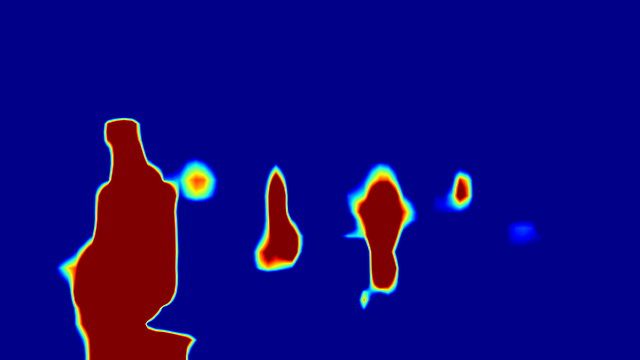}
  \\[-0.1mm]

  \raisebox{3.5\normalbaselineskip}[0pt][0pt]{\rotatebox[origin=c]{90}{\texttt{Tomyum 1}}} &
  \includegraphics[width=\sz\linewidth]{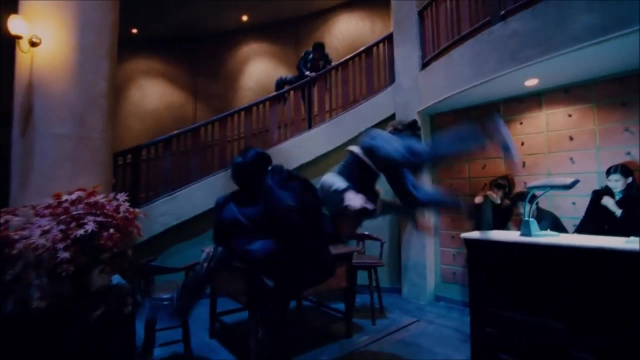}  &
  \includegraphics[width=\sz\linewidth]{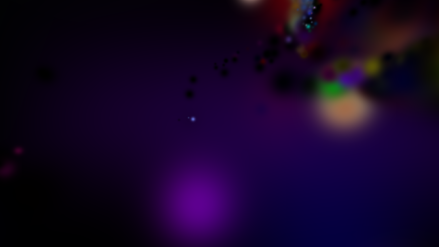} &
  \includegraphics[width=\sz\linewidth]{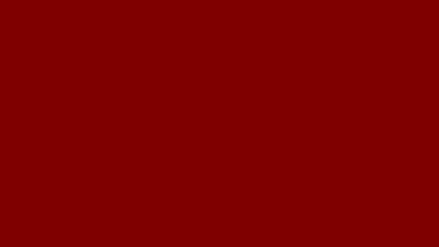} &
  \includegraphics[width=\sz\linewidth]{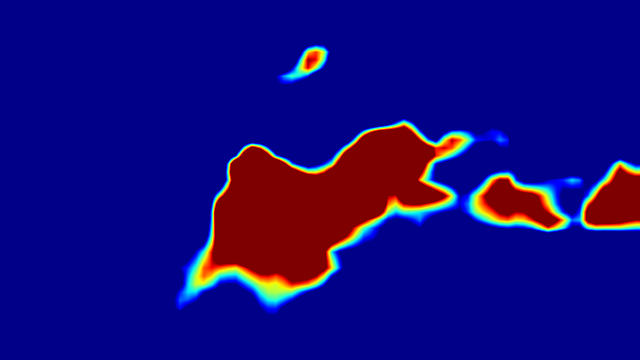}
  \\[-0.1mm]

  \raisebox{3.5\normalbaselineskip}[0pt][0pt]{\rotatebox[origin=c]{90}{\texttt{Tomyum 2}}} &
  \includegraphics[width=\sz\linewidth]{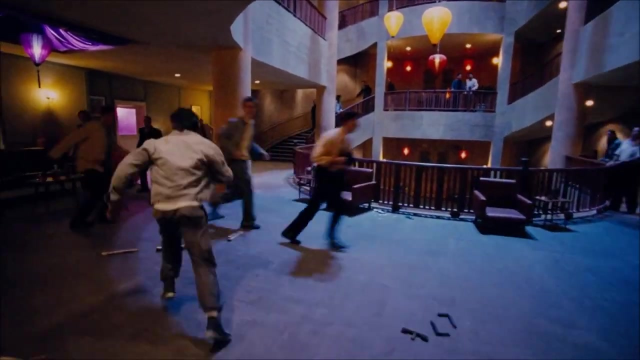}  &
  \includegraphics[width=\sz\linewidth]{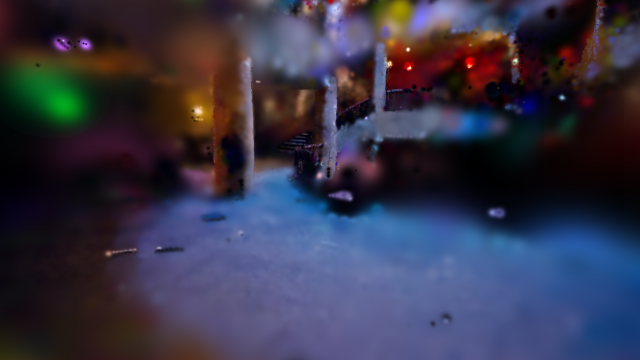} &
  \includegraphics[width=\sz\linewidth]{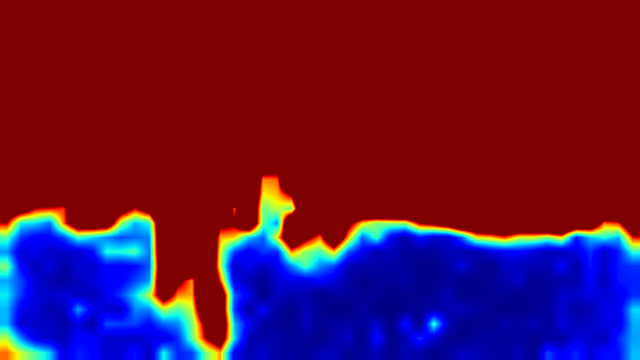} &
  \includegraphics[width=\sz\linewidth]{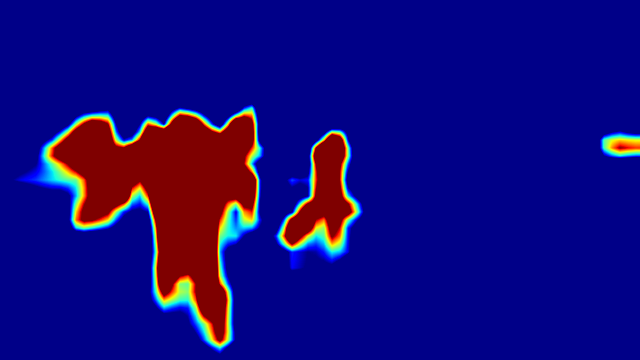}
  \\[-0.1mm]

  & {\fontsize{7}{8} \selectfont Input}
  & {\fontsize{7}{8} \selectfont Rendered (WildGS-SLAM)} 
  & {\fontsize{7}{8} \selectfont Uncertainty (WildGS-SLAM)} 
  & {\fontsize{7}{8} \selectfont Uncertainty (Ours)}
  \\
     
  \end{tabular}
    \caption{\textbf{Uncertainty Estimation.}}
  \label{fig:uncer_youtube}
\end{figure*} 

\begin{figure*}
    \centering
    \includegraphics[width=0.81\linewidth]{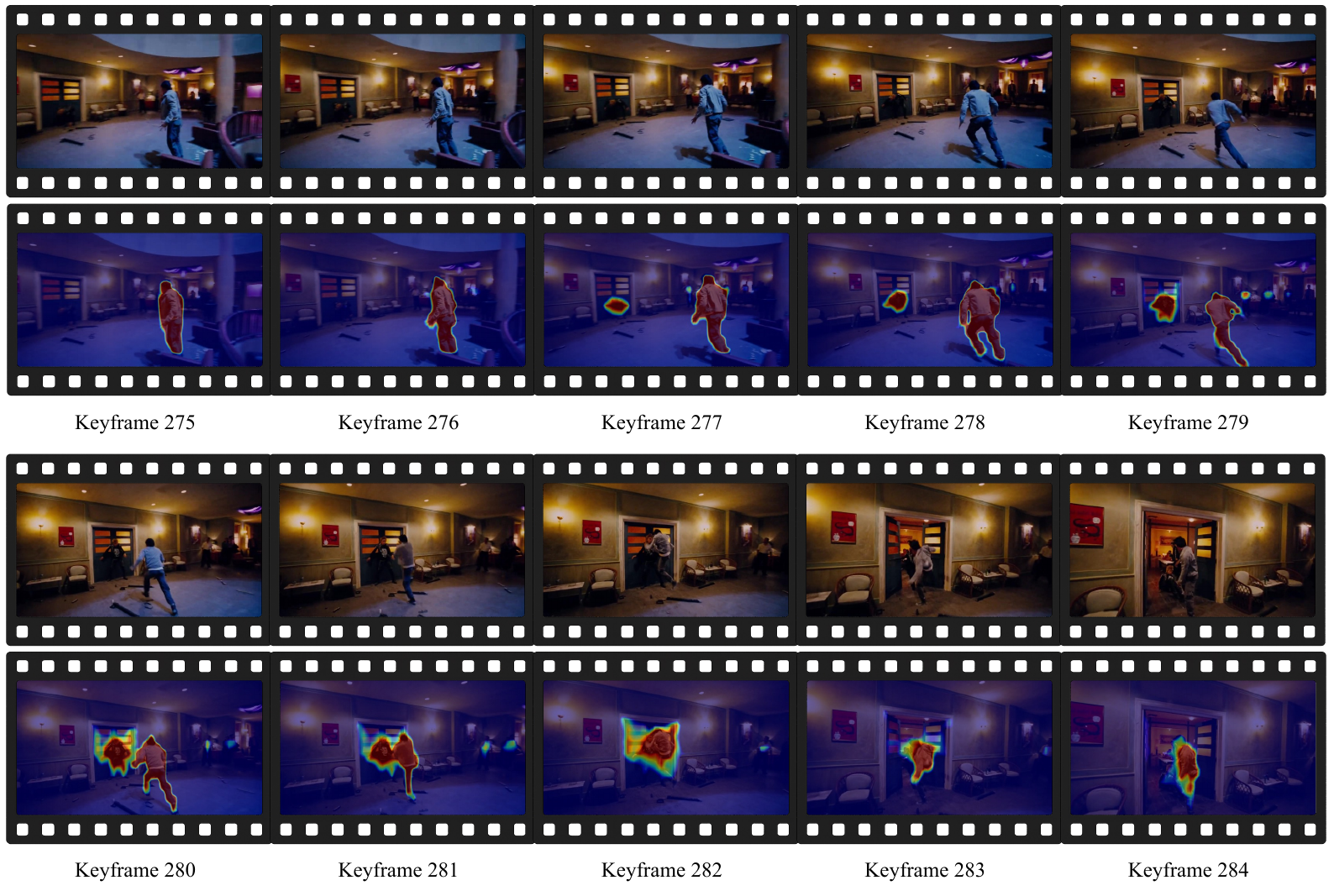}
    \caption{\textbf{Uncertainty Visualization for Consecutive Frames of YouTube \texttt{Tomyum 1}.} We observe that our method robustly handles scenarios in which an object transitions from a static state to dynamic motion, such as a door being pushed open by a person. Prior to the onset of motion, our approach leverages stable visual correspondences on the door to help tracking, since our uncertainty optimization is based on frame-to-frame feature alignment.}
    \label{fig:uncer_series1}
\end{figure*}

\begin{figure*}
    \centering
    \includegraphics[width=0.81\linewidth]{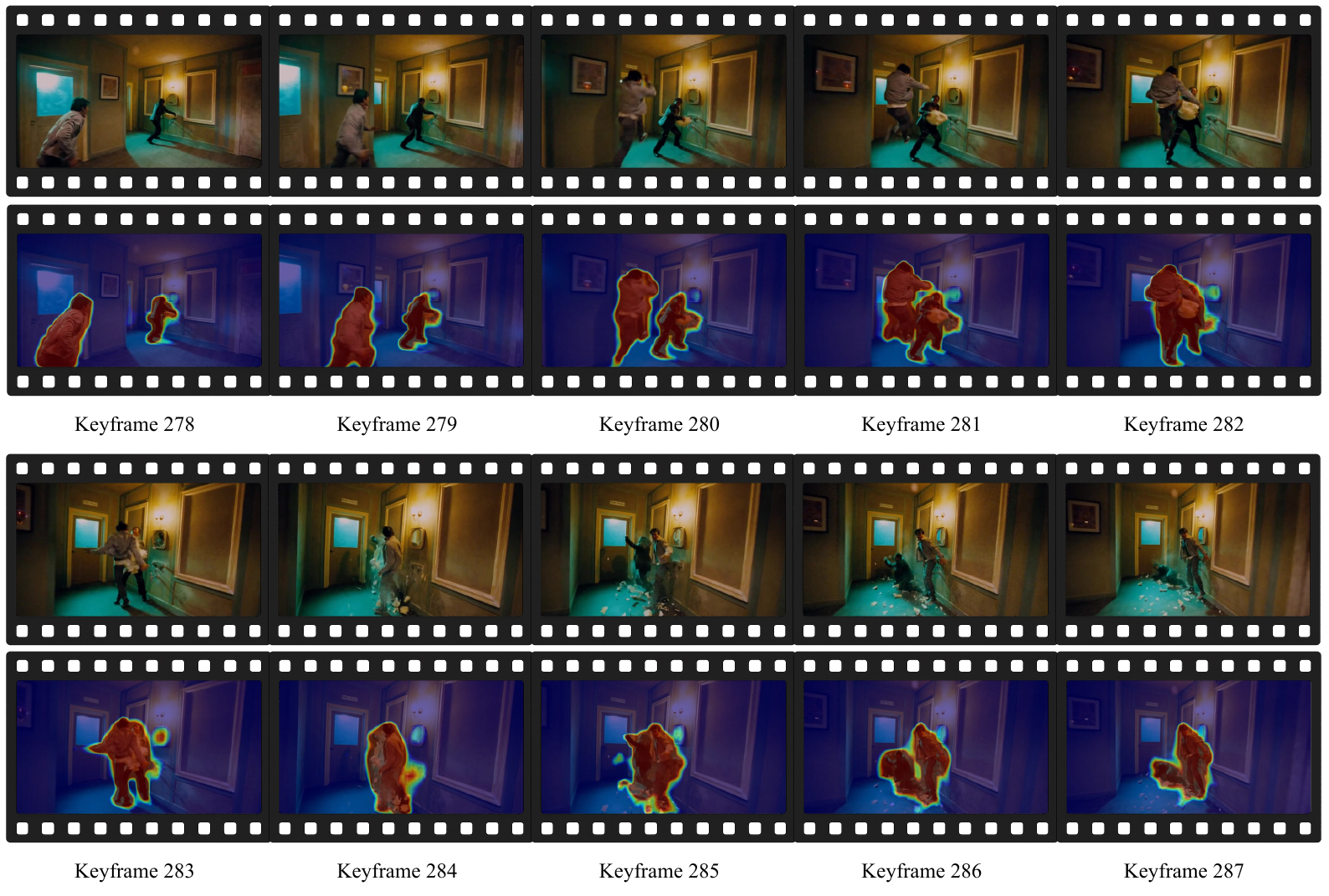}
    \caption{\textbf{Uncertainty Visualization for Consecutive Frames of YouTube \texttt{Tomyum 2}.} Our approach assigns high uncertainty to regions exhibiting strong view-dependent effects (e.g., the mirror).}
    \label{fig:uncer_series3}
\end{figure*}
\paragraph{Uncertainty Comparisons on YouTube Videos}
We provide additional qualitative results on uncertainty estimation in \figref{fig:uncer_youtube}. As shown in \figref{fig:uncer_youtube}, WildGS-SLAM~\cite{zheng2025wildgs} always fails to construct the reliable Gaussian map, leading to inaccurate and noisy uncertainty predictions. In contrast, our method leverages frame-to-frame feature alignment, demonstrating significantly greater robustness in visually complex and truly in-the-wild environments.

Moreover, our approach effectively handles strong view-dependent effects such as reflections and shadows (\eg reflections in \texttt{Taylor 22} and \texttt{Tokyo Walking 2}). It is also highly sensitive to small dynamic objects, enabling precise uncertainty estimation even under challenging real-world conditions. Our method further exhibits strong robustness to severe motion blur and low dynamic range (\eg \texttt{Tomyum 1} and \texttt{Tomyum 2}), which are extremely difficult for conventional segmentation or detection approaches.

Overall, \figref{fig:uncer_youtube} highlights the accuracy and robustness of our uncertainty optimization in unconstrained in-the-wild settings, effectively delineating uncertain regions while maintaining high confidence in static areas.

\paragraph{Uncertainty Visualization for Consecutive Keyframes}
We visualize the optimized uncertainties across consecutive keyframes in \figref{fig:uncer_series1} and \figref{fig:uncer_series3}. Our method optimizes frame-wise uncertainty by exploiting multi-view feature similarity, allowing it to fully leverage static-scene information whenever available. As shown in \figref{fig:uncer_series1}, the system effectively utilizes the door region for camera tracking prior to keyframe 280.
In addition, our uncertainty estimation integrates multi-view cues from frames connected through the frame graph, \ie it captures multi-view inconsistency within a local window. Thus, our approach assigns the reasonable higher uncertainty to the door region of keyframes 280/281 before the door begins to move.

We observe strong mirror-reflection effects in \figref{fig:uncer_series3}, with keyframe 283 exhibiting the largest appearance change. Accordingly, our method assigns the highest uncertainty to keyframe 283, while maintaining relatively low uncertainty for the remaining keyframes that show only minor appearance differences. This behavior demonstrates the effectiveness of our approach and the precision of the resulting uncertainty estimates.

\subsection{Point Cloud Reconstruction}
\label{sec:additional_recon}
\begin{figure*}[!ht]
  \centering
  \footnotesize
  \setlength{\tabcolsep}{4.0pt}
  \newcommand{\wid}{0.3} 
  \newcommand{\het}{0.2} 

  \begin{tabular}{c c: c : c}
  \raisebox{5.0\normalbaselineskip}[0pt][0pt]{\rotatebox[origin=c]{90}{\texttt{Taylor 22}}} &
  \includegraphics[height=\het\linewidth]{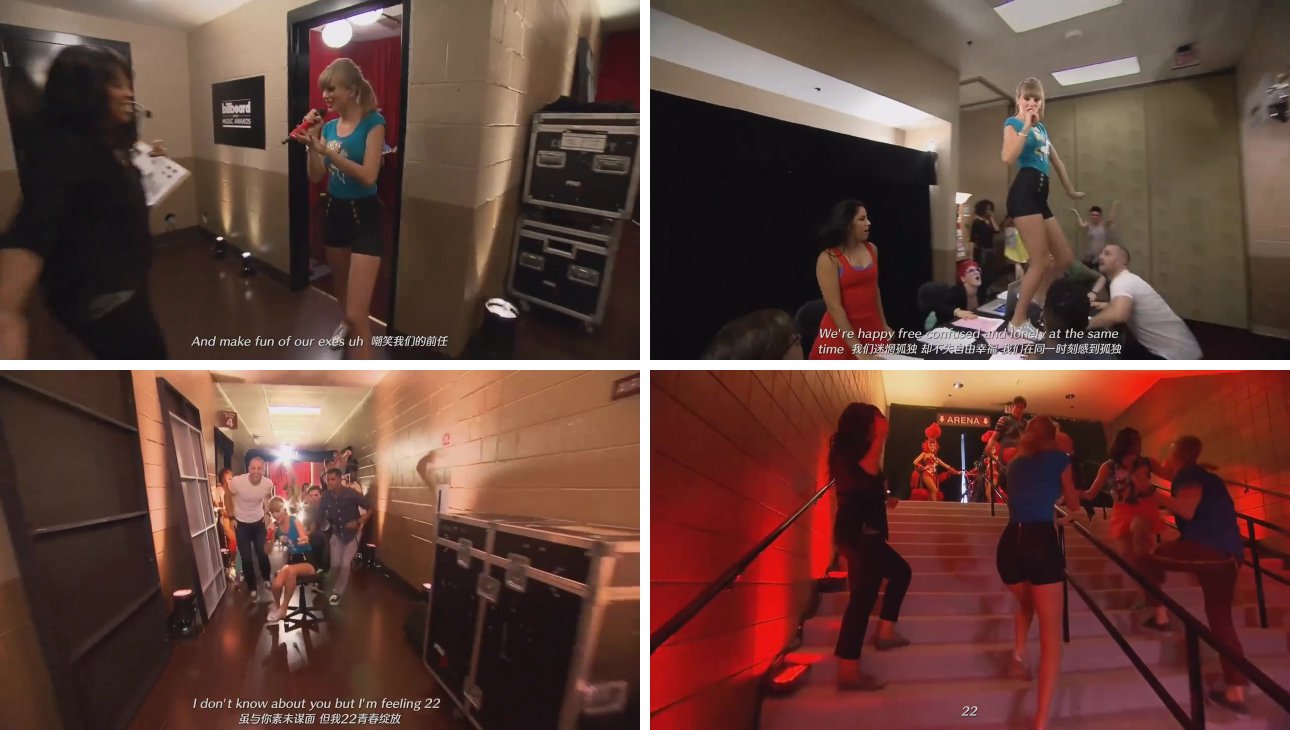} &
  \includegraphics[height=\het\linewidth]{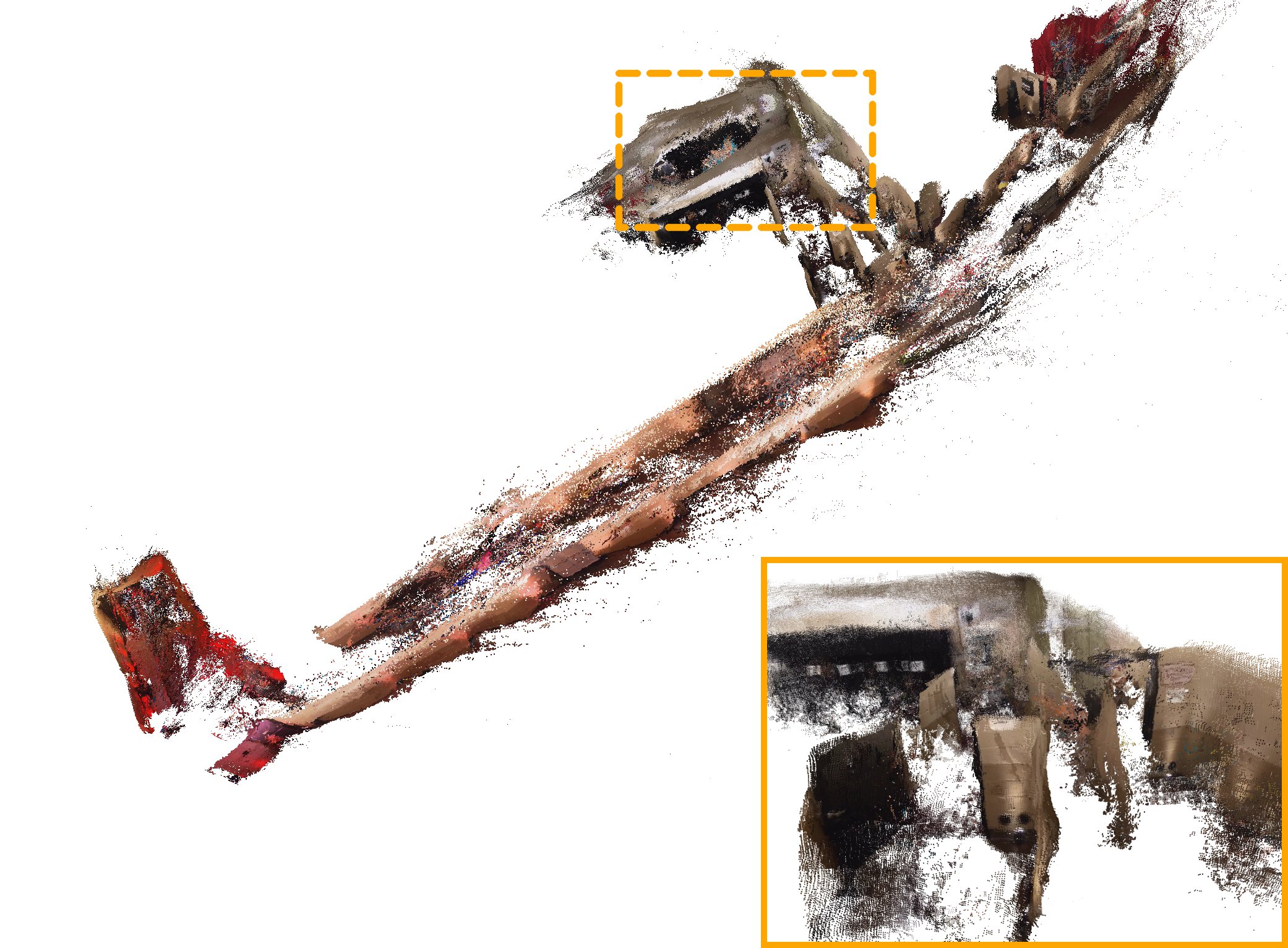} &
  \includegraphics[height=\het\linewidth]{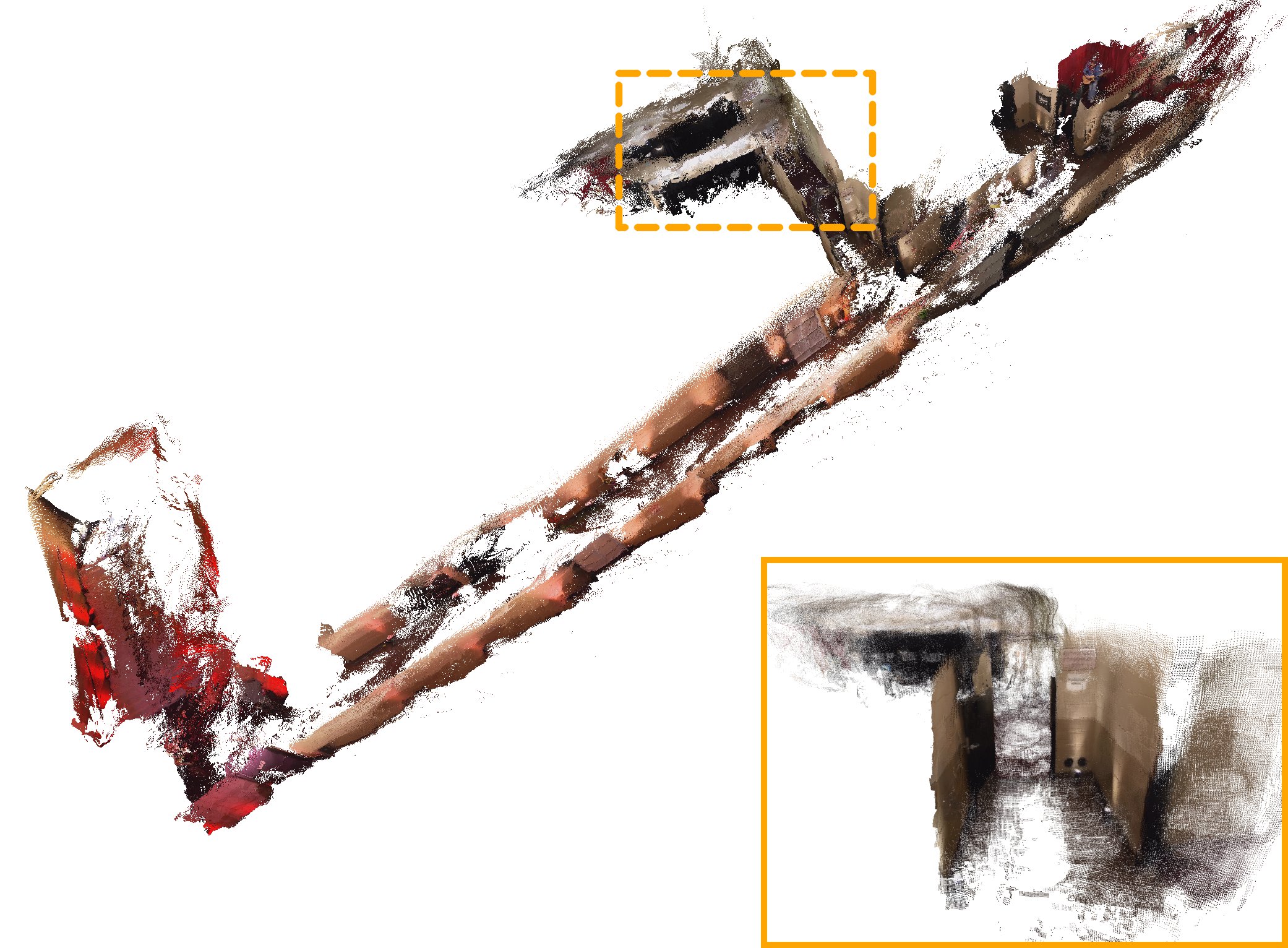} 
  \\[1.0mm]

  \raisebox{5.0\normalbaselineskip}[0pt][0pt]{\rotatebox[origin=c]{90}{\texttt{Tokyo Walking 1}}} &
  \includegraphics[height=\het\linewidth]{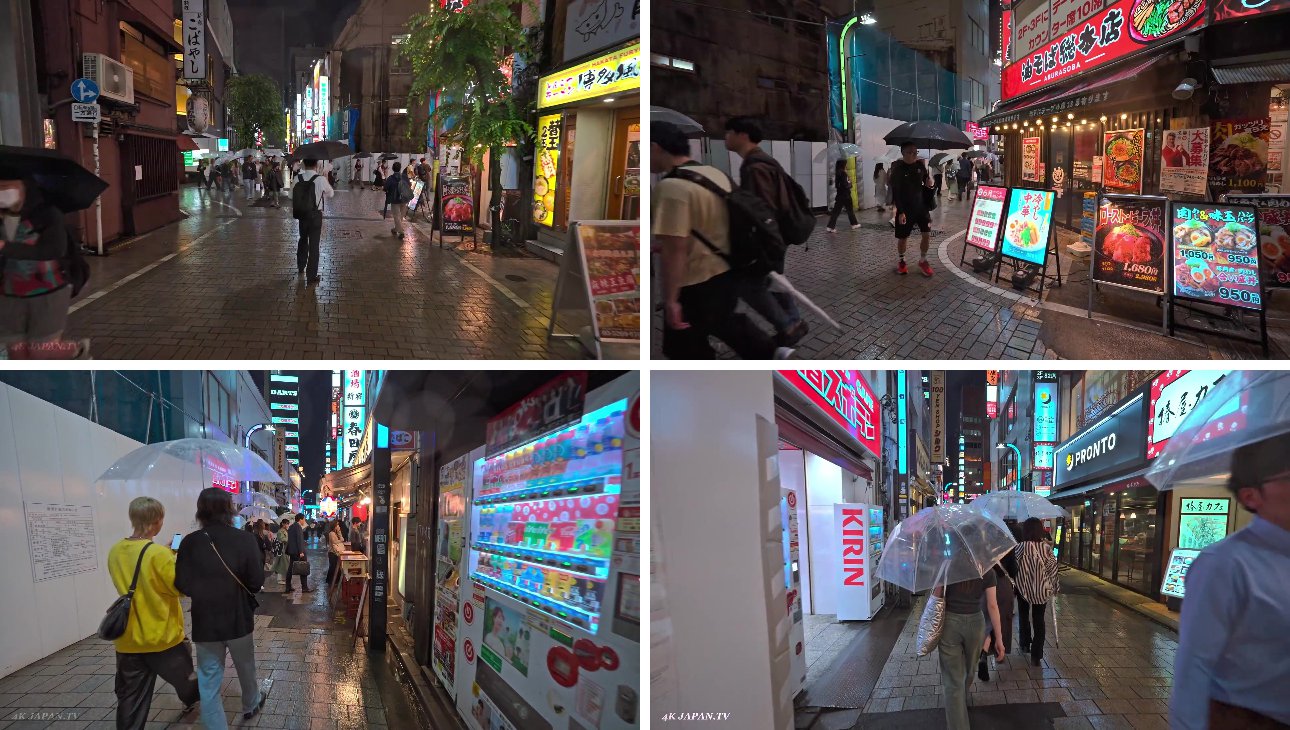} &
  \includegraphics[height=\het\linewidth]{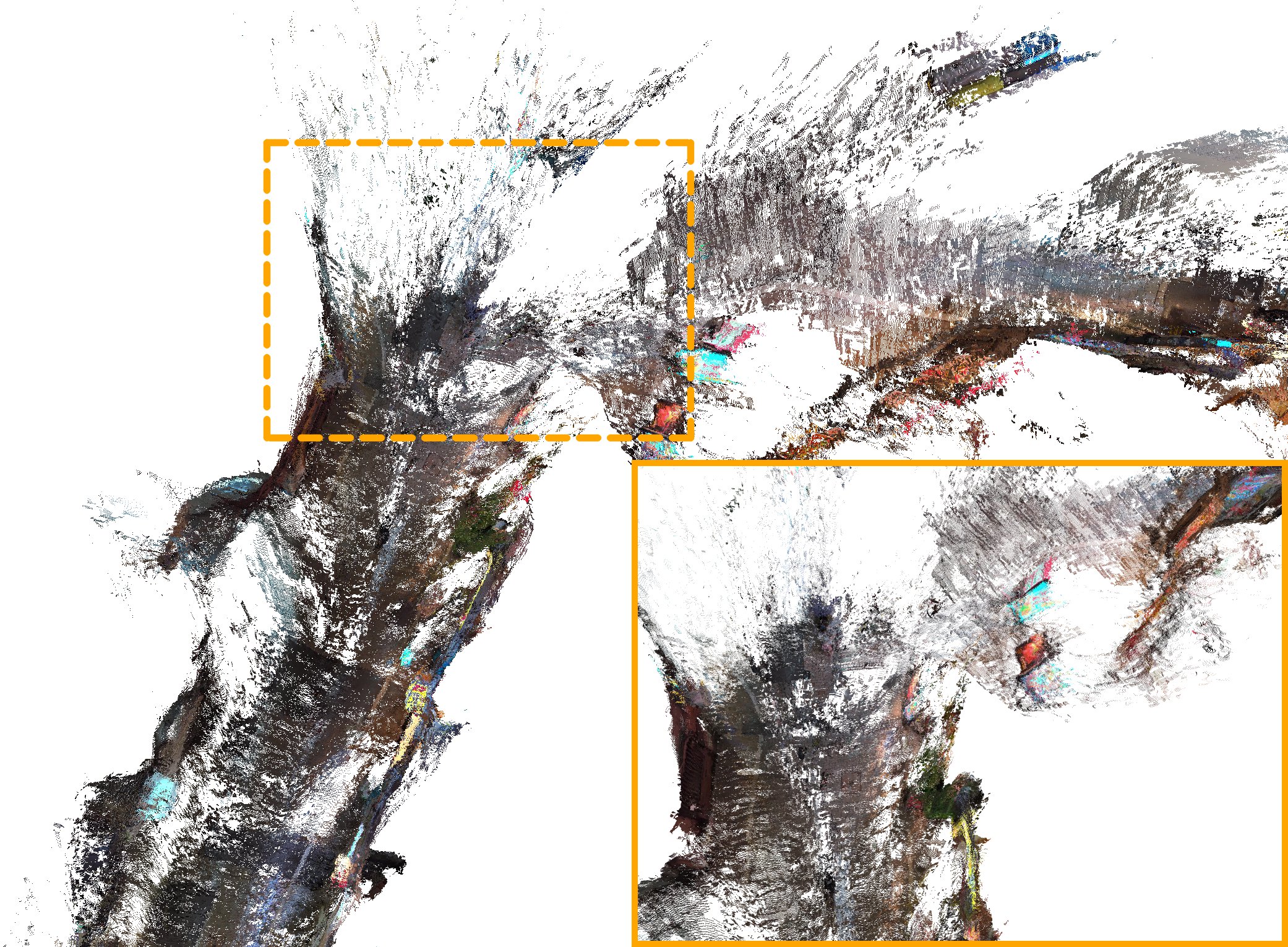} &
  \includegraphics[height=\het\linewidth]{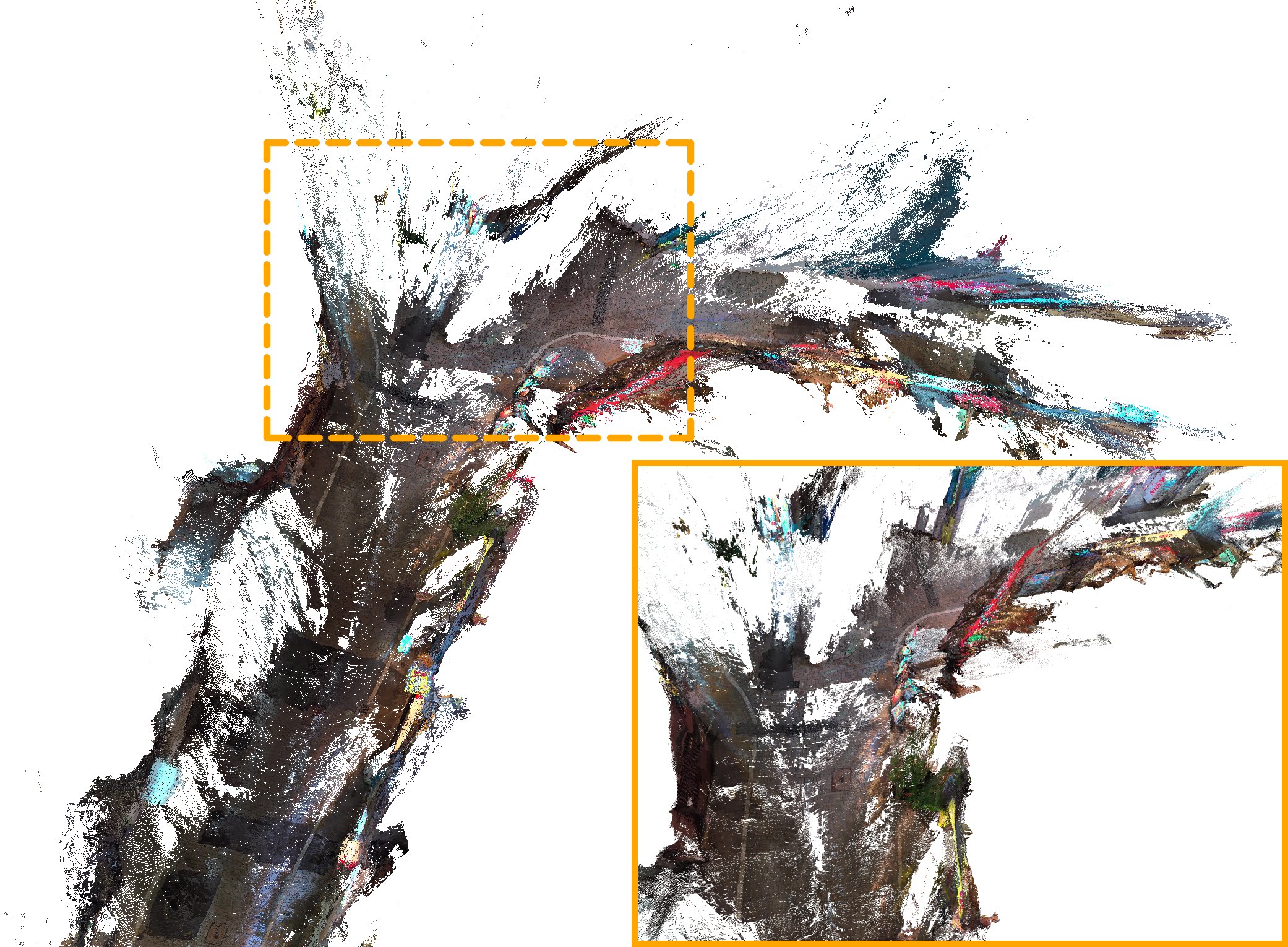} 
  \\[1.0mm]
  
  & Inputs & DROID-SLAM \cite{teed2021droid} & \project \\

  \end{tabular}

  \caption{\textbf{3D Reconstruction Comparisons on YouTube Sequences.}
  We compare reconstructed static point clouds between DROID-SLAM~\cite{teed2021droid} and our method. Our approach produces more accurate and consistent reconstructions across highly dynamic and visually challenging real-world sequences. For the \texttt{Taylor 22} scene, DROID-SLAM reconstructs two separate corridor structures due to inaccurate pose tracking, as highlighted in the boxed region, whereas our method produces a consistent geometric reconstruction. Moreover, the white lane marking on the asphalt road of \texttt{Tokyo Walking 1} is faint and fragmented in the point cloud reconstructed by DROID-SLAM, while in our reconstruction it remains cleanly visible, continuous, and highly accurate.
  }
  \label{fig:sup_point_clouds_comparisons}
\end{figure*}

\paragraph{Static Reconstruction Comparisons}
We present 3D reconstruction comparisons between DROID-SLAM~\cite{teed2021droid} and our method in \figref{fig:sup_point_clouds_comparisons}. In the top row, DROID-SLAM fails to recover consistent geometry -- erroneously reconstructing a single corridor as two separate structures -- and exhibits noticeable tracking drift in sequences with the presence of strong dynamic distractors. In contrast, our approach produces coherent geometric reconstructions and accurate pose estimates. The second row further highlights the robustness of our method: it reconstructs cleanly visible and highly accurate white lane markings on the asphalt road, even under challenging real-world conditions.

\begin{figure*}
    \centering
    \includegraphics[width=1.0\linewidth]{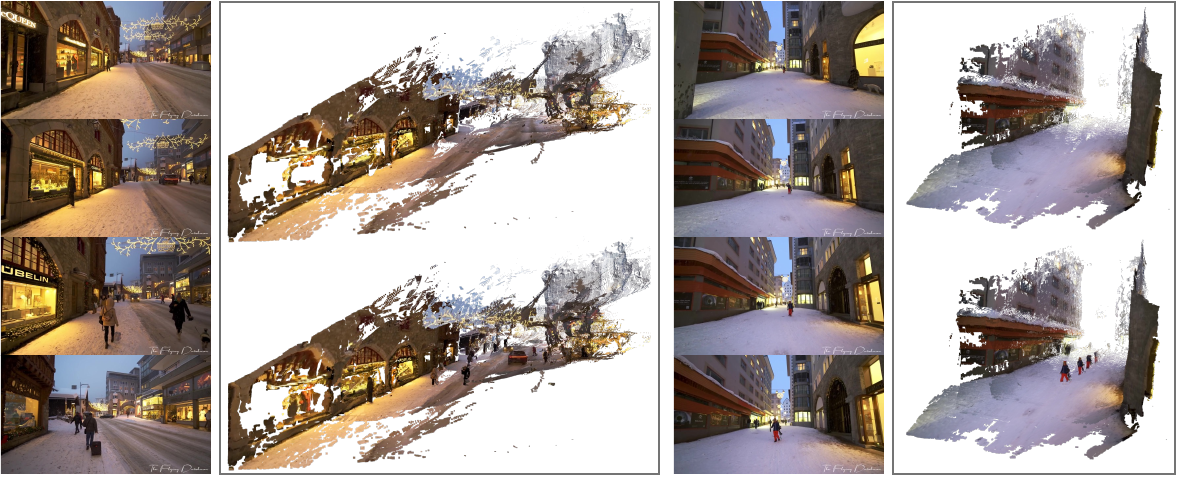}
    \caption{\textbf{Point Clouds Visualization from 4 Views.} 
    \textit{Top view}: reconstructed static scene from 4 input views shown in the figure. \textit{Bottom view}: reconstructed dynamic point clouds. 
    The comparisons between the dynamic and static point clouds further demonstrate the effectiveness of our uncertainty estimation. 
    In both visualizations, the static point clouds remain highly consistent, indicating that our method reliably preserves static geometry while filtering dynamic regions.
    }
    \label{fig:sup_point_clouds_4_views}
\end{figure*}

\begin{figure*}[!ht]
  \centering
  \footnotesize
  \setlength{\tabcolsep}{4.0pt}
  \newcommand{\sz}{0.45} 

  \begin{tabular}{c c: c}

  \raisebox{7.0\normalbaselineskip}[0pt][0pt]{\rotatebox[origin=c]{90}{\texttt{St. Moritz 2}}} &
  \includegraphics[width=\sz\linewidth]{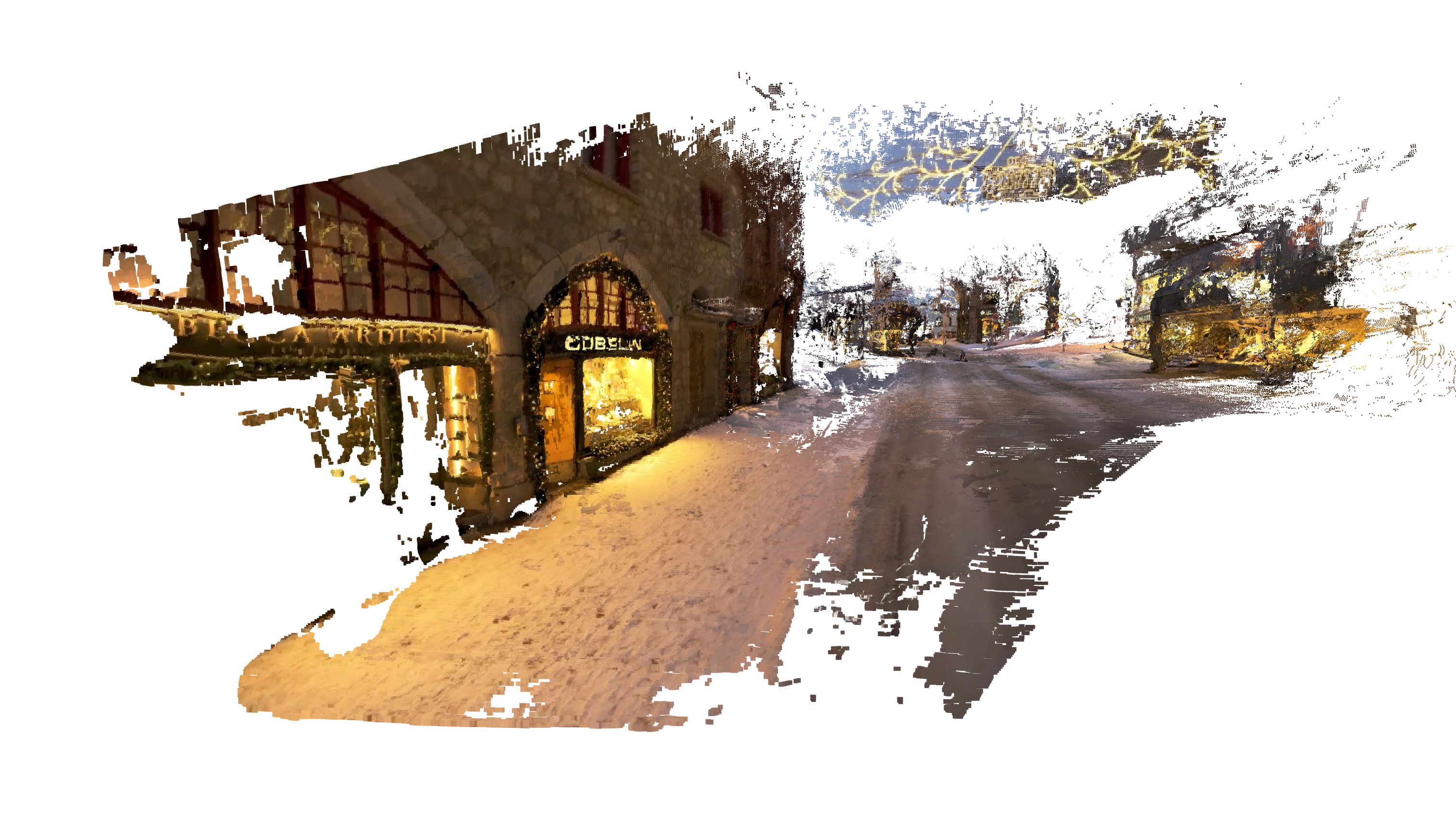} &
  \includegraphics[width=\sz\linewidth]{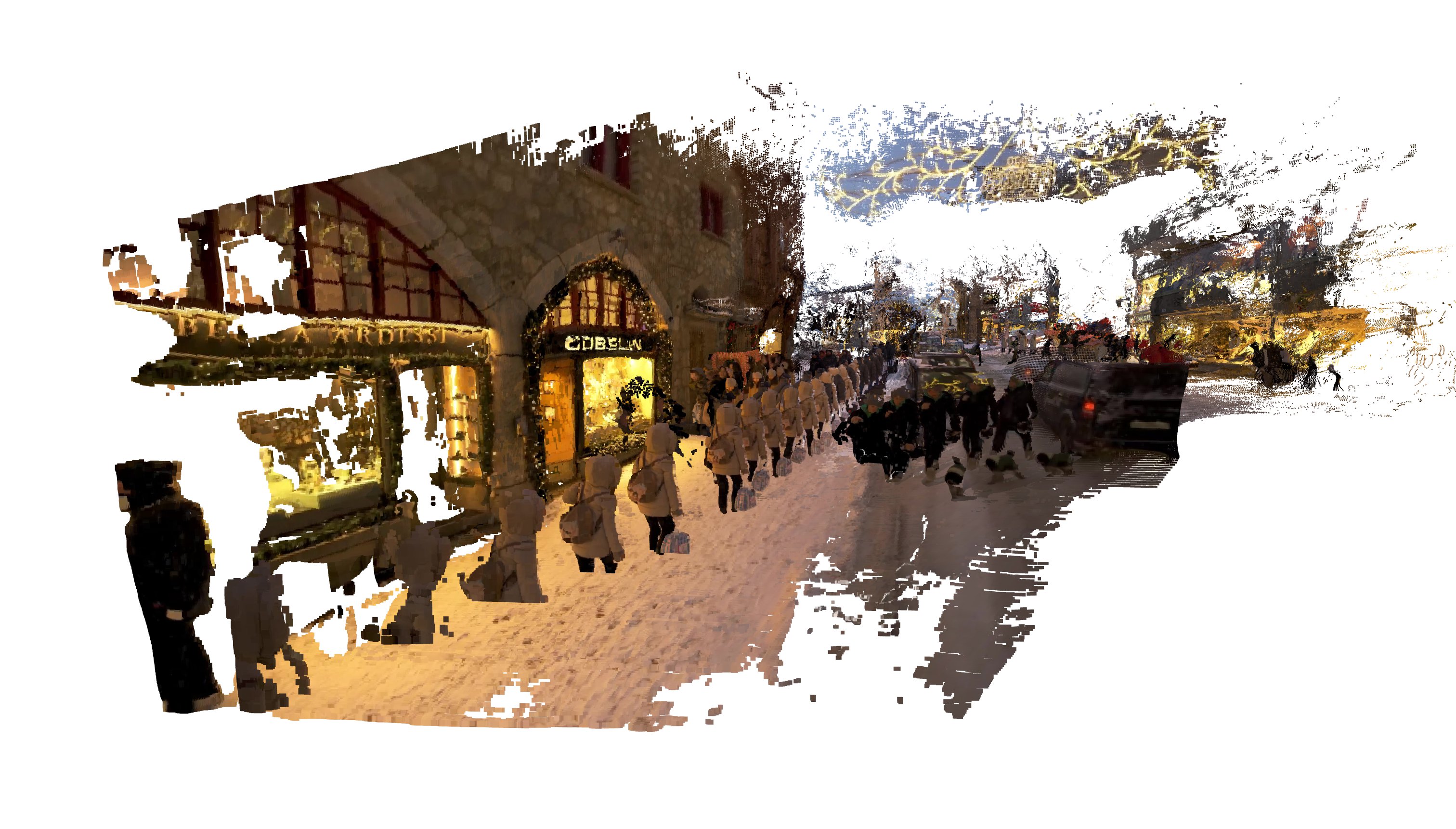} 
  \\[1.0mm]

  \raisebox{7.0\normalbaselineskip}[0pt][0pt]{\rotatebox[origin=c]{90}{\texttt{St. Moritz 3}}} &
  \includegraphics[width=\sz\linewidth]{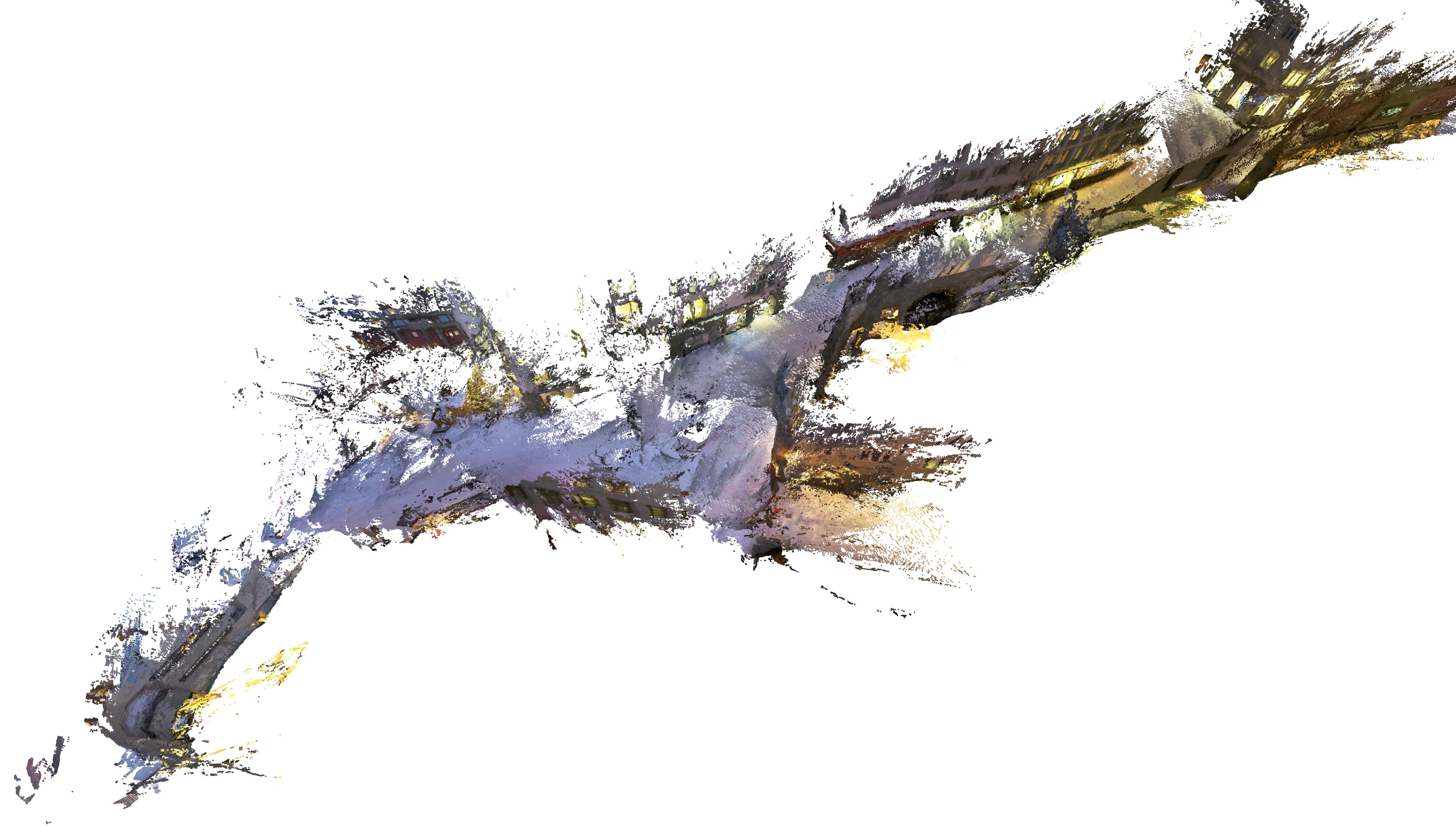} &
  \includegraphics[width=\sz\linewidth]{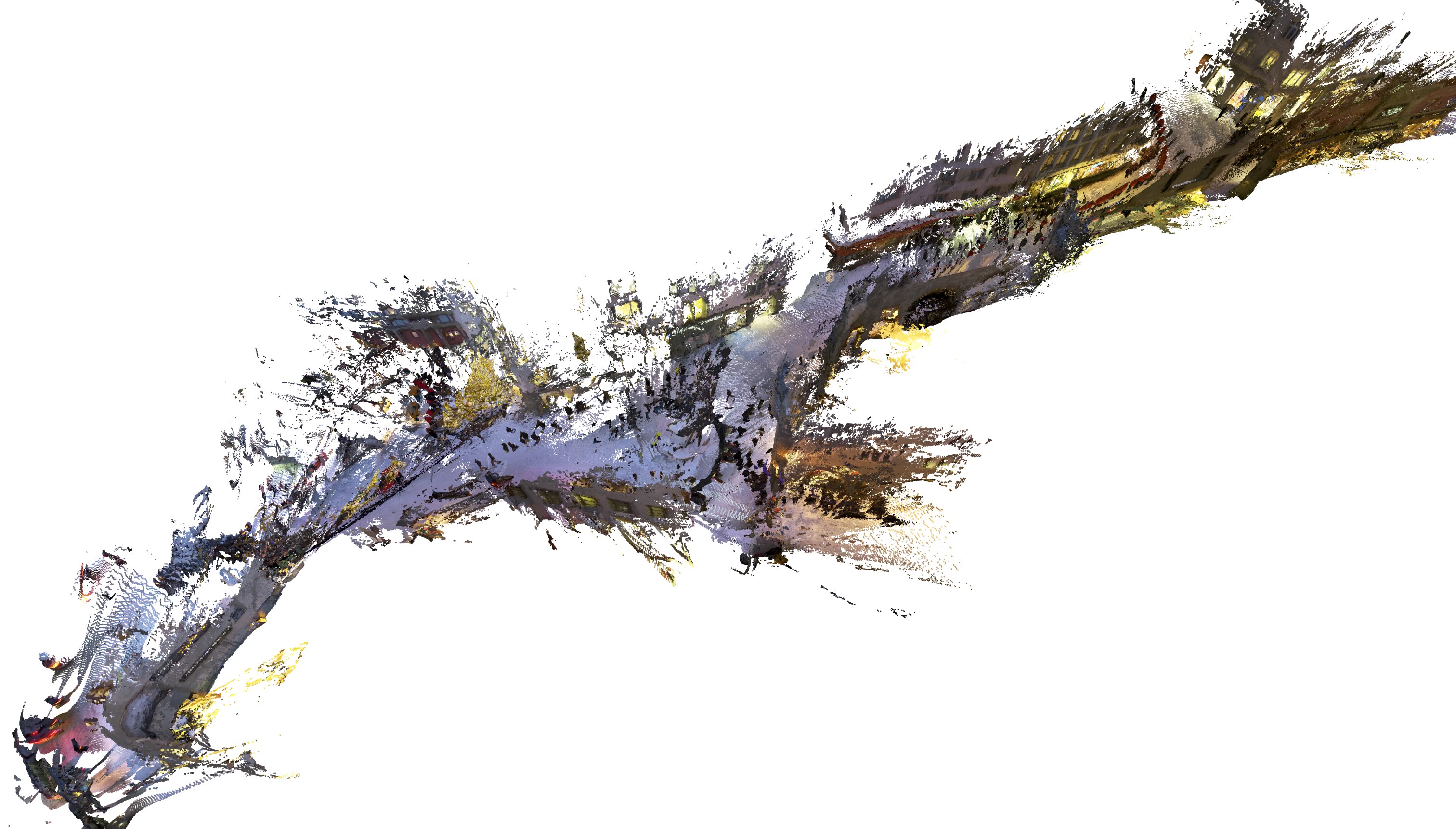} 
  \\[1.0mm]

  \raisebox{7.0\normalbaselineskip}[0pt][0pt]{\rotatebox[origin=c]{90}{\texttt{St. Moritz 4}}} &
  \includegraphics[width=\sz\linewidth]{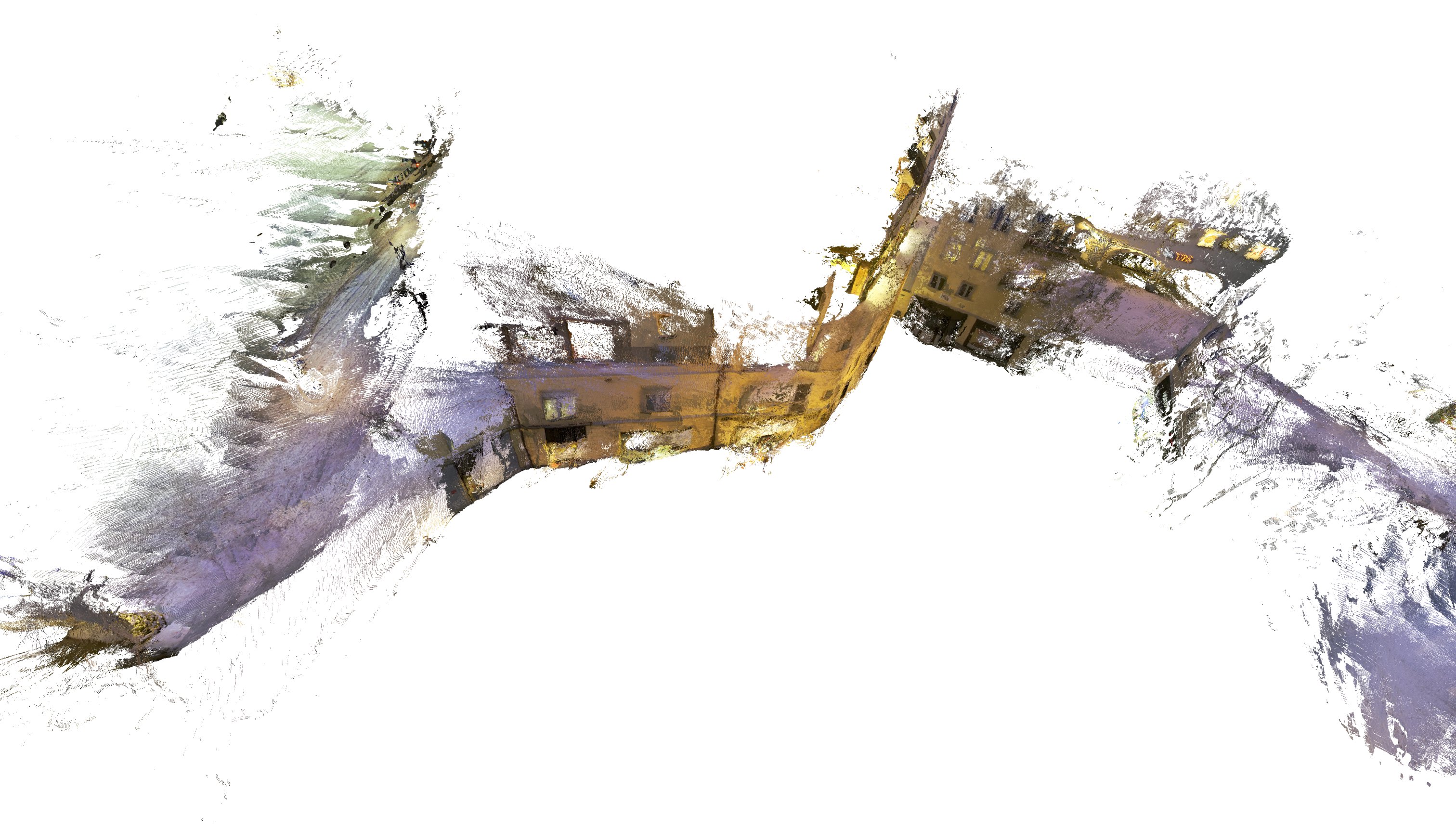} &
  \includegraphics[width=\sz\linewidth]{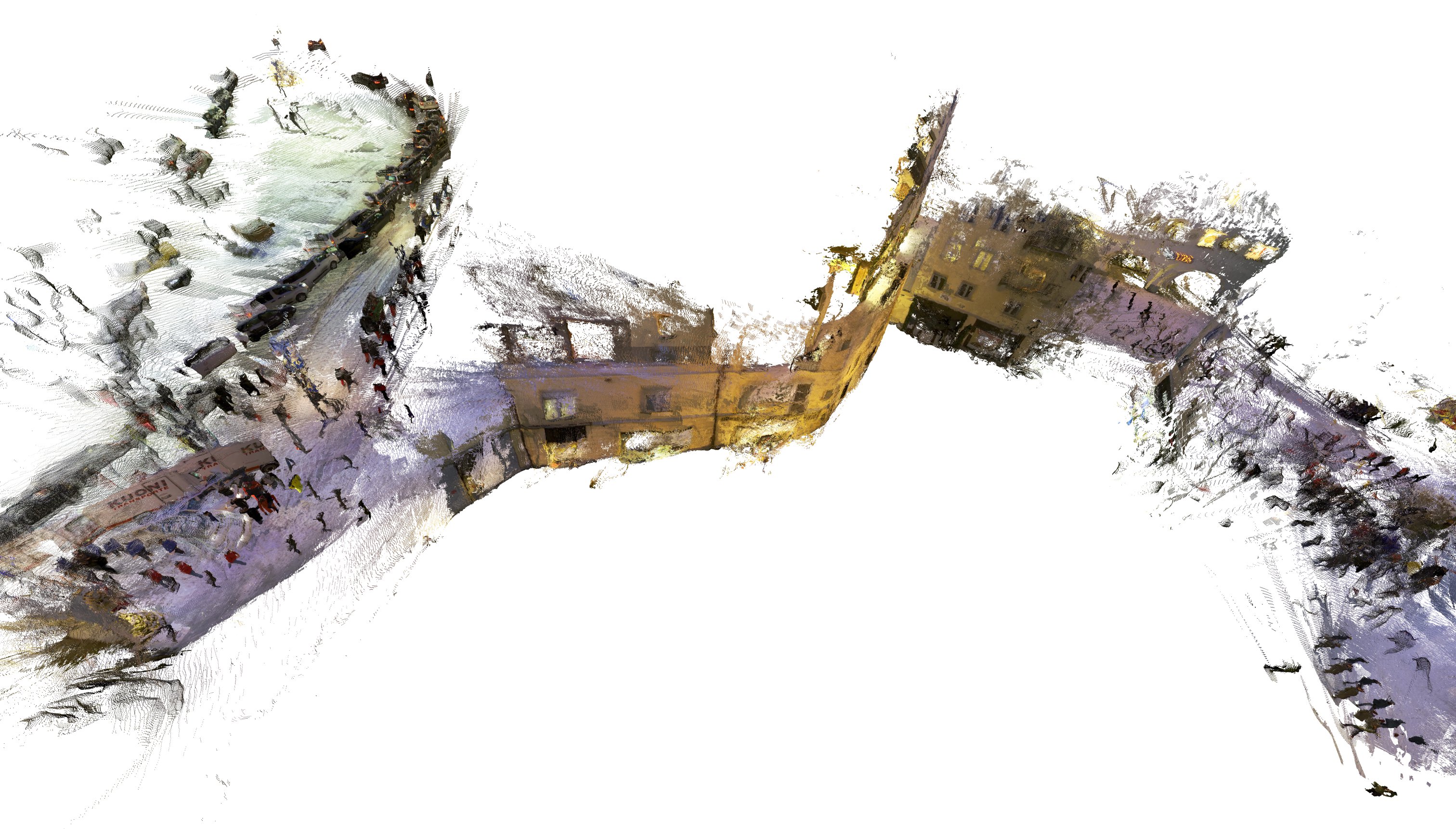} 
  \\[1.0mm]
  
  \raisebox{7.0\normalbaselineskip}[0pt][0pt]{\rotatebox[origin=c]{90}{\texttt{St. Moritz 5}}} &
  \includegraphics[width=\sz\linewidth]{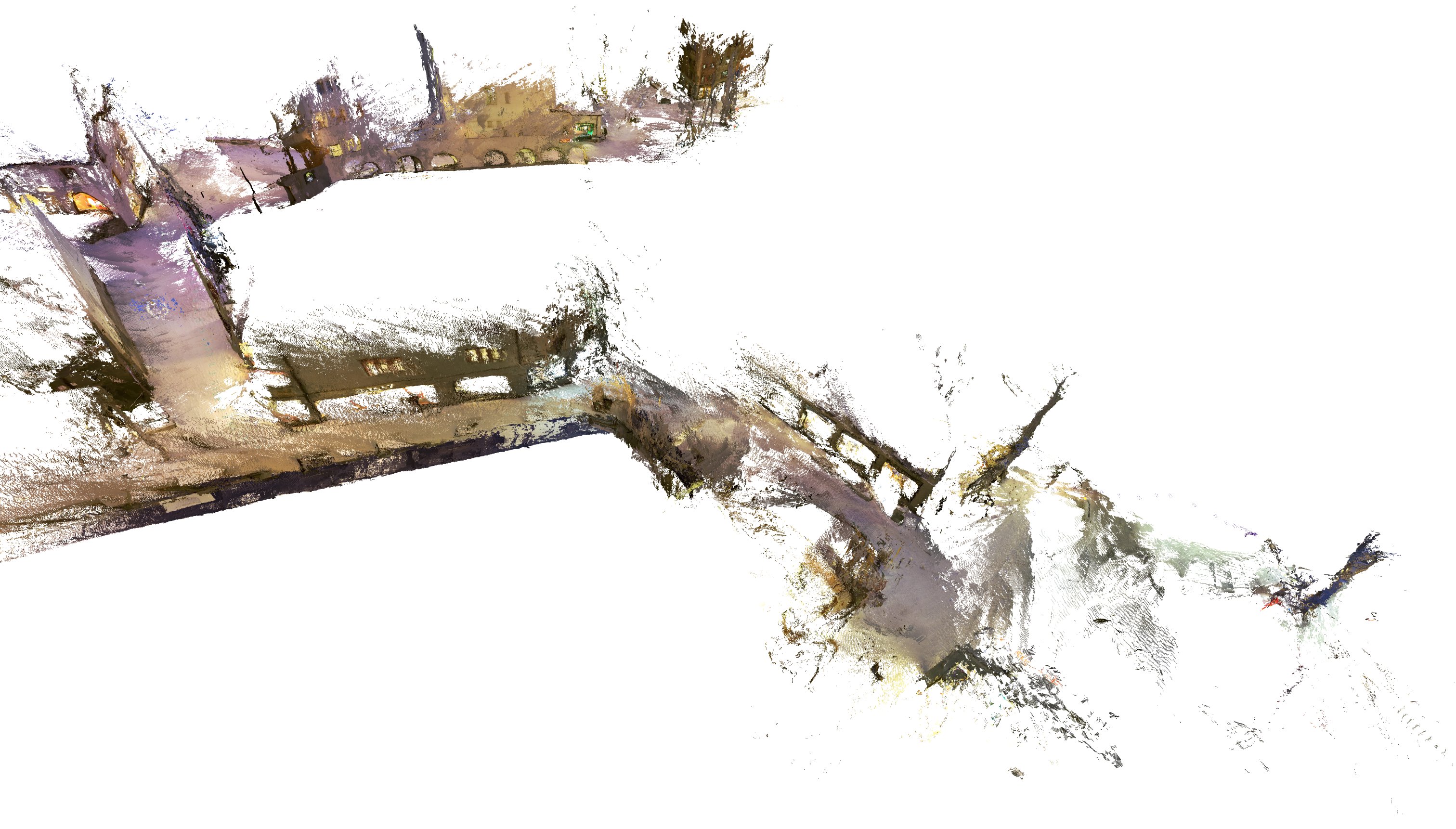} &
  \includegraphics[width=\sz\linewidth]{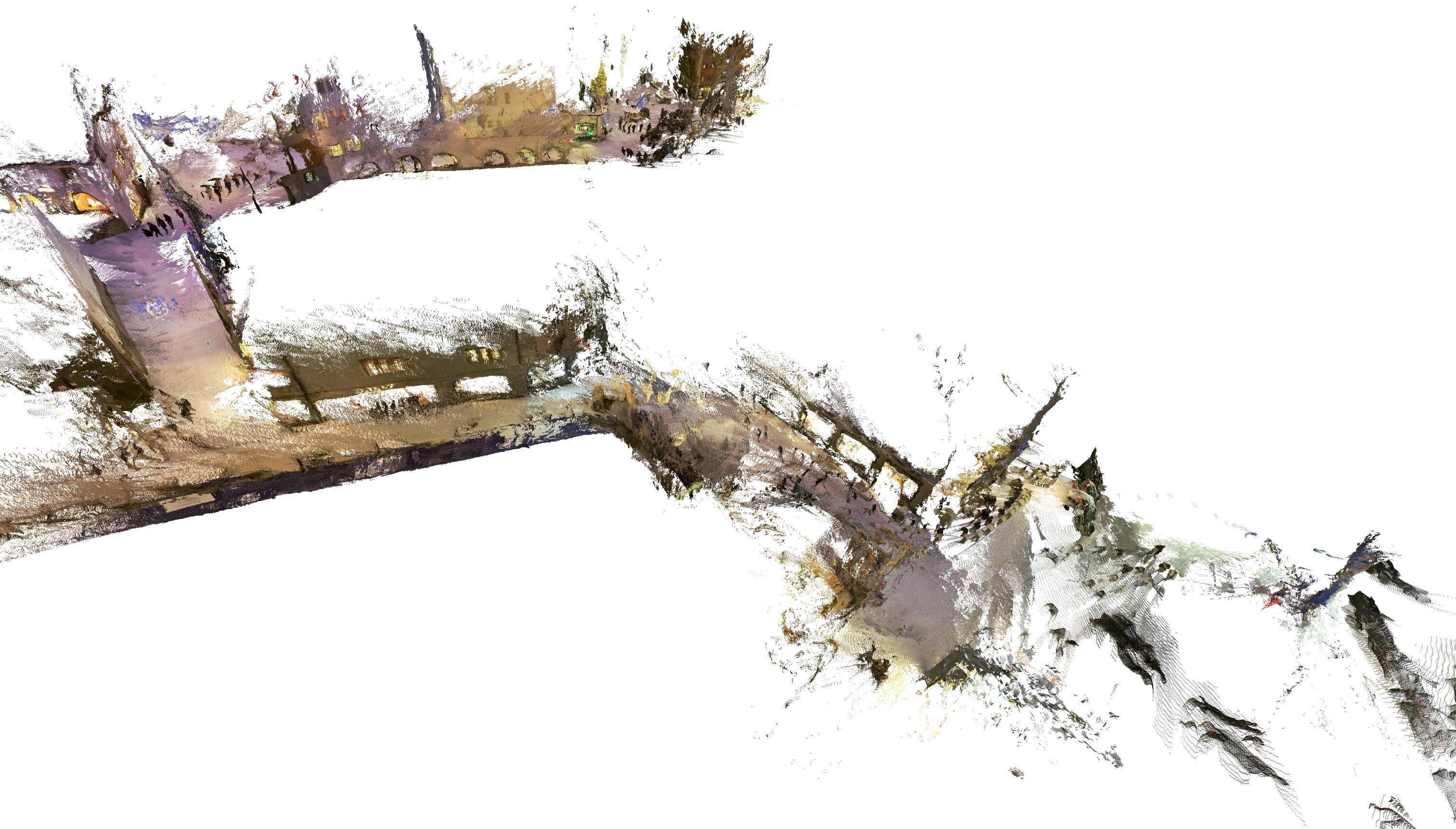} 
  \\[1.0mm]
  
  & {Static Reconstruction} & {Dynamic Reconstruction} \\

  \end{tabular}

  \caption{\textbf{Qualitative Results of Our Static and Dynamic Reconstruction.}
  We visualize globally aligned static reconstructions alongside dynamic point clouds across all keyframes. Notably, we apply the estimated per-frame dynamic uncertainty to filter out dynamic points. 
  The left and right columns show the static and dynamic reconstructions, respectively. 
  These comparisons highlight the accuracy of our uncertainty estimation, as our method effectively suppresses dynamic regions while preserving geometric fidelity in static areas.
  }
  \label{fig:sup_point_clouds_all_views}
\end{figure*}

\paragraph{Static / Dynamic Reconstruction}
We visualize the static and dynamic point clouds in \figref{fig:sup_point_clouds_4_views} and \figref{fig:sup_point_clouds_all_views}, providing complementary perspectives from both top-down and interior viewpoints. The high geometric consistency between our static reconstruction and the static regions within the dynamic point clouds illustrates the precision of our uncertainty estimation. These results demonstrate that our method effectively suppresses dynamic or uncertain regions while preserving the underlying static scene structure.

\subsection{Additional Ablation Study}
\begin{table}[t]
\centering
\footnotesize
\setlength{\tabcolsep}{8pt}
{
    \begin{tabular}{lc}
        \toprule
        {Method} & ATE RMSE [cm] \\
        \midrule
        w/o prior term & 5.18 \\
        \textbf{Full} & \textbf{2.30} \\
        \bottomrule
    \end{tabular}
}
\caption{\textbf{Ablation Studies on Bonn RGB-D Dataset~\cite{palazzolo2019iros}.}
}
\label{tab:sup_ablation_study}
\end{table}
\tabref{tab:sup_ablation_study} shows that the prior regularization term effectively avoid the trivial solution $\bu \rightarrow \infty$. Without this prior, the system assigns uniformly large uncertainties to all pixels, resulting in results similar to the \textit{w/o Uncertainty-aware BA} configuration reported in \tabrefn{tab:ablation_study} of the main paper.

\end{document}